%% file: article.tex
\documentclass[article, shortnames, nojss]{jss}

\usepackage{orcidlink,thumbpdf,lmodern}

\usepackage{framed}

\usepackage{amsmath, amssymb}
\usepackage{bm}
\usepackage{caption}
\usepackage{tikz}
\usepackage{wrapfig}
\usepackage{booktabs}
\usepackage{multirow}
\usepackage{adjustbox}
\usepackage{subcaption}
\usepackage{pifont}
\usepackage{array}
\usepackage{footmisc}

\usetikzlibrary{patterns}
\usetikzlibrary{arrows.meta}

\def\checkmark{\tikz\fill[scale=0.4](0,.35) -- (.25,0) -- (1,.7) -- (.25,.15) -- cycle;}
\newcommand{\xmark}{\ding{54}}%

\definecolor{BIPSBlue}{RGB}{23, 99, 170}
\definecolor{torchorange}{HTML}{EE4C2C}

\author{Niklas Koenen~\orcidlink{0000-0002-4623-8271}\\Leibniz Institute for Prevention\\ Research and Epidemiology – BIPS,\\University of Bremen 
   \And Marvin N. Wright~\orcidlink{0000-0002-8542-6291}\\Leibniz Institute for Prevention\\ Research and Epidemiology – BIPS,\\University of Bremen,\\University of Copenhagen}
\Plainauthor{Niklas Koenen, Marvin N. Wright}
\title{Interpreting Deep Neural Networks with the Package \pkg{innsight}}
\Plaintitle{Interpreting Deep Neural Networks with the Package innsight}
\Shorttitle{Interpreting DNNs with \pkg{innsight}}

\Abstract{
  The \proglang{R} package \pkg{innsight} offers a general toolbox for revealing variable-wise interpretations of deep neural networks' predictions with so-called feature attribution methods. Aside from the unified and user-friendly framework, the package stands out in three ways: It is generally the first \proglang{R} package implementing feature attribution methods for neural networks. Secondly, it operates independently of the deep learning library, allowing the interpretation of neural networks from any \proglang{R} package, including \pkg{keras}, \pkg{torch}, \pkg{neuralnet}, and even custom models. Despite its flexibility, \pkg{innsight} benefits internally from the \pkg{torch} package's fast and efficient array calculations, which builds on \pkg{LibTorch} -- \pkg{PyTorch}'s \proglang{C++} backend -- without a \proglang{Python} dependency. Finally, it offers a variety of visualization tools for tabular, signal, image data, or a combination of these. Additionally, the plots can be rendered interactively using the \pkg{plotly} package. 
}

\Plainkeywords{neural networks, feature attribution, interpretable machine learning, explainable artificial intelligence, XAI, IML, torch, keras} 

\Address{
    Niklas Koenen\\
    Leibniz Institute for Prevention Research and Epidemiology – BIPS\\
    Achterstr. 30\\
    28359 Bremen, Germany\\
    \emph{and}\\
    Faculty of Mathematics and Computer Science\\
    University of Bremen\\
    E-mail: \email{koenen@leibniz-bips.de}\\

    Marvin N. Wright\\
    Leibniz Institute for Prevention Research and Epidemiology – BIPS\\
    Achterstr. 30\\
    28359 Bremen, Germany\\
    \emph{and}\\
    Faculty of Mathematics and Computer Science\\
    University of Bremen, Germany\\
    \emph{and}\\
    Section of Biostatistics, Department of Public Health\\
    University of Copenhagen, Denmark\\
    E-mail: \email{wright@leibniz-bips.de}
}

\begin{document}


\section{Introduction}
\label{sec:intro}

Throughout the past decade, neural networks have unleashed a tremendous surge of attention and infiltrated almost all conceivable domains of science, industry, and public life. Mainly, their increasing popularity is due to their natural ability to extract patterns and knowledge from vast amounts of structured raw data thanks to modern computing capacities and deliver outstanding performance \citep{imagenet_2012, lecun_2015, silver_2016, bengio_2021}. However, the intelligently learned decision-making process of a neural network remains inscrutable and hidden from the user due to its enormous complexity. Interpretations cannot be inferred as straightforward from this so-called \textit{black box} as, for example, the coefficients of a linear model. As a consequence, the gain in predictive accuracy and model flexibility generally comes at the price of an increasingly opaque and intricate machine learning model, as was already noted by \cite{darpa_2019}. Nevertheless, it is precisely this question of interpretability -- or, informally speaking: Why did a network make a certain prediction? -- that is becoming more and more relevant for applications with high-stake decisions and possibly becoming a legal requirement, e.g., in autonomous systems \citep{sullivan_2019}, healthcare \citep{schneeberger_2020} or data processing in general \citep{gdpr_2016, goodman_2016}.

Arising from this question and need, several methods have been proposed to explain predictions of machine learning models in a supervised learning setting. These methods are mainly classified according to the criteria for which model class they are applicable and at which level they provide explanations: Regarding the first criterion, \textit{model-agnostic} approaches analyze the association of input data and model predictions of arbitrary models. Contrary, \textit{model-specific} methods additionally exploit internal structures to reveal insights, but their application is restricted to a specific group of models. Secondly, interpretability methods for machine learning models can be categorized into \textit{local} and \textit{global} in terms of the explanation level. Established local model-agnostic methods, such as Shapley additive explanations (SHAP) \citep{Lundberg_2017}, individual conditional expectations (ICE) \citep{Goldstein_2015}, and local interpretable model-agnostic explanations (LIME) \citep{Ribeiro_2017}, explain only individual or groups of instances from the dataset, e.g., a single picture for image classification or one patient in the context of medical disease prediction. In contrast, global approaches describe the entire model behavior independent of individual effects. Commonly applied methods from this category are permutation feature importance \citep{Fisher_2019}, accumulated local effect (ALE) plots \citep{Apley_2020} and partial dependence plots \citep{Friedman_2001, Greenwell_2018}. One way to describe this distinction is to look at the classical linear model with input variables $\bm{x} = (x_1, \ldots, x_p)^\top$ and prediction $\hat{y}$:
\begin{align*}
    \hat{y} = \beta_0 + x_1\, \beta_1 + \ldots + x_p\, \beta_p.
\end{align*}
In this case, the coefficients $\beta_1, \ldots, \beta_p$ describe the global effect of the input variables $x_1, \ldots, x_p$ taking their values independent of $\bm{x}$, i.e., they indicate how much the variables affect the prediction $\hat{y}$ in general. On the other hand, the local explanations $x_1\, \beta_1, \ldots, x_p\, \beta_p$ demonstrate how much each input variable contributes to or impacts the prediction for a chosen input instance $\bm{x}$, commonly leading to a variable-wise decomposition of $\hat{y}$ in additive effects. 

\begin{figure}[t]
    \centering
    \resizebox{0.95\textwidth}{!}{
	\input{tikz/feature_attribution.tikz}
	}
    \caption{General procedure of feature attribution methods: First, an input instance $\bm{x}$ flows through the model $f$ to obtain a prediction $\bm{\hat{y}}$. Then, the desired output node or class $\hat{y}_c$ to be explained is selected. Finally, the relevance $R_i^c$ of the individual input variables $i$ at the selected output $c$ is calculated in a backward pass.}
    \label{fig:feat_attr}
\end{figure}
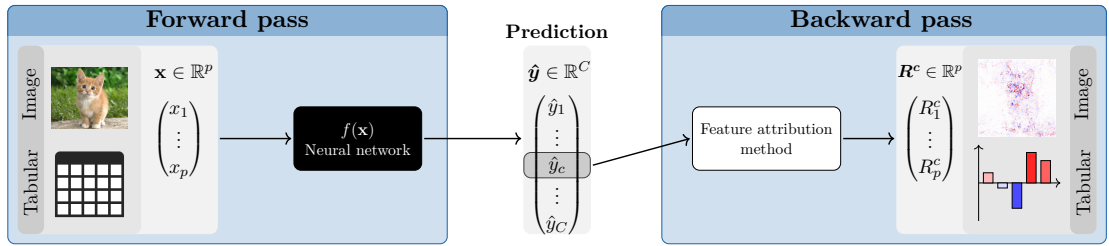

Despite the quantity and universality of model-agnostic methods, they are barely applicable to modern deep neural networks mainly because of two reasons: Firstly, many of these model-agnostic approaches are based on repeated evaluation of perturbed or permuted input instances, and secondly, they scale poorly for a higher number of input variables. Since the forward pass of deep neural networks is computationally intensive and the inputs are often high-dimensional RGB images, applying model-agnostic approaches is time-consuming and challenging. Moreover, global variable-wise methods are generally not appropriate for image data since the importance of pixels is rather independent of the exact localization and depends more on the neighboring pixels varying in each individual image. Alternatively, global model-specific approaches such as feature visualizations \citep{olah_2017} or concept-based methods \citep{kim_2018} can only be used with further optimization procedures or concept-labeled datasets. This gap of interpretability methods for neural networks is being filled by \textit{feature attribution} methods, which leverage all model-internal components in addition to the input and output relation, requiring only the model and the input instance to be explained. Particularly, this group of local model-specific methods, which assign relevance scores to each input variable for one of the output nodes or classes, prevailed and has successfully been applied in many domains \citep{zuallaert_2018, anders_2019, lauritsen_2020}. Furthermore, only a single regular forward pass and then a modified backward pass need to be performed for an explanation, i.e., they are generated without an optimization or estimation procedure. This two-step technique is illustrated in Figure~\ref{fig:feat_attr}: First, a prediction is produced in the standard forward pass, and the class to be explained is selected. In the subsequent method-specific backward pass, each input variable is assigned a relevance score to the chosen output class and can be visualized in a heat map or bar chart depending on the input type. An overview of the most popular feature attribution methods can be found in Section~\ref{sec:method}.

Several software packages have been developed in the last few years to make feature attribution methods widely accessible to users and to provide them with a unified and easy-to-use interface. The most popular packages are \pkg{innvestigate} \citep{innvestigate_2019} for the deep learning library \pkg{Keras} \citep{chollet2015}, and \pkg{captum} \citep{captum_2020} and \pkg{zennit} \citep{zennit_2021} for \pkg{PyTorch} \citep{pytorch}. In addition, the \pkg{shap} package \citep{Lundberg_2017} implements many Shapley-value-based methods and can handle both \pkg{Keras} and \pkg{PyTorch} models. However, all these packages are \proglang{Python}-exclusive and mostly only support networks of specific deep learning libraries. Despite the existing \proglang{R} packages for model-agnostic interpretability methods, such as \pkg{iml} and \pkg{dalex} \citep{iml, dalex}, we want to make feature attribution methods easily accessible to the \proglang{R} community and therefore provide the software package \pkg{innsight}, which pursues the following goals:
\begin{itemize}
    \item \textit{First feature attribution \proglang{R} package}: \pkg{innsight} is the first \proglang{R} package that implements the most popular feature attribution methods for neural networks unified in a single user-friendly package. 
    \item \textit{Computationally efficient}: The powerful \pkg{torch} \citep{torch_R} package is utilized internally for all calculations, which builds on \pkg{LibTorch}, the \proglang{C++} variant of \pkg{PyTorch} \citep{pytorch}, and doesn't rely on a \proglang{Python} dependency. 
    \item \textit{Deep-learning-library-agnostic}: The passed trained models are not limited to a specific deep learning library. The package supports models from the \proglang{R} packages \pkg{keras} \citep{keras_R}, \pkg{torch} and \pkg{neuralnet} \citep{neuralnet_2010}. However, under some constraints, an arbitrary model can be passed as a list to be fully flexible.
    \item \textit{Visualization tools}: \pkg{innsight} offers several visualization methods for individual or summarized results regardless of whether it is tabular, signal, image data, or a combination of these. Additionally, interactive plots can be created based on the \pkg{plotly} package \citep{plotly_R}.
\end{itemize}
The \pkg{innsight} package is available from the Comprehensive \proglang{R} Archive Network (CRAN) at \url{https://CRAN.R-project.org/package=innsight} or from our GitHub repository at \url{https://github.com/bips-hb/innsight/}. 

The rest of the paper is structured as follows: First, we overview the most popular feature attribution methods for neural networks in Section~\ref{sec:method}. Then, in Section~\ref{sec:usage}, we elaborate on the package's design, functionality, and capabilities. Next, the package is applied to a basic example on a penguin dataset and an advanced example for melanoma detection based on image and tabular data as input types. In the concluding Section~\ref{sec:validation}, the obtained package's outputs are compared and validated with the already mentioned \proglang{Python} equivalents.

\section{Methodology of feature attribution}\label{sec:method}

Feature attribution methods for neural networks describe a group of local interpretation methods that assign to each input variable the contribution or impact to a chosen model output. For example, suppose an input instance $\bm{x} \in \mathbb{R}^p$ with $p \in \mathbb{N}$ variables is fed forward through a neural network $f:\mathbb{R}^p \to \mathbb{R}^C$ resulting in an output $f(\bm{x}) = \bm{\hat{y}} \in \mathbb{R}^C$ with $C \in \mathbb{N}$ classes or regression outputs. In this case, a feature attribution method assigns relevance scores $R_1^c, \ldots, R_p^c$ to each of the input features $x_1, \ldots, x_p$ of $\bm{x}$ on a chosen output class or node $\hat{y}_c$ of the prediction $\bm{\hat{y}}$ to be explained, as already described in Figure~\ref{fig:feat_attr}.

\subsection{Gradient-based methods}\label{sec:gradient_based}

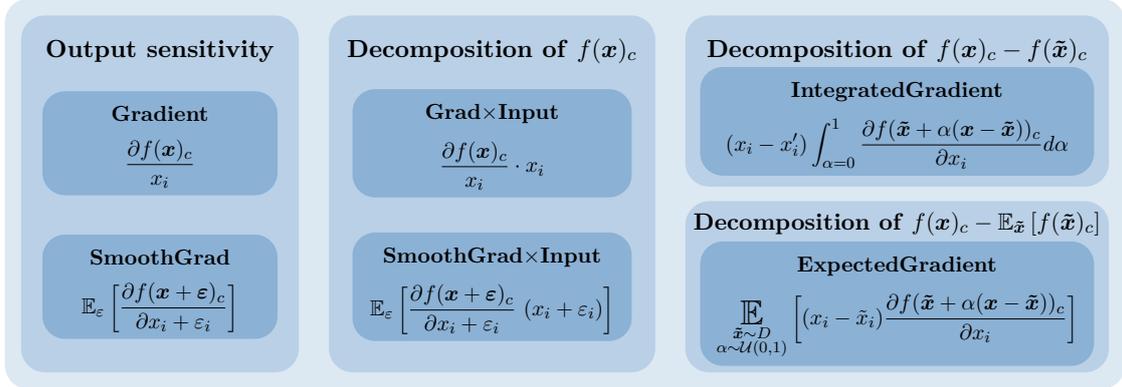
\begin{figure}
    \centering
    \resizebox{\linewidth}{!}{
        \input{tikz/gradientbased.tikz}
    }
    \caption{A summary of gradient-based feature attribution methods, including their mathematical representation. They are divided into blocks based on their underlying objectives. For example, in the case of feature-wise relevances $R_i^c$ obtained from Gradient$\times$Input, the goal is to achieve a sum that equals $f(\bm{x})_c$, i.e., $\sum_{i = 1}^p R_i^c = f(\bm{x})_c$.}
    \label{fig:overview_gradient_based}
\end{figure}

Gradient-based methods are the fastest and most straightforward interpretation methods because they operate on the default techniques of the high-level deep learning libraries for computing gradients during the training loop. However, these methods -- in a sense -- calculate the derivatives of chosen output to the input variables instead of the derivatives of the loss value to the model parameters during gradient descent. Although the terminology of gradient-based methods can often be interpreted more broadly, we only consider techniques that use the default gradient methods. For example, \cite{ancona_2018} showed that variants of the \textit{layer-wise relevance propagation (LRP)} and \textit{deep learning important features (DeepLift)}, discussed later in Section~\ref{sec:LRP} and \ref{sec:Deeplift}, can approximately be considered gradient-based. Regardless, they are not mentioned in this section, since not all LRP and DeepLift variants can be considered gradient-based. Additionally, these variants require an overwriting of the standard gradients, i.e., they don't use the mathematical definition of the gradient anymore. In the following, the most common gradient-based feature attribution methods are briefly explained, including their underlying objectives. For a more mathematical overview, see Figure~\ref{fig:overview_gradient_based}.

One of the first and most intuitive methods for interpreting neural networks is the \textit{Gradient} method introduced by \citet{simonyan_2014}, also known as \textit{Vanilla Gradients} or \textit{Saliency maps}. This method computes the gradients of the selected output with respect to the input variables. Therefore, the resulting relevance values indicate prediction-sensitive variables that can be locally perturbed the least to change the outcome the most. Assuming that the model $f$ behaves linearly overall, increasing $x_i$ by one raises the output by the calculated gradient. In general, neural networks are highly nonlinear, which forces the gradients to have large fluctuations or abrupt changes. This phenomenon can introduce noise and potential misinterpretations in the Gradient method. A simple extension of this basic Gradient method to tackle this issue is the \textit{smoothed gradients (SmoothGrad)} approach introduced by \cite{smilkov_2017}. They proposed computing the gradients of randomly Gaussian perturbed copies of $x_i$ and determining the average gradient from that, instead of calculating only the gradient in $x_i$. As a result, locally very noisy gradients are smoothed out and the method provides the average behavior in a larger neighborhood of $x_i$. The estimation accuracy and size of the neighborhood can be adjusted with the hyperparameters $n$ for the number of perturbations and $\sigma^2$ for the variance of the Gaussian noise. With the value of $n$, the estimation accuracy for the average gradient can be increased, but this goes hand in hand with a higher computational effort. The second parameter $\sigma^2$ is mostly specified indirectly via a noise level $\lambda \geq 0$ determining the proportion of the total range of the input domain that is covered by the standard deviation $\sigma$, i.e., $\lambda = \frac{\sigma}{x_\text{max} - x_\text{min}}$. Especially for images, this argument can be used to control the visual smoothness of the explanation.

A simple modification can change both previously discussed methods to the methods \linebreak \textit{Gradient$\times$Input} and \textit{SmoothGrad$\times$Input}. The gradients are calculated as for the respective methods and then multiplied by the corresponding feature values. The \textit{Gradient$\times$Input} method was introduced by \cite{shrikumar_2016} and relies on a well-grounded mathematical background despite its simple idea: The basic concept is decomposing the output prediction $\hat{y}_c$ according to its relevance to each input variable $x_i$, i.e., into variable-wise additive effects
\begin{align} \label{Eq_decompose}
    \hat{y}_c = f(\bm{x})_c = \sum_{i = 1}^p R_i^c.
\end{align}
Mathematically, this method is based on the first-order Taylor decomposition. Assuming that a function $g:\mathbb{R}^p \to \mathbb{R}$ is continuously differentiable in $\bm{x} \in \mathbb{R}^p$, a remainder term $\varepsilon(g,\bm{z},\bm{x}):\mathbb{R}^p \to \mathbb{R}$ with $\lim_{\bm{z} \to \bm{x}} \varepsilon(g, \bm{z}, \bm{x}) = 0$ exists such that
\begin{align*}
    g(\bm{z}) &= g(\bm{x}) + \nabla_{\bm{x}} g(\bm{x}) \cdot (\bm{z}-\bm{x})^\top + \varepsilon(g, \bm{z}, \bm{x})\\
         &= g(\bm{x}) + \sum_{i = 1}^p \frac{\partial\, g(\bm{x})}{\partial\, x_i} (z_i - x_i) + \varepsilon(g, \bm{z}, \bm{x}), \quad \bm{z}\in \mathbb{R}^p.
\end{align*}
The first-order Taylor formula thus describes a linear approximation of the function $g$ at the point $\bm{x}$ since only the first derivatives are considered. Consequently, a highly nonlinear and continuous function $g$ is well approximated only in a small neighborhood around $\bm{x}$. For larger distances from $\bm{x}$, sufficient small values of the residual term are not guaranteed anymore. The Gradient$\times$Input method considers the data point $\bm{x}$ and sets $\bm{z} = \bm{0}$. In addition, the residual term $\varepsilon(f_c, \bm{0},\bm{x})$ and the summand $f(\bm{0})_c$ are ignored. Analogously, this multiplication can be applied to all gradients in the summation of the SmoothGrad method in order to compensate for local fluctuations.

Even though the multiplication of gradients by the inputs provides an approximate decomposition of $f(\bm{x})_c - f(\bm{0})_c$, this approach only captures the feature-wise effects of $\bm{x}$ concerning a baseline of $\bm{0}$. However, this value does not necessarily reflect a prediction-neutral reference value and can be challenging to interpret or even lie outside the data distribution. \cite{sundararajan2017} proposed a method called \textit{IntegratedGradient} as a way to find a decomposition of $f(\bm{x})_c - f(\bm{\tilde{x}})_c$ into feature-wise effects for an arbitrary reference value $\bm{\tilde{x}}$, using integration over the Gradient$\times$Input values along an integration path. In practice, the integral is approximated by the sum of gradients multiplied by the inputs along an interpolated path from $\bm{x}$ to $\bm{\tilde{x}}$. Nevertheless, choosing the reference value $\bm{\tilde{x}}$ remains a challenging task and ideally requires domain-specific knowledge. The \textit{ExpectedGradient} method \citep{Lundberg_2017, erion2021} addresses this issue by estimating the mean value of the IntegratedGradient method through Monte Carlo integration, considering the whole distribution of reference values instead of a single baseline. In this sense, the method finds a feature-wise decomposition of $f(\bm{x})_c - \mathbb{E}_{\bm{\tilde{x}}} [f(\bm{\tilde{x}})_c]$ and, thus, calculates approximately Shapley values.

\subsection{Layer-wise relevance propagation (LRP)}\label{sec:LRP}

\begin{figure}[t]
    \centering
    \begin{subfigure}[t]{0.47\textwidth}
        \resizebox{\textwidth}{!}{
            \input{tikz/lrp.tikz}
        }
        \caption{Backward pass of the LRP method.}
        \label{fig:lrp_overview}
    \end{subfigure}
    \hfill
    \begin{subfigure}[t]{0.52\textwidth}
        \resizebox{\textwidth}{!}{
            \input{tikz/lrp_detail.tikz}
        }
        \caption{Calculation of the relevance of layer $l$ based on the relevances of the upper layer $l+1$.}
        \label{fig:lrp_detail}
    \end{subfigure}
    \caption{(a) illustrates the layer-by-layer backpropagation of relevances $R_i^l$ from the prediction score to the input variables through the use of relevance messages $r_{i \leftarrow j}$. For a hidden layer, (b) demonstrates how the relevance of the lower layer $l$ results from summing all incoming relevance messages.}
    \label{fig:enter-label}
\end{figure}
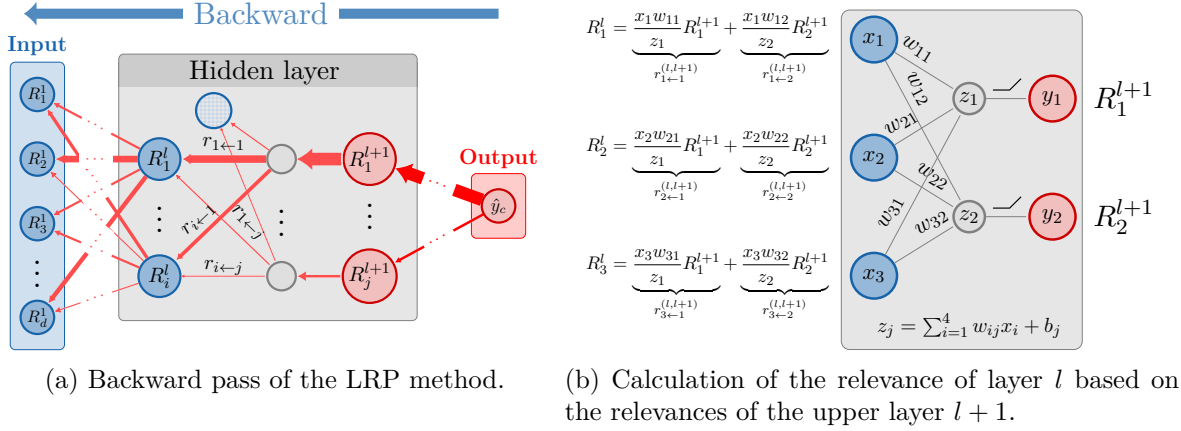

The \textit{layer-wise relevance propagation (LRP)} method was introduced by \cite{bach_2015} and has a similar goal as the Gradient$\times$Input approach explained in the previous section: decomposing the output into variable-wise relevances conforming to Equation~\ref{Eq_decompose}. The distinguishing aspect is that the prediction $\hat{y}_c$ is redistributed layer by layer from the output node back to the inputs according to the layer's weights and intermediate values. The entire procedure is accomplished by rule-based relevance messages defining how to redistribute the upper-layer relevance to the lower layer. A high-level overview of this method applied to a neural network with $L$ layers can be seen in Figure~\ref{fig:lrp_overview}. The method mainly consists of the following iterative steps: As the starting point, the relevance for the considered output node $R_1^L$ is set to the respective prediction score $\hat{y}_c$. Subsequently, the relevance of the lower layer's node $R_i^{L -1}$ is calculated using the sum of all incoming relevance messages. A relevance message describes the proportion of the upper-layer relevance $R_1^{L}$ that is sent to a node $i$ in the lower layer. This process is repeated layer by layer backwards, as shown in Figure~\ref{fig:lrp_overview}, until the input layer is reached and relevances are obtained for each input feature. More precisely, for a hidden layer $l$, the relevance message $r_{i \leftarrow j}^{(l, l + 1)}$ from node $j$ in the upper layer $l+1$ to node $i$ in the preceding layer defines the proportion of the relevance from $R_j^{l+1}$ attributed to the node $i$ in the lower layer. Since the relevance messages are based solely on the contribution from a single upper-layer node, the overall relevance of the node $i$ is obtained by summing up all incoming relevance messages (see Figure~\ref{fig:lrp_detail}), i.e.,
\begin{align} \label{eq:lrp_conservation}
R_i^{l} = \sum_j r_{i \leftarrow j}^{(l, l +1)}.
\end{align}

Since the publication of the LRP method, various variations of relevance messages flowing from the upper layer to the lower-layer node have emerged. However, the fundamental rule on which all other variations of relevance messages are more or less based is the \textit{simple rule} (also known as \textit{LRP-0}). The relevances are redistributed to the lower layers according to the ratio between local and global pre-activation. Despite being a rule-based approach, if the neural network only includes ReLU activations, this rule makes the method LRP equivalent to Gradient$\times$Input presented in Section~\ref{sec:gradient_based} \citep{ancona_2018}. Let $\bm{x}$ be the input, $\bm{w}$ the weight matrix and $\bm{b}$ the bias vector of layer $l$, and $R_j^{l+1}$ the upper-layer relevance; then $x_i\, w_{ij}$ is the local and $z_j = b_j + \sum_{k} x_k\, w_{kj}$ the global pre-activation defining the simple rule as (also used in Fig.~\ref{fig:lrp_detail})
\begin{align}\label{eq:lrp_simple}
    r_{i \leftarrow j}^{(l,\, l+1)} = \frac{x_i\, w_{ij}}{z_j} \, R_j^{l +1}.
\end{align}
Many other rules for relevance messages are built upon this principle. The well-known variations \textit{$\varepsilon$-rule} and \textit{$\alpha$-$\beta$-rule} of the simple rule and their advantages and disadvantages are explained in the Appendix~\ref{sec:app_LRP}. Additionally, a brief summary of other rules discussed in the literature is provided there.

\subsection{Deep learning important features (DeepLift)}\label{sec:Deeplift}

One method that, to some extent, echoes the idea of LRP is the so-called \textit{deep learning important features (DeepLift)} method introduced by \cite{shrikumar_2017}. It behaves similarly to LRP in a layer-by-layer backpropagation fashion from a selected output node back to the input variables, considering the simple rule. However, it incorporates a reference value $\bm{\tilde{x}}$ to compare the relevances with each other, analogously to the IntegratedGradient method discussed in Section~\ref{sec:gradient_based}. Hence, the relevances of DeepLift represent the relative effect of the outputs of the instance to be explained $f(\bm{x})_c$ and the output of the reference value $f(\bm{\tilde{x}})_c$. By taking the difference, the bias term is eliminated in the relevance messages, preventing the relevance absorption and leading to an exact variable-wise decomposition of the difference-from-reference output $\Delta \hat{y}_c = f(\bm{x})_c - f(\bm{\tilde{x}})_c$, i.e.,
\begin{align*}
    \Delta \hat{y}_c = f(\bm{x})_c - f(\bm{\tilde{x}})_c = \sum_{i=1}^p R_i^c.
\end{align*}
Similar to the relevance messages for LRP, DeepLift defines so-called \textit{multipliers} for each layer or part of a layer. Based on these multipliers, the contribution of an arbitrary input (or intermediate) variable to the difference-from-reference output can be obtained by multiplying it by the corresponding difference-from-reference value. For an arbitrary layer with the layer's input $\bm{x}$, reference input $\bm{\tilde{x}}$, and multiplier $m_{\Delta \bm{x} \Delta \hat{y}_c}$, this means:
\begin{align}\label{eq:DeepLift_multiplier}
    \sum_{i}  m_{\Delta x_i \Delta \hat{y}_c} \left( x_i - \tilde{x}_i \right) = m_{\Delta \bm{x} \Delta \hat{y}_c} \cdot \left(\Delta \bm{x}\right)^\top = \Delta \hat{y}_c.
\end{align}
The multipliers fulfill a chain rule allowing the computation of the multiplier for the preceding layer given the already calculated one $m_{\Delta \bm{t}, \Delta \hat{y}_c}$, i.e.,
\begin{align}\label{eq:DeepLift_chain_rule}
    m_{\Delta x_i \Delta \hat{y}_c} = \sum_{j} m_{\Delta x_i \Delta t_j}\, m_{\Delta t_j \Delta \hat{y}_c}.
\end{align}
In other words, the chain rule justifies defining the multipliers for each layer or part of a layer separately before combining them with the upper-layer multipliers. For linear components, such as the matrix multiplication in dense or convolutional layers, the weights are used as the multipliers, i.e., $m_{\Delta x_i \Delta z_j} = w_{ij}$. For nonlinear components, like, e.g., all point-wise activations such as ReLU, tanh, or sigmoid, \cite{shrikumar_2017} propose the \textit{Rescale} and \textit{RevealCancel} rule. While the Rescale rule uses the ratio of the layer's difference-from-reference output and difference-from-reference pre-activation as the multiplier, the RevealCancel rule is designed to propagate meaningful relevances even for saturated activations and discontinuous gradients through the layers' activation part. For a more detailed explanation, we refer to the Appendix~\ref{sec:app_Deeplift}.
These rules, along with the chain rule (Eq.~\ref{eq:DeepLift_multiplier}-\ref{eq:DeepLift_chain_rule}), enable the successive computation of the input variables' contributions $R_i^c$ to the difference-from-reference output $\Delta \hat{y}_c$ in a single backward pass. 

The reference value is the only crucial hyperparameter for the DeepLift method, apart from the rule for non-linearities. This choice depends significantly on the application and usually requires proficient domain-specific knowledge. Nevertheless, the authors suggest asking oneself the question of what one wants to measure an effect against. For example, taking the background color or blurred versions of the original picture as the reference values for images are reasonable choices. In many cases, zeros as a baseline are also used. \cite{ancona_2018} showed that using the Rescale rule with activations crossing the origin (i.e., $\sigma(0) = 0$) and a zero baseline as reference value $\bm{\tilde{x}}$ coincides with the Gradient$\times$Input method discussed in Section~\ref{sec:gradient_based} and with LRP with the simple rule. Similar to how the ExpectedGradient method generalizes IntegratedGradient considering the distribution of baseline values instead of a single reference value, the \textit{DeepSHAP} \citep{Lundberg_2017} method extends the DeepLift technique. It calculates the average DeepLift value across various baseline values, thereby achieving a decomposition of $f(\bm{x}) - \mathbb{E}_{\bm{\tilde{x}}}[f(\bm{\tilde{x}})]$ into feature-wise effects and, thus, gives approximately Shapley values.

\subsection{Connection weights}

One of the earliest methods specifically designed for neural networks is the \textit{connection weights (CW)} method invented by \cite{olden_2004}, resulting in a global relevance score for each input variable. The basic idea of this approach is to multiply all path weights for each possible connection between an input variable $x_i$ and the output node or class $\hat{y}_c$ and then calculate the sum of all of them. However, this method ignores all bias vectors and all activation functions during calculation. Analogously to the previous methods, CW can also be defined layer by layer, deriving the relevance for layer $l$ from the upper layer as follows:
\begin{align*}
    R_i^l = \sum_{j} w_{ij} R_j^{l+1}.
\end{align*}
Since only the model weights are used, this method is independent of input data and, thus, a global interpretation method. Inspired by the method Gradient$\times$Input, it can also be extended into a local method by taking the point-wise product of the global CW method and the input data. 

\subsection{Choice of the method}

Overall, the choice of methods for a user remains an open research question, but there are several recommendations that can be derived from the complexity or stated goals of the methods. Firstly, consider the computational efficiency of the method. Standard gradient methods such as Gradient and Gradient$\times$Input, as well as LRP, are very fast but can be very noisy as they only examine a very local behavior. On the other hand, methods like SmoothGrad, IntegratedGradient, and ExpectedGradient often require a high number of forward passes to deliver accurate results, but also consider the local neighborhood or baseline values. Furthermore, it has been demonstrated that many methods (such as Gradient$\times$Input, LRP, IntegratedGradient, DeepLift) are not invariant to constant shifts of the inputs \citep{kindermans2019reliability,haug2021baselines,nielsen2022robust}. This is because the Taylor approximation is only accurate around zero (or the reference value), and the choice of the reference value has a crucial impact. One solution to this is provided by Shapley-based methods like ExpectedGradient and DeepSHAP. However, these methods require a highly informative reference dataset in addition to the instance to be explained, and they are noticeably slower. These considerations highlight the trade-offs involved in selecting a feature attribution method for a given task, and it is crucial for users to weigh the speed, accuracy, and invariance characteristics based on the specific requirements of their application.

\section{Functionality and usage}\label{sec:usage}

\begin{wrapfigure}[14]{r}{0.39\textwidth}
    \centering
    \resizebox{0.3\textwidth}{3cm}{
        \input{tikz/innsight.tikz}
    }
    \captionof{figure}{\pkg{innsight} utilizes the package \pkg{torch}, which builds directly on the \proglang{C++} library \pkg{LibTorch} without a \proglang{Python} dependency.}
    \label{fig:innsight}
\end{wrapfigure}
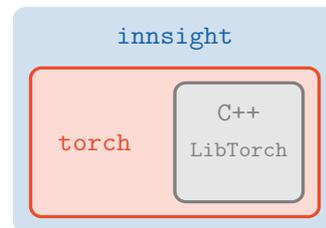

The \proglang{R} package \pkg{innsight} combines all the methods discussed in the previous section in a user-friendly structure and a unified step-based workflow from the trained model to the visualization of the relevances of a feature attribution method. For efficient high-dimensional array calculations, the package utilizes the \proglang{R} package \pkg{torch} \citep{torch_R}, which builds on \pkg{LibTorch} (the \proglang{C++} variant of \pkg{PyTorch} \citep{pytorch}), and consequently runs without a \proglang{Python} dependency (see Fig.~\ref{fig:innsight}). The following three steps yield the requested results, regardless of the class of the passed model or the chosen feature attribution method:
\begin{itemize}
    \item[-] Step 1: Convert the model
    \item[-] Step 2: Apply selected method
    \item[-] Step 3: Get or visualize results.
\end{itemize}
Internally, a class structure is being built using \code{R6} classes based on the equally named package \pkg{R6} \citep{r6_R}. This type of object-oriented programming class is used because \pkg{torch} also relies on it, and it simplifies inheritance and argument passing compared to conventional \code{S3} and \code{S4} classes. Apart from the utilized packages for the internal workflows, calculations, and visualizations discussed in the following sections, the packages \pkg{checkmate} \citep{checkmate_R} and \pkg{cli} \citep{cli_R} are generally used for all argument verifications, internal checks, and terminal outputs of messages, warnings, and errors.

For illustration and better comprehension, the three steps will be exemplified from a user's perspective using the \textit{bike sharing dataset} \citep{bike_sharing}. The internal mechanisms and more detailed descriptions are provided in the subsequent sections. This regression dataset contains information about the number of bicycles rented on a given day, along with various features such as weather conditions, holidays, and temperature. When using the \pkg{innsight} package, object-oriented \code{R6} objects are internally created at each step, but their initialization is simplified through function calls providing a conventional \proglang{R} usage without prior knowledge about \code{R6} classes. This facilitates easy application for \proglang{R} users of the package.

In the following code snippet, the dataset is loaded and restricted to a few variables. The outcome in this regression dataset is \code{"cnt"}, indicating the total number of bicycles rented on the given day. The model is trained on this dataset using the \pkg{neuralnet} package \citep{neuralnet_2010} containing one hidden layer with $64$ neurons. In addition, the outcome variable \code{"cnt"} is scaled to bikes per $10\,000$.
\begin{CodeInput}
R> library("neuralnet")
R> set.seed(42)
R> bike <- read.csv("additional_files/bike_sharing/day.csv")
R> bike <- bike[, c("cnt", "holiday", "workingday",
+    "temp", "hum", "windspeed")]
R> bike$cnt <- bike$cnt / 10000
R> bike <- as.matrix(bike)
R> model <- neuralnet(cnt~., data = bike, hidden = c(64), 
+    linear.output = TRUE)
\end{CodeInput}

To enable a deep-learning-library-agnostic implementation, the given model \code{model} is analyzed in the first step, and internally, a replica based on the \pkg{torch} package is reconstructed. However, for the user, this process is abstracted through the \code{convert()} function, allowing the adjustment of the used variable and outcome names, e.g., with the argument \code{output_names}:
\begin{CodeInput}
R> library("innsight")
R> conv <- convert(model, output_names = c("Number of rented bikes/10,000"))
\end{CodeInput}

In the second step, the user's method of choice can be applied to the provided data (argument \code{data}). Internally, an \code{R6} class for the respective method is initialized but hidden from the user again. In this case, the DeepSHAP method (\code{run_deepshap()}) is run on the first $20$ instances with the entire dataset as reference values. For computational reasons, the internal calculation uses $100$ samples from this dataset, as this is the default value of \code{limit_ref}. In addition, the data must always be passed as input data only, which is why the outcome variable \code{"cnt"} is removed in the following code:
\begin{CodeInput}
R> res_deepshap <- run_deepshap(conv, bike[1:20, -1], data_ref = bike[, -1])
\end{CodeInput}

\begin{figure}[!t]
    \centering
    \begin{subfigure}[t]{0.5\textwidth}
        \includegraphics[width = \textwidth]{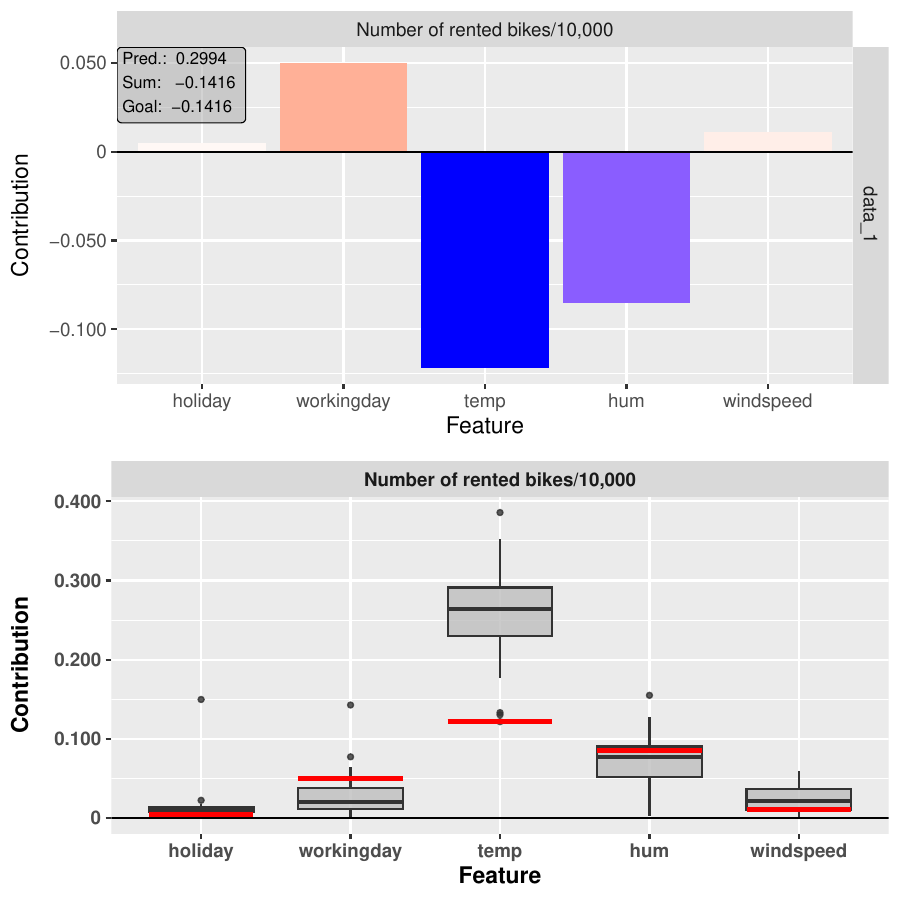}
        \caption{Visual results of \code{plot()} and \code{boxplot()} from the bike sharing example.}
        \label{fig:bike_example}
    \end{subfigure}%
    \hskip10pt
    \begin{subfigure}[t]{0.4\textwidth}
        \resizebox{\textwidth}{!}{
            \input{tikz/innsight_intern.tikz}
        }
        \caption{Internal conversion process of an \code{Converter} object.}
        \label{fig:innsight_intern}
    \end{subfigure}
    \caption{(a) displays the visualizations of the \code{plot()} and \code{boxplot()} functions applied to the DeepSHAP method on the bike sharing dataset. In (b), the internal conversion process of creating a new \code{Converter} object is shown, which is identical to calling the shortcut function \code{convert()}.}
\end{figure}

In the final step, results can be extracted, for instance, using the \code{get_result()} function, or visualized using \code{plot()} or \code{plot_global()}/\code{boxplot()}. The \code{boxplot()} method is an alias for \code{plot_global()} in case of tabular and signal data, as boxplots are created for these data types. It is noted that the variable \code{"hum"} for humidity is scaled between $0$ and $1$, and \code{"tmp"} and \code{"windspeed"} are divided by the respective maximal value, i.e., $100$ and $67$. The visualization relies on the \pkg{ggplot2} package and can be customized accordingly (the plots are shown in Fig~\ref{fig:bike_example}):
\begin{CodeInput}
R> library("ggplot2")
R> head(get_result(res_deepshap))
R> plot(res_deepshap)
R> boxplot(res_deepshap, ref_data_idx = 1) +
+    theme(text = element_text(face = "bold"))
\end{CodeInput}
\begin{CodeOutput}
, , Number of rented bikes/10,000

          holiday   workingday       temp         hum     windspeed
[1,]  0.004659358  0.050143935 -0.1218946 -0.08540524  0.0109126559
[2,]  0.007893519  0.064216673 -0.1305959 -0.01469872 -0.0320372544
[3,] -0.009826845  0.007002956 -0.2871930  0.09675989 -0.0359177366
[4,] -0.011643781  0.028260766 -0.2517902  0.04398110  0.0213681515
[5,] -0.009875727  0.015752951 -0.2339799  0.09297864 -0.0008396581
[6,]  0.002854239 -0.004088928 -0.2964459  0.05857125  0.0088869939
\end{CodeOutput}

For example, in the box plots (see Fig.~\ref{fig:bike_example} bottom), bold font is used for the labels, which is achieved through the \code{theme(text = element_text(face = "bold"))} feature of \pkg{ggplot2}. By default, the orientation of relevances is concealed by absolute values, as global attention is usually focused on the strength of the effects. Consequently, it can be observed that the model considers temperature as the most crucial feature for its predictions. Furthermore, the argument \code{ref_data_idx} is employed to highlight the relevances of the first instance of the dataset as a reference value with red lines. This instance and its local effects are more precisely visualized in the upper illustration of Figure~\ref{fig:bike_example}. It also shows that temperature is an influential factor for the model since it predicts a relatively low count of $2\,994$ bicycles (compared to the average of $4\,504$ bicycles). This aligns with the data, as on that particular day, the temperature is $8.18$°C (normalized value $0.344167$), while the dataset's average temperature is $15.28$°C (normalized value $0.49538$). Additionally, the plot includes a small box displaying the model's prediction of the instance ($0.2994$), the sum of calculated relevances ($-0.1416$), and the decomposition target of the method ($-0.1416$). For example, the DeepSHAP method is based on the decomposition of the difference between the prediction and the average prediction in the baseline dataset (i.e., $f(\bm{x})_c - \mathbb{E}_{\bm{\tilde{x}}}[f(\bm{\tilde{x}})_c]$), which is for the first instance $-0.1416$, i.e., $1\,416$ bikes below the average.

\subsection{Step 1 - Convert the model}\label{sec:step_1}

The key step that turns the \pkg{innsight} package into a deep-learning-library-agnostic approach and unlocks the provided \pkg{torch} toolbox to all methods is this first step, which essentially analyzes a passed model and creates a \pkg{torch}-based replication. For the user, however, the internal processes remain hidden, and the entire conversion step is accomplished by creating a new instance of the class \code{Converter}:
\begin{Code}
Converter$new(model,
  input_dim = NULL, input_names = NULL, output_names = NULL, 
  dtype = "float", save_model_as_list = FALSE)
\end{Code}
As previously mentioned, this object is implemented using the object-oriented \code{R6} class from the \pkg{R6} package \citep{r6_R}. To overcome prior knowledge of \code{R6} classes, the shortcut function \code{convert()} is implemented, which simply forwards all arguments to the \code{Converter}'s initialization function from above. 

The only necessary argument is the passed \code{model}, which can be either a \code{nn_sequential} object from \pkg{torch}, a \code{keras_model} or \code{keras_model_sequential} object from \pkg{keras} \citep{chollet2015}, a \code{neuralnet} object from \pkg{neuralnet} \citep{neuralnet_2010}, or a named list in a specific style. The other arguments \code{input_dim}, \code{input_names} and \code{output_names} are optional -- except \code{input_dim} in combination with \pkg{torch} models -- and are used for internal validation of the copied model or to assign labels to the input and output nodes used for the visualizations in Step 3 (see Section~\ref{sec:step_3}). This is already demonstrated in the previous example with the bike sharing dataset, where the output name \code{"Numbers of rented bikes/10,000"} is passed. In addition, the arguments \code{dtype} and \code{save_model_as_list} specify the calculations' numerical precision and save the entire model as a named list in the instance's field \code{model_as_list}, which is created as an intermediate step during the conversion process and are explained in more detail in the next paragraph and Figure~\ref{fig:innsight_intern}.

To be as flexible as possible and to interpret neural networks from almost any \proglang{R} package, a conversion method is implemented for each of the model classes mentioned above of the packages \pkg{torch}, \pkg{keras} and \pkg{neuralnet}, summarizing all decisive components and layers of the passed model in an ordered and unified way into a list. Then, a \pkg{torch}-based model \code{ConvertedModel} (i.e., a subclass of \code{nn_module}) is created internally from this list. In addition, the rule-based interpretation methods described in Section~\ref{sec:method} are pre-implemented for each valid layer type, which can be called layer by layer in the following Step 3. Since the creation of the converted model is consequently independent of the class of the given model, the conversion call can be bypassed by directly passing the desired model as a list. Hence, custom wrappers for other packages' models can be written, allowing an interpretation of models not being created by the packages \pkg{torch}, \pkg{keras}, or \pkg{neuralnet}. An overview of the individual steps that are performed internally when initializing a new instance of the \code{Converter} class is summarized in Figure~\ref{fig:innsight_intern}. In addition to the fields shown in Figure~\ref{fig:innsight_intern}, there are also fields containing the labels (\code{\$input_names}, \code{\$output_names}) and shapes (\code{\$input_dim}, \code{\$output_dim}) of the input and output layers in a unified list structure. What kind of list structure is required for a model passed as a list, which layers are generally accepted and even more is explained in detail in Appendix~\ref{app:model} or in the vignette "In-depth explanation" (see \code{vignette("detailed\_overview", package = "innsight")} and is only referred to at this point.

\subsection{Step 2 - Apply selected method}\label{sec:step_2}

As previously mentioned, the \pkg{innsight} package provides the most popular feature attribution techniques in a unified framework. For package users, simple functions are provided to apply the respective method to the data, as demonstrated at the beginning of this section with the bike sharing dataset using the \code{run_deepshap()} function. However, before delving into the individual user-facing functions, their internal class-related origin is explained, since the fundamental structure of the methods remains consistent. Internally, the unification is achieved by the \code{R6} super class \code{InterpretingMethod}, from which all methods intended for users inherit and only add method-specific arguments to those of the super class. The rudimentary call of initializing a new method object looks like this:
\begin{Code}
InterpretingMethod$new(converter, data,
  channels_first = TRUE, output_idx = NULL, output_label = NULL, 
  ignore_last_act = TRUE, verbose = interactive(), dtype = "float")
\end{Code}
The key arguments for every method are the \code{converter} object from the first step (see Sec.~\ref{sec:step_1}), containing the \pkg{torch}-converted model, and the \code{data} to be interpreted. The data can be passed in any format as long as the \proglang{R} base method \code{as.array()} can convert it into an array, and it matches the expected input dimension of the model. In addition, it is common for image or signal data to place either the channel axis directly after the batch axis or at the last position. However, this placement can generally not be extracted unambiguously from the data, which is why the \code{channels\_first} argument specifies where the channel axis is located, allowing the use of both formats. The remaining arguments \code{output\_idx}/\code{output_label}, \code{ignore\_last\_act}, \code{verbose} and \code{dtype} set which output nodes or labels are to be explained, whether the last activation function is ignored, whether a progress bar is displayed, or change the numerical precision for the calculations.

The feature attribution techniques designed for the package user's regular use cases and applications are inheritors of the super class \code{InterpretingMethod} and extend it by method-specific arguments. Since each method is implemented as an \code{R6} class, its application involves initializing a new class object through the \code{$new()} call. Therefore, helper functions are implemented, such as \code{run_deepshap(...)} in the example from the section's beginning, serving as a more user-friendly alternative to \code{DeepSHAP$new(...)}. For clarity, the subsequent presentation of the methods from Section~\ref{sec:method} focuses solely on these shortcut functions, although it is noted which \code{R6} class they initialize:
\begin{itemize}
    \item The methods \textit{Gradient} and \textit{Gradient$\times$Input} are implemented as the \code{R6} class \code{Gradient}, which has \code{times_input} as the only additional argument apart from the inherited ones. This argument switches between the usual gradients (\code{times_input = FALSE}) and the gradients multiplied by the corresponding inputs (\code{times_input = TRUE}):
    \begin{Code}
run_grad(converter, data, times_input = FALSE, ...)
    \end{Code}
    
    \item Similarly, the methods \textit{SmoothGrad} and \textit{SmoothGrad$\times$Input} are realized in the \code{R6} class \code{SmoothGrad} containing the arguments \code{n} for the number of perturbations and \code{noise\_level} for the noise scale in addition to the \code{times\_input} argument:
    \begin{Code}
run_smoothgrad(converter, data, times_input = FALSE, n = 50, 
  noise_level = 0.1, ...)
    \end{Code}

    \item The method \textit{IntegratedGradient} is implemented in the \code{R6} class \code{IntegratedGradient}. The method's baseline value can be specified using the argument \code{x_ref} defaulting to a zero baseline, and the number of discretization points for the integral approximation can be set with \code{n}:
    \begin{Code}
run_intgrad(converter, data, x_ref = NULL, n = 50, ...)
    \end{Code}

    \item Similarly, the \textit{ExpectedGradient} method can be applied by initializing an object of the class \code{ExpectedGradient}. The reference dataset can be passed with the \code{data_ref} argument defaulting to a zero baseline, and the number of samples with \code{n}:
    \begin{Code}
run_expgrad(convert, data, data_ref = NULL, n = 50, ...)
    \end{Code}
    
    \item The \textit{LRP} method, including the simple rule (\code{"simple"}), $\varepsilon$-rule (\code{"epsilon"}), $\alpha$-$\beta$-rule (\code{"alpha_beta"}), and a composition of these rules, is implemented in the \code{R6} class \code{LRP}. The rule and its corresponding parameter (if available) are set with the arguments \code{rule\_name} and \code{rule\_param}. The default value and meaning of \code{rule_param} depends on the selected rule, more precisely, for \code{"epsilon"} the rule's $\varepsilon$ value is set to $0.01$ and for \code{"alpha_beta"} a value of $\alpha = 0.5$ (i.e., $\beta = 1 - \alpha$) is used. For both arguments, named lists can also be passed to assign a rule or parameter to each layer type separately. Since many zeros are produced in a maximum pooling layer during the backward pass due to the selection of the maximum value in the pooling kernel, the argument \code{winner\_takes\_all} can be used to treat a maximum as an average pooling layer in the backward pass instead.
    \begin{Code}
run_lrp(converter, data, rule_name = "simple", rule_param = NULL, 
  winner_takes_all = TRUE, ...)
    \end{Code}
    
    \item Analogously, the method \textit{DeepLift} is realized in the \code{R6} class \code{DeepLift} including the argument \code{rule\_name} for selecting the \textit{Rescale} (\code{"rescale"}) or \textit{RevealCancel}\linebreak 
    (\code{"reveal_cancel"}) rule for non-linearities. The reference value is set with \code{x\_ref} defaulting to a baseline of zeros. DeepLift can also run into problems in maximum pooling layers since the maximum values in the pooling kernel from the normal and reference input generally do not coincide. Hence, with the \code{winner\_takes\_all} argument, this layer type can be treated as an average pooling layer in a backward pass.
    \begin{Code}
run_deeplift(converter, data, rule_name = "rescale", x_ref = NULL, 
  winner_takes_all = TRUE, ...)
    \end{Code}

    \item In the same way, the \textit{DeepSHAP} method is implemented as an \code{R6} class named \code{DeepSHAP}. Instead of a single reference value, the entire reference dataset is passed with the \code{data_ref} argument. For computational reasons, by default, a maximum of $100$ baseline values is considered for calculation, which can be adjusted using the \code{limit_ref} argument:
    \begin{Code}
run_deepshap(converter, data, data_ref = NULL, limit_ref = 100, ...)
    \end{Code}
    
    \item The last method provided by \pkg{innsight} is the \textit{connection weights (CW)} method realized in the \code{R6} class \code{ConnectionWeights}. The argument \code{times\_input} specifies whether the global result of the CW method is calculated or whether it is additionally multiplied by the inputs to obtain local instance-wise explanations. A notable aspect, in this case, is that the \code{data} argument is not needed for the global variant, but it is required for the local one.
    \begin{Code}
run_cw(converter, times_input = FALSE, ...) # global
run_cw(converter, data, times_input = TRUE, ...) # local
    \end{Code}
\end{itemize}

Although \pkg{innsight} primarily focuses on feature attribution methods specifically designed for neural networks, it also includes two well-known model-agnostic approaches, \textit{local interpretable model-agnostic explanation (LIME)} \citep{Ribeiro_2017} and \textit{Shapley values} \citep{Lundberg_2017}. LIME locally fits an intrinsic surrogate model (e.g., a generalized linear model) on the original model prediction to explain individual instances, while the game-theoretical Shapley values attribute the contributions of each feature by considering all possible feature combinations and their impact on the model output. Internally, they utilize the suggested packages \pkg{lime} \citep{lime} and \pkg{fastshap} \citep{fastshap}, and are incorporated into the class structure based on the \code{R6} class \code{InterpretingMethod}. They are realized in the \code{R6} classes \code{LIME} and \code{SHAP} and can be applied as follows:
\begin{Code}
run_lime(converter, data, data_ref, pred_fun = NULL, ...)
run_shap(converter, data, data_ref, pred_fun = NULL, ...)
\end{Code}
Since these methods are model-agnostic, any other predictive model can be passed instead of a \code{Converter} object in the argument \code{converter}. However, for this, the prediction function \code{pred_fun} must be specified so that \pkg{innsight} knows how to make predictions. This function is already pre-implemented for \code{Converter} objects and for models from the packages \pkg{neuralnet}, \pkg{keras} and \pkg{torch}. Additionally, both methods require a reference dataset, which is passed with \code{data_ref}. Inheriting from \code{InterpretingMethod} are only the arguments \code{channels_first}, \code{output_idx}, and \code{output_label}. Similarly to the \code{Converter} class, \code{input_dim}, \code{input_names}, and \code{output_names} can be passed. All other arguments are forwarded to the corresponding methods \code{lime::explain()} or \code{fastshap::explain()}, which are called internally.

\subsection{Step 3 - Get and visualize the results}\label{sec:step_3}

After creating an object of a selected method, in the third step the results can be extracted or, if required, presented in a descriptive and visual way. For this purpose, the \pkg{innsight} package provides three generic methods \code{get_result()}, \code{plot()} and \code{plot_global()} that either return the results as an \proglang{R} object (such as \code{array}, \code{torch_tensor} or \code{data.frame}) or create visualizations for individual instances or aggregated results over the whole passed dataset. All three generic functions call the respective class methods in the \code{InterpretingMethod} super class, which are inherited by all the interpreting methods from the second step by design. For instance, in the bike sharing example from the section's beginning, the plot is generated using \code{plot(res_deepshap)} instead of \code{res_deepshap$plot()}.

\subsubsection[Generic function getresult()]{Generic function \code{get\_result()}} \label{sec:get_result}

The function \code{get_result()} can be used to obtain the results in various forms, whatever is favored according to the user's subsequent workflow or application. This method has only the argument \code{type} (besides the method object), which determines the representation of the returned results. By default (\code{type = "array"}), the result is returned as an \proglang{R} base \code{array}, including the input and output names in the corresponding dimensions specified in the first step in the \code{Converter} object (see Sec.~\ref{sec:step_1}). The shape of the array is composed of the input shape including the batch size and the number of computed output nodes, i.e., for a tabular input with ten instances and four input variables, the shape is $10$$\times$$4$$\times$$3$ if the method was applied to three output nodes in Step 2. In the example with the bike sharing dataset, an array of size $20$$\times$$5$$\times$$1$ was returned using \code{get_result(res_deepshap)} which also includes the class label \code{"Number of rented bikes/10,000"}. This is because $20$ instances were explained, and the model has $5$ features and one output node. In the same way, \code{type = "torch_tensor"} returns a \code{torch_tensor} object having the same shape as the array, but without dimension labels. However, both variants can also return a list or list of lists with the related results as an \code{array} or \code{torch_tensor} for models with multiple input or output layers. The third and last format of the results is an \proglang{R} base \code{data.frame} obtained with \code{type = "data.frame"}. Included are columns for the input instance (\code{"data"}), the input and output layer of the model (\code{"model_input"} and \code{"model_output"}), the input variable (\code{"feature"}) -- possibly also a second one for images (\code{"feature_2"}) and the channel for signal and image data (\code{"channel"}) -- the output node or class (\code{"output_node"}), and the relevance (\code{"value"}) for the corresponding values. Moreover, the generated \code{data.frame} contains variables that show the prediction of the instance or already aggregated relevances for the respective output node and instance. The column \code{decomp_goal} indicates the decomposition goal of the method aimed by the aggregated relevances explained in Section~\ref{sec:method}, e.g., $f(\bm{x})_c - \mathbb{E}_{\bm{\tilde{x}}}[f(\bm{\tilde{x}})_c]$ for the DeepSHAP method.

\subsubsection[Generic function plot()]{Generic function \code{plot()}}

The generic function \code{plot()} visualizes individual instances of the result of the method applied before based on the graphic package \pkg{ggplot2} \citep{ggplot2_R} or the package \pkg{plotly} \citep{plotly_R} for interactive graphics if the corresponding argument \code{as_plotly} is set. The call for a method's object \code{method} is executed as follows:
\begin{Code}
plot(method, data_idx = 1, output_idx = NULL, output_label = NULL,
  aggr_channels = "sum", as_plotly = FALSE, same_scale = FALSE, 
  show_preds = TRUE)
\end{Code}
The key arguments are \code{data_idx} and \code{output_idx}/\code{output_label}, which specify the indices for the dataset instances and the indices/labels for the desired output nodes or classes whose result is to be visualized. By default, the first data instance and the first computed output node are used. In the argument \code{output_idx}, no arbitrary indices can be passed, but only those for which the results were calculated previously in the second step. The same applies to \code{output_label}, which must additionally be a subset of the output names \code{output_names} in the \code{Converter} object. The further argument \code{aggr_channels} can be used to define how the channels are aggregated for image and signal data. There are various options for choosing this aggregation function, which may significantly influence the quality of the explanation \citep{pooling}. Nevertheless, for \pkg{innsight}, the default is to compute the sum over the channels, which aligns with the additivity axiom for Shapley values to accurately reflect the collective group effect \citep{strumbelj2010}. Since visualization depends on the data type, it is internally distinguished between tabular/signal data and image data; accordingly, a bar chart or a raster chart is created, as shown in Figure~\ref{fig:viz_plot} on the left. The relevances in the bars or the pixels are also scaled by color, facilitating a visual comparison; red means positive, blue negative, and white the absence of relevance. Since, in general, the scales vary significantly for the selected output class or data instance, the plots are scaled separately for each value in \code{output_idx}/\code{output_label} and \code{data_idx}. When several input layers are to be visualized, the remaining argument \code{same_scale} can be used to select whether the individual input layers are also scaled separately in terms of color. This decision depends on the use case, as illustrated in the melanoma example in Section~\ref{sec:illu_melanom}. Additionally, in each plot, a small box displaying information for the respective instance and output node is presented. This includes the prediction, the sum of relevances, and, if available, the method's decomposition target of the sum of relevance. The appearance of the box can be toggled with the \code{show_preds} argument. For the bike rental example, Figure~\ref{fig:bike_example} top shows the visualized relevances of the first instance and the box with the additional predictive and decomposition information. Furthermore, instead of returning objects from the \pkg{ggplot2} or \pkg{plotly} packages, instances of the \code{S4} classes \code{innsight_ggplot2} and \code{innsight_plotly} are produced, which are explained in the Appendix~\ref{sec:app_usage_advanced} for advanced visualizations.

\subsubsection[Generic function plotglobal()]{Generic function \code{plot\_global()}}

Global behavioral patterns and insights into the model's decision-making process can be derived from the results of multiple instances by appropriately summarizing and aggregating them. The generic function \code{plot_global()} visualizes these global interpretations over the whole or parts of the given dataset based on the graphics package \pkg{ggplot2} \citep{ggplot2_R} or the package \pkg{plotly} \citep{plotly_R} for interactive charts analogous to the previously discussed function \code{plot()}. Box plots are created for the features of tabular and signal data, and the median pixels' relevance is shown for image data due to the high dimensionality. For the former, the alias function \code{boxplot()} can also be used analogously. The call for a method's object \code{method} is the following:
\begin{figure}[!t]
    \centering
    \resizebox{\textwidth}{!}{
	\input{tikz/visualization_tools.tikz}
	}
    \caption{Overview of the visualization tools \code{plot()} and \code{plot\_global()} provided by the \pkg{innsight} package depending on the type of input and the argument \code{as\_plotly}. The function \code{boxplot()} is an alias for \code{plot\_global()} in case of tabular or signal data.}
    \label{fig:viz_plot}
\end{figure}
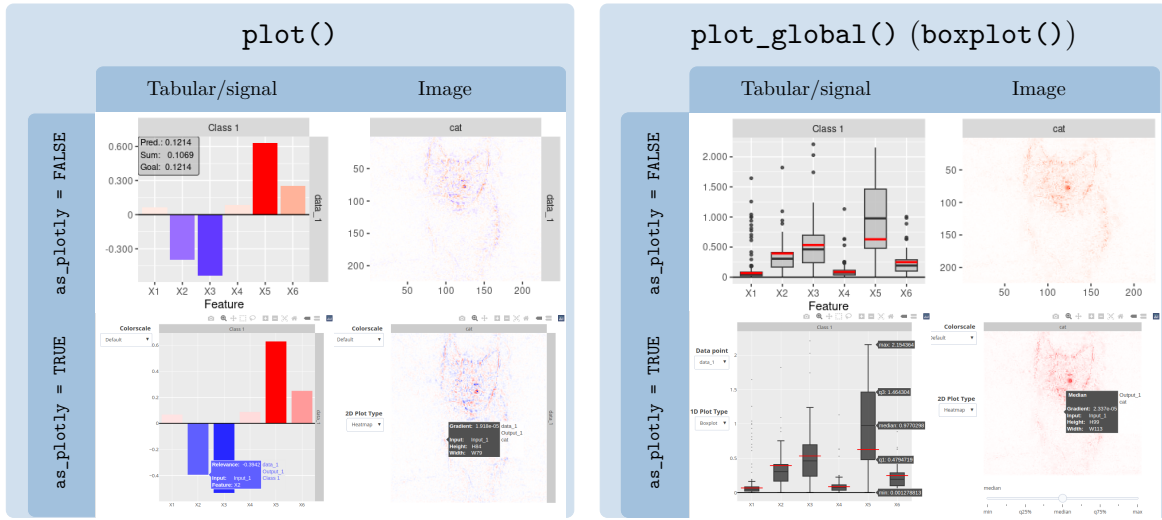
\begin{Code}
plot_global(method, output_idx = NULL, output_label = NULL, data_idx = "all", 
  ref_data_idx = NULL, aggr_channels = "sum", preprocess_FUN = abs, 
  as_plotly = FALSE, same_scale = FALSE, ...)
# For tabular and signal data
boxplot(method, ...)
\end{Code}
In addition to the identical arguments \code{output_idx}/\code{output_label}, \code{aggr_channels},\linebreak \code{as_plotly}, and \code{same_scale} for the \code{plot()} function, options for selecting the data points to be aggregated (\code{data_idx}), for drawing a reference data point (\code{ref_data_idx}) and a pre-process function of the results (\code{preprocess_FUN}) are added. By default, the absolute values of the relevances are calculated, as in global contexts, the focus is usually more on the intensity of effects rather than their orientation. However, this can be adjusted with the \code{preprocess_FUN} argument, e.g., showing the orientation with \code{identity}. Analogous to \code{plot()}, the visualization style depends on the type of input data; tabular and signal data are displayed as box plots, whereas only a raster plot with the pixel-wise median is rendered for image data due to the high dimensionality. For example, a box plot is created in the bike sharing example since it contains tabular data (see Fig.~\ref{fig:bike_example} bottom). In this visualization, the red reference lines are drawn for the first dataset instance using the argument \code{ref_data_ix}. However, if the chart is \pkg{plotly}-based, there is a slider to select which quantile to display. Basic examples and an overview of the \code{plot_global()} function are given in Figure~\ref{fig:viz_plot} on the right. Despite the creation of \pkg{ggplot2} or \pkg{plotly} graphs, instances of the \code{S4} class \code{innsight_ggplot2} or \code{innsight_plotly} are returned, which are explained in the Appendix~\ref{sec:app_usage_advanced}.

\section{Illustrations} \label{sec:illustrations}

To exemplify the methods and step-by-step execution of the \pkg{innsight} package, a standard dataset with only numerical tabular inputs on a simple model and a more complex dataset with image and tabular data on an extensive non-sequential network are analyzed in the following. The penguin dataset from the \pkg{palmerpenguins} package \citep{palmer_R} is used as the simple dataset, taking only the numerical variables of bill length and depth, flipper length, and body weight as inputs. The melanoma dataset \citep{rotemberg_2020} of the Kaggle competition\footnote{See the following link for the official dataset description \url{https://www.kaggle.com/competitions/siim-isic-melanoma-classification/overview/description}} is taken as the second dataset, which classifies the malignancy or benignity of the skin cell based on images of skin lesions and moles, and patient-level contextual information. Both datasets are classification tasks. Even though the bike sharing dataset has already exemplified a regression problem, feature attribution methods typically treat classification problems similarly to regression problems: The activation of the last layer -- usually sigmoid or softmax functions -- is ignored, and the pre-activation score is explained instead of the actual probability $\hat{y}_c$ \citep{shrikumar_2017,montavon_2019}.

\subsection{Example 1: Penguin dataset}

In the first example, the penguin dataset provided by the \pkg{palmerpenguins} package \citep{palmer_R} is used, and a neural network consisting of a dense layer is trained using the \pkg{neuralnet} package \citep{neuralnet_2010}. Before the \pkg{innsight} package can be applied, the dataset must be processed, and the neural network must be trained on the modified dataset. As a first pre-processing step, only the variables with the species, bill length and depth, flipper length, and body weight are selected, cleaned of missing values, and numerical variables are normalized:
\begin{CodeChunk}
\begin{CodeInput}
R> library("palmerpenguins")
R> set.seed(42)
R> data <- na.omit(penguins[, c(1, 3, 4, 5, 6)])
R> data[, 2:5] <- scale(data[, 2:5])
\end{CodeInput}
\end{CodeChunk}
Next, the dataset is divided into training data and test data at a ratio of $75\%$ to $25\%$:
\begin{CodeChunk}
\begin{CodeInput}
R> train_idx <- sample.int(nrow(data), as.integer(nrow(data) * 0.75))
R> train_data <- data[train_idx, ]
R> test_data <- data[-train_idx, -1]
\end{CodeInput}
\end{CodeChunk}
In the second pre-processing step, a network with $128$ units in a single hidden layer and the logistic function as activation is fitted on the training data \code{train_data}:
\begin{CodeChunk}
\begin{CodeInput}
R> library("neuralnet")
R> model <- neuralnet(species ~ ., data = train_data, hidden = 128, 
+    act.fct = "logistic", err.fct = "ce", linear.output = FALSE)
\end{CodeInput}
\end{CodeChunk}
Now, we follow the three steps that provide and visualize an explanation of the model \code{model} on the test data \code{test_data}, which were described in detail in Section~\ref{sec:usage}. As a reminder, the first step uses the \code{convert()} -- a shortcut function for initializing an object of the \code{R6} class \code{Converter} -- to convert the given \code{model} to a \pkg{torch}-based model:
\begin{CodeChunk}
\begin{CodeInput}
R> library("innsight")
R> conv <- convert(model)
\end{CodeInput}
\end{CodeChunk}
Then, in the second step, the desired method is selected and applied to the test data \code{test_data} via the corresponding function \code{run_*} which is identical to initializing the respective \code{R6} class. In this example, the IntegratedGradient method is applied with the average feature value as a baseline:
\begin{CodeChunk}
\begin{CodeInput}
R> intgrad <- run_intgrad(conv, test_data, 
+    x_ref = matrix(colMeans(test_data), 1))
\end{CodeInput}
\end{CodeChunk}
\begin{figure}[!t]
    \centering
    \begin{subfigure}[b]{0.48\textwidth}
        \resizebox{\linewidth}{!}{
            \includegraphics{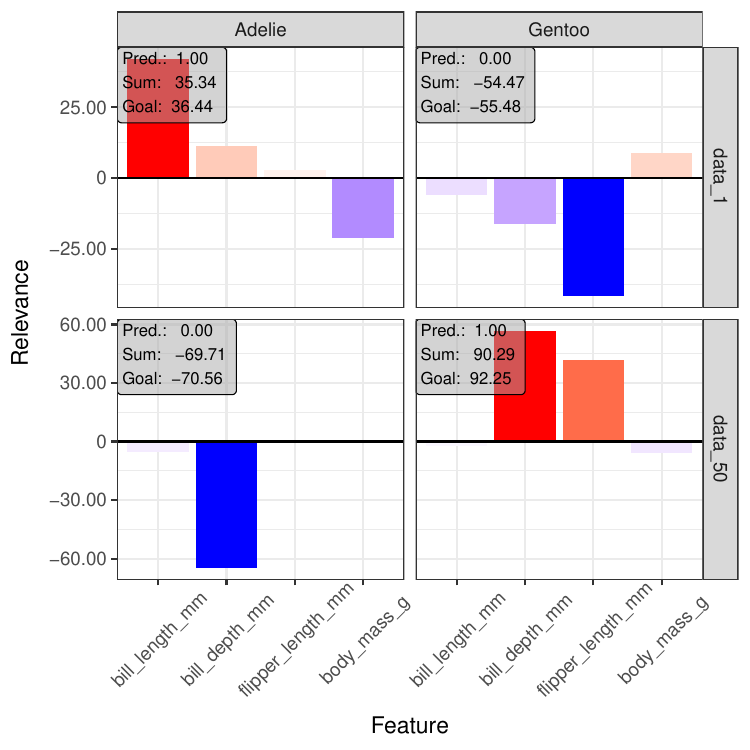}
        }
        \caption{Visualization created with \code{plot()}}
        \label{fig:illustr_1_a}
    \end{subfigure}%
    \hfill
    \begin{subfigure}[b]{0.48\textwidth}
        \resizebox{\linewidth}{!}{
            \includegraphics{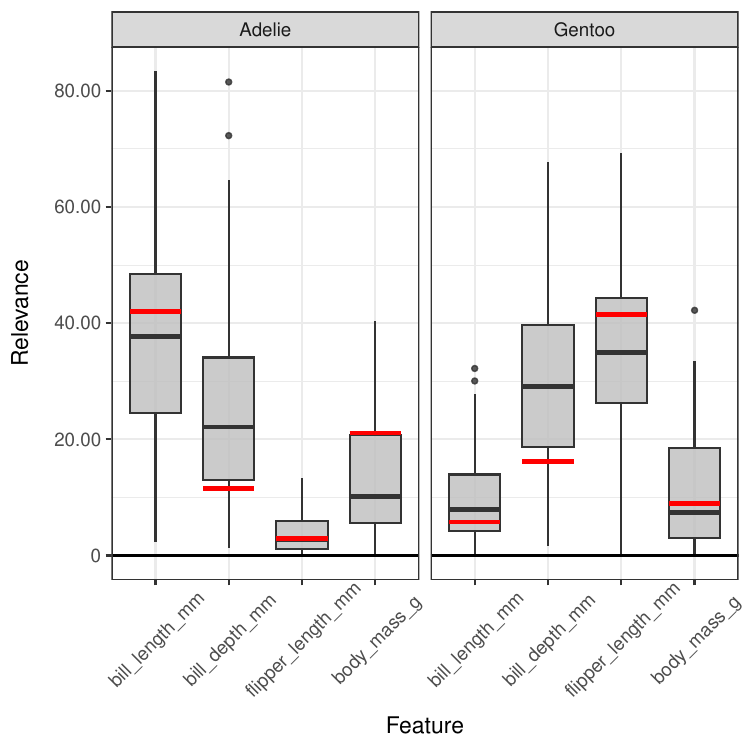}
        }
        \caption{Visualization created with \code{boxplot()}}
        \label{fig:illustr_1_b}
    \end{subfigure}
    \caption{Generated visualizations of IntegratedGradient results with average feature value as a baseline on the penguin dataset. Sub-figure (a) shows the individual results from data points $1$ and $50$ from the test data \code{test\_data} for the Adelie and Gentoo classes. In contrast, the summarized results as box plots across the whole test data for the same two classes can be found in (b), including the individual result of the first data point with the red line.}
    \label{fig:illustr_1}
\end{figure}
In the last step, the results are visualized in two ways: Using the \code{plot()} function, the relevances of one instance of the species Adelie (data index $1$) and one of the species Gentoo (data index $50$) are displayed for both corresponding classes. Secondly, the results for the two classes, Adelie and Gentoo, are aggregated over the entire test data, and box plots are generated using the \code{boxplot()} function, showing the first penguin as a reference. As mentioned in Section~\ref{sec:usage}, these two variants can be treated and modified like ordinary \pkg{ggplot2} objects, e.g., adding themes or rotating the x-axis labels. Both visualizations are executed by the following code and can be viewed in Figure~\ref{fig:illustr_1}:
\begin{CodeChunk}
\begin{CodeInput}
R> library("ggplot2")
R> plot(intgrad, data_idx = c(1, 50), output_label = c("Adelie", "Gentoo")) +
+   theme_bw() +
+   theme(axis.text.x = element_text(angle = 45, vjust = 0.6))
R> boxplot(intgrad, output_label = c("Adelie", "Gentoo"), ref_data_idx = 1) +
+   theme_bw() +
+   theme(axis.text.x = element_text(angle = 45, vjust = 0.6))
\end{CodeInput}
\end{CodeChunk}
In Figure~\ref{fig:illustr_1_a}, it can be seen that the bill length for the chosen penguin of the Adelie class (index $1$ in the dataset \code{test_data}) is highly relevant -- based on the trained model -- for this particular class aligning to the prediction of $100\%$ which is shown in the info box. At the same time, the penguin's flipper length argues against the Gentoo class due to its strong negative relevance. For the Gentoo penguin, the bottom row in Figure~\ref{fig:illustr_1_a} reveals that the bill depth is decisively in favor of the Gentoo class and concurrently against the Adelie species. The respective info boxes show that the model is generally very confident in its prediction and that the method has achieved its decomposition goal very well. As mentioned at the beginning of the section, the pre-activations are decomposed by default for classification problems and not the probability scores. Besides these instance-wise explanations, the \code{boxplot()} function provides aggregate insights across the entire test data \code{test_data}, summarized in Figure~\ref{fig:illustr_1_b}. The box plots show that, indeed, the bill length and depth are the most crucial variables for the Adelie class and consequently strongly influence it. Simultaneously, however, the length of the bill is not as decisive for the Gentoo class, but the bill depth is. It further emerges that the flipper length is another crucial feature for the Gentoo class. The red lines for the first penguin also show that the bill depth is not as relevant as it is on a global scale.

\subsection{Example 2: Melanoma dataset}\label{sec:illu_melanom}

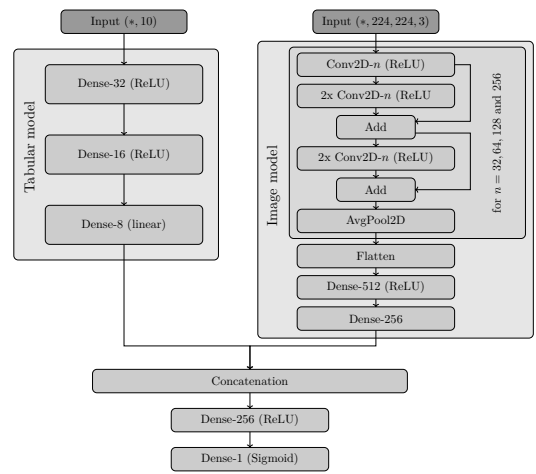
\begin{wrapfigure}{r}{0.46\textwidth}
    \centering
    \resizebox{0.45\textwidth}{!}{
        \input{tikz/melanoma_model.tikz}
    }
    \captionof{figure}{Model architecture for the melanoma dataset.}
    \label{fig:melanoma_model}
\end{wrapfigure}

The second example examines the melanoma dataset \citep{rotemberg_2020} from the Kaggle challenge\footnote{See the following link for the official dataset description \url{https://www.kaggle.com/competitions/siim-isic-melanoma-classification/overview/description}} in 2020, issued by the society of imaging informatics in medicine (SIIM) and based on the international skin imaging collaboration (ISIC) archive, the most extensive publicly available collection of quality-controlled dermoscopic images of skin lesions. This dataset consists of $33\,126$ labeled images with associated patient-level contextual information, such as the age, gender, and image location of the skin lesion or mole. 

Due to the complexity and high dimensionality of the data, training a neural network is not straightforward and overall not the main focus of this paper thus, reference is made to the GitHub repository for reproduction (\url{https://github.com/bips-hb/JSS_innsight/}), and only the most notable points are summarized in the following: The tabular input part's numerical and one-hot encoded categorical variables are fed into a sequential model of dense layers. On the other hand, an architecture based on the established residual layers \citep{he_2016} considering skip connections between convolutional layers is used for the image data. Afterward, the two outputs of the respective input parts are merged by concatenation and finally flow in a sequential model with only dense layers to obtain a prediction probability for the skin lesion status. The coarse structure is summarized in Figure~\ref{fig:melanoma_model}, where additional dropout layers are used between dense layers. Furthermore, the numerical variable age and the one-hot encoded variables gender and location yield ten features as inputs for the tabular model, and the images are resized to $224\!\times\!224\!\times\!3$ for the image model. This model architecture is trained on the melanoma dataset with a validation split of $20\%$ and a batch size of $256$ instances using the \pkg{Keras} library \citep{chollet2015} with stochastic gradient descent as the optimizer and class-weighted binary cross-entropy as the loss function. The best model is selected based on the highest value of the area under the ROC curve (AUC) on the validation data. This metric is chosen because the dataset is highly imbalanced with only $584$ of the $33\,126$ images containing a malignant skin lesion. Since the model is trained from scratch and the image model has significantly more parameters than the tabular one, training starts with $300$ warm-up epochs on the image model using the image data only. Then, the image model is joined with the tabular and the dense output model. Afterward, training continues on the image and tabular data, saving the model with the highest value of the AUC metric on the validation data. In addition, the initial learning rate of $0.01$ is reduced by a factor of $0.1$ after $20$ epochs without a validation AUC improvement, and training is terminated after $40$ unimproved epochs. With this approach, an AUC value of $87.71\%$ and an accuracy of $84.19\%$ on the validation data are achieved, and the model to be interpreted is selected.

\begin{figure}[!t]
    \centering
    \begin{subfigure}[b]{0.25\textwidth}
        \resizebox{\linewidth}{0.5\linewidth}{
            \includegraphics{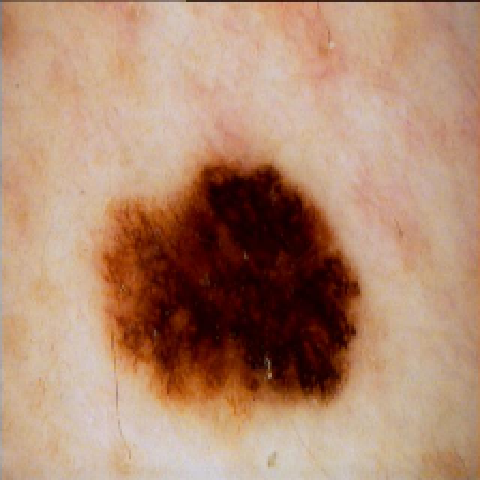}
        }
        \vskip5pt
        \resizebox{\linewidth}{0.5\linewidth}{
            \includegraphics{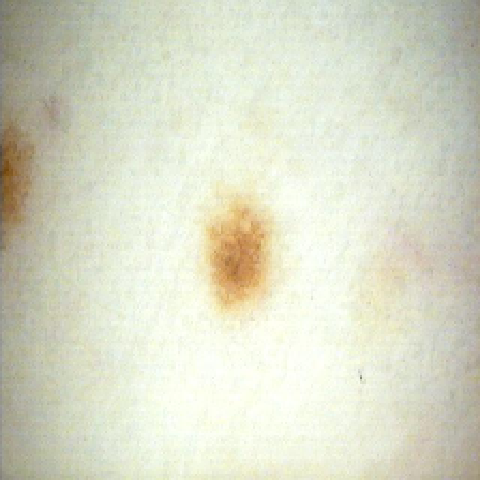}
        }
        \vskip5pt
        \resizebox{\linewidth}{0.55\linewidth}{
            \includegraphics{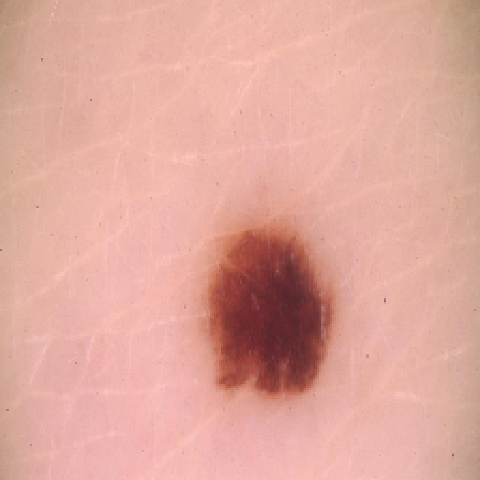}
        }
        \vskip12pt
        \caption{Input images}
        \label{fig:melanom_res_original}
    \end{subfigure}%
    \begin{subfigure}[b]{0.75\textwidth}
        \resizebox{\linewidth}{!}{
            \includegraphics{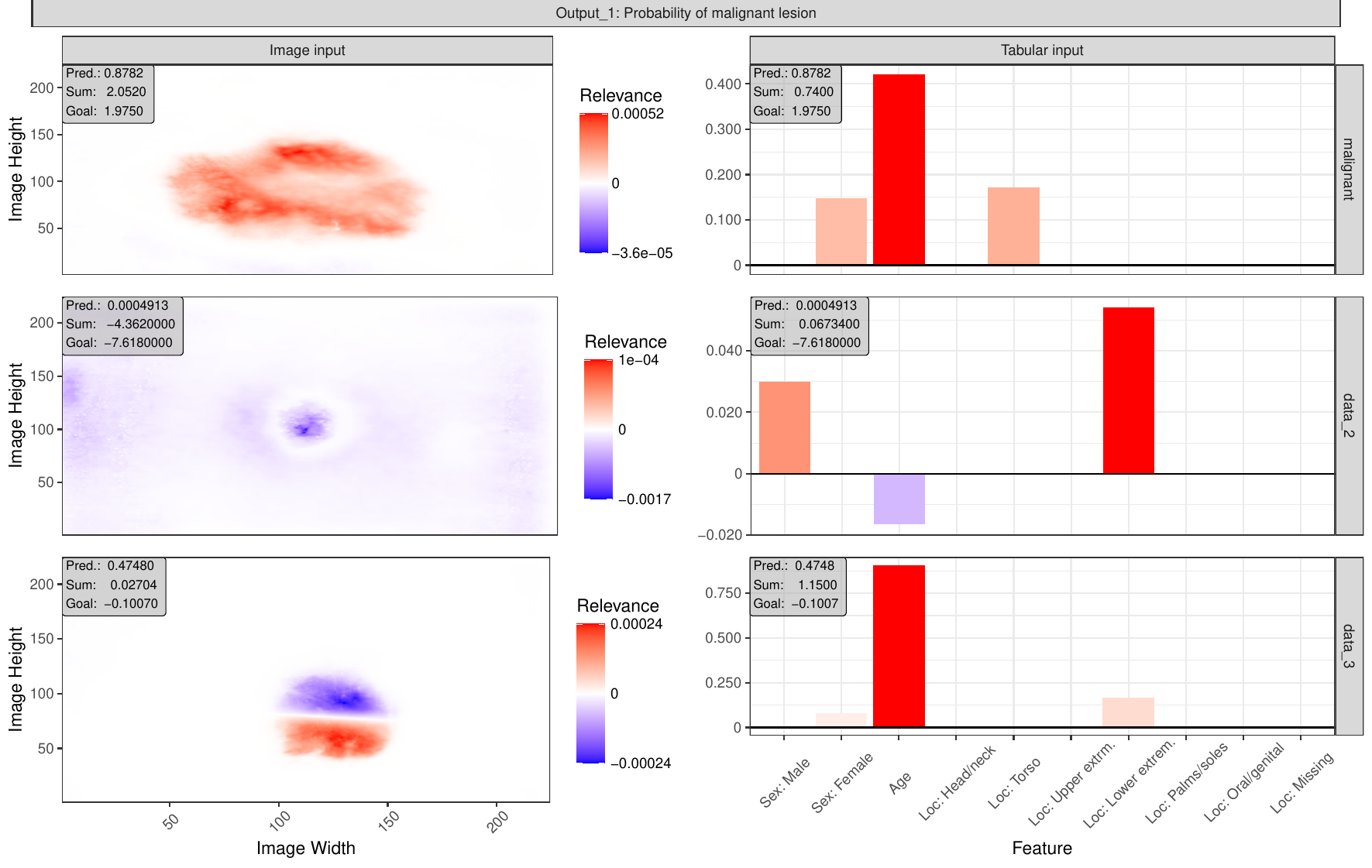}
        }
        \caption{Plot generated by \pkg{innsight}}
        \label{fig:melanom_res_innsight}
    \end{subfigure}
    \caption{The image part of the instance of the melanoma dataset to be explained and the associated visualization generated by \pkg{innsight}. Figure (a) shows a (top) malignant lesion image of a $65$-year-old female, (middle) benign lesion of a $40$-year-old male, and (bottom) malignant lesion of a $90$-year-old female patient. Figure (b) displays the LRP explanation of the patients from (a) created with the \code{plot()} function and subsequent minor modifications such as facet and x-axis labels.}
    \label{fig:melanom_res}
\end{figure}

Based on this model, the obtained predictions can now be explained using the 3-step approach of \pkg{innsight}: In the first step, the trained model is loaded and converted to a \pkg{torch}-based model using the \code{convert()} function for initializing a \code{Converter} object. However, since \pkg{keras} models do not include names of the input variables and output nodes, these can be passed along when initializing the converter to preserve meaningful labels of the input and output variables in the visualizations. Thus, the first step is executed by the following \proglang{R} code:

\begin{CodeChunk}
\begin{CodeInput}
R> library("keras")
R> library("innsight")
R> model <- load_model_tf("additional_files/melanoma_model.h5")
R> input_names <- list(
+   list(paste0("C", 1:3), paste0("H", 1:224), paste0("W", 1:224)),
+   list(c("Sex: Male", "Sex: Female", "Age",
+          "Loc: Head/neck", "Loc: Torso", "Loc: Upper extrm.",
+          "Loc: Lower extrem.", "Loc: Palms/soles", "Loc: Oral/genital",
+          "Loc: Missing")))
R> output_name <- c("Probability of malignant lesion") 
R> converter <- convert(model, input_names = input_names, 
+    output_names = output_name)
\end{CodeInput}
\end{CodeChunk}
Next, the LRP method with composite rules is applied, which selects the propagation rule depending on the layer type. For convolutional layers, the $\alpha$-$\beta$-rule with $\alpha = 1.5$ is used to favor the positive over the negative relevances. In addition, the $\varepsilon$-rule with $\varepsilon = 0.01$ is performed on all dense layers and the simple rule -- the rule used by default -- on average pooling layers. This second step is performed with \pkg{innsight} as follows:
\begin{CodeChunk}
\begin{CodeInput}
R> rule_name <- list(Conv2D_Layer = "alpha_beta", Dense_Layer = "epsilon")
R> rule_param <- list(Conv2D_Layer = 1.5, Dense_Layer = 0.01)
R> res <- run_lrp(converter, inputs, channels_first = FALSE, 
+    rule_name = rule_name, rule_param = rule_param)
\end{CodeInput}
\end{CodeChunk}
For the sake of simplicity, the loading of the input data \code{inputs} is omitted in the above code snippet and can be found in the reproduction material together with the whole example. In addition, the channel axis of the images is located at the last position, which is why the argument \code{channels_first} must be set to \code{FALSE}. The results can be visualized using the implemented \code{plot()} function. By default, the results are scaled using colors (red for positive and blue for negative relevances) for each instance, each considered output node and each input layer individually. This behavior is especially appropriate for models with multiple input layers consisting of images mixed with tabular data. Because even if the relevances are the same at the end of the tabular and image model before merging, they are further propagated to only ten input variables for the tabular and $224$$\times$$224$$\times3$ variables for the image model, leading to potentially different relevance scales. The following code produces the plot object based on the \code{S4} class \code{innsight_ggplot2} for the first three dataset instances, which can be treated as  a \pkg{ggplot2} object (see Appendix~\ref{sec:app_usage_advanced} for details):
\begin{CodeChunk}
\begin{CodeInput}
R> library("ggplot2")
R> p <- plot(res, data_idx = 1:3) + theme_bw()
\end{CodeInput}
\end{CodeChunk}
Since this model has no standard architecture and the visualization is more extensive, the suggested packages \pkg{gridExtra} \citep{gridextra_R} and \pkg{gtable} \citep{gtable_R} are required. However, each individual plot in the object \code{p} can now be modified individually based on the \pkg{ggplot2} syntax. The indexing works as the objects are plotted in a matrix-wise fashion, provided by the facet rows and columns. It is pointed out that each plot object is based on the same dataset, which is also created by the method \code{get_result(type = "data.frame")}, i.e., the same column names can be used within the \pkg{ggplot2} syntax. In the following code snippet, the facet and the x-axis labels are changed manually, and the plot is visualized, which can be found in Figure~\ref{fig:melanom_res}:
\begin{CodeChunk}
\begin{CodeInput}
R> p[1, 1] <- p[1, 1, restyle = FALSE] +
+    facet_grid(cols = vars(model_input),
+               labeller = as_labeller(c(Input_1 = "Image input")))
R> p[1, 2] <- p[1, 2, restyle = FALSE] +
+    facet_grid(cols = vars(model_input), rows = vars(data),
+               labeller = as_labeller(c(Input_2 = "Tabular input",
+                                        data_1 = "malignant")))
R> p <- p + theme(axis.text.x = element_text(angle = 45, vjust = 0.6))
R> plot(p, heights = c(0.31, 0.31, 0.38))
\end{CodeInput}
\end{CodeChunk}
 The argument \code{restyle} is set when indexing the \code{innsight_ggplot2} object, ensuring that the subplots are extracted in the same way as they are displayed in the whole plot. Otherwise, the entire plot's corresponding facet stripes and axis labels are transferred to the selection. In addition, the arguments in the generic function \code{plot()} for \code{innsight_ggplot2} objects are forwarded to the function \code{gridExtra::arrangeGrob()} when the plot is finally rendered. This feature allows adjusting the relative heights and widths, demonstrated in the last line of code, to slightly compensate for the increased vertical space of the rotated axis labels.

 The three instances in Figure~\ref{fig:melanom_res} describe different explanatory approaches to the trained model's predictions: The top image in Figure~\ref{fig:melanom_res_original} of a malignant lesion was recorded on the torso of a $65$-year-old female patient. In the associated interpretation generated by \pkg{innsight} (top row in Fig.~\ref{fig:melanom_res_innsight}), it can be observed that, on the one hand, the model identifies the lesioned skin area. On the other hand, the darker and patchy pigmentation and the ragged borders positively influenced the prediction of $87.82\%$ for melanoma (shown as $0.8782$ in the corresponding infobox in Fig.~\ref{fig:melanom_res_innsight}). This observation is also consistent with the official ABCD checklist for melanoma \citep{abcde}, which states that asymmetry, irregular borders, varying color, and large diameters are indicative of a malignant skin lesion. However, the patient's age also positively affected the prediction, as evident from the tabular patient-level information explanation in Figure~\ref{fig:melanom_res_innsight}. A complimentary picture results from the middle image in Figure~\ref{fig:melanom_res_original}, showing a benign mole located on the lower extremities of a $40$-year-old man. The model predicted a probability of only $0.05\%$ for a malignant lesion and explains its decision with the symmetrical shape, uniform color pigmentation, and lack of notched borders. In addition, the age of $40$ also has a slightly negative influence on the prediction, whereas the location seems to contribute a large positive effect (middle row in Fig.~\ref{fig:melanom_res_innsight}). The last instance exemplifies a situation where the model is uncertain whether it is a malignant or benign skin lesion since its prediction is $47.48\%$. The truly malignant skin area originates from the lower extremities of a $90$-year-old woman (bottom image in Fig.~\ref{fig:melanom_res_original}). Especially the image input explanation in the last row in Figure~\ref{fig:melanom_res_innsight} shows the model's uncertainty because the mole's upper part looks very regular, arguing for a healthy lesion and consistently highlighted with negative relevance (blue) by the model's explanation. In contrast, the lower part contains some notches potentially favoring melanoma, which the model also correctly identified. Furthermore, the high age of the $90$-year-old patient has a strong positive relevance to the model's prediction, demonstrating the strong effect of the feature age. The corresponding infoboxes in Figure~\ref{fig:melanom_res_innsight} also show that -- while the decomposition goals are not achieved, which, however, is expected in LRP due to the absorption of relevance into the bias vectors -- the majority of relevance for the top two patients comes from the image and less from patient-level data. Only in the case of the last patient do the image relevances seem to cancel each other out so that the tabular features have the strongest contribution to the prediction.

\section{Validation and runtime} \label{sec:validation}

To evaluate the validity and computational performance of \pkg{innsight}, the results of the presented feature attribution methods on simulated models and data are compared with the results of the \proglang{Python} implementations \pkg{zennit} \citep{zennit_2021}, \pkg{innvestigate} \citep{innvestigate_2019}, \pkg{captum} \citep{captum_2020}, \pkg{deeplift} \citep{shrikumar_2017}, and \pkg{shap} \citep{Lundberg_2017}. The packages \pkg{deeplift} and \pkg{innvestigate} are based on the high-level machine learning library \pkg{Keras} \citep{chollet2015} and utilize \pkg{TensorFlow} \citep{tensorflow} as the backend for all calculations. In addition, both packages initially create a replication of the passed model with the interpretation methods pre-implemented in the individual layers, similar to \pkg{innsight}. In contrast, the packages \pkg{zennit} and \pkg{captum} use \pkg{PyTorch} \citep{pytorch} and run without a conversion step since hooks are used to modify the automated backward pass according to the applied method on the fly (see Appendix~\ref{app:model} for more details). The package \pkg{shap} can handle both \pkg{Keras} and \pkg{PyTorch} models, and also uses hooks to modify automatic gradients. However, this only enables the application of methods that can be considered independent of the preceding and following layers, which complicates, for example, an implementation of DeepLift with the RevealCancel rule. Furthermore, not every package supports all methods. For example, \pkg{innvestigate} and \pkg{zennit} are more geared towards standard gradient methods and LRP, whereas \pkg{deeplift} is more or less an implementation of the methods from its associated paper and focuses on the DeepLift method. \pkg{shap}, originating from a methods paper, primarily considers Shapley-value-based methods. On the other hand, \pkg{captum} is a good all-rounder, but still has some gaps in the context of LRP. A summary of the packages' implemented feature attribution methods is provided in Table~\ref{tab:pkgs}.

\begin{table}[h]
		\caption{Summary of the implemented feature attribution methods in each package.}\label{tab:pkgs}
		\centering
		\begin{adjustbox}{width=0.95\textwidth}
			\begin{tabular}{l*{6}{>{\centering\arraybackslash}p{0.13\textwidth}}}
				\toprule\toprule
				& \multicolumn{6}{c}{\textbf{Package}}\\
				\cmidrule[0.8pt]{2-7}
				& \pkg{captum} & \pkg{deeplift}  & \pkg{innvestigate} & \pkg{shap} & \pkg{zennit} & \pkg{innsight}\\
				\midrule[0.6pt]
				Gradient & \checkmark & \checkmark & \checkmark & \xmark & \checkmark & \checkmark \\
				SmoothGrad & \checkmark & \checkmark & \checkmark & \xmark & \checkmark	 & \checkmark \\
				Gradient$\times$Input & \checkmark & \checkmark & \checkmark & \xmark & \checkmark  & \checkmark \\
				IntegratedGradient & \checkmark & $\hphantom{^*}$\checkmark$^*$ & \checkmark & \xmark & \checkmark & \checkmark \\ 
				ExpectedGradient & \checkmark &  \xmark & \xmark & \checkmark & \xmark & \checkmark \\ \hline
				LRP (simple rule) & \checkmark & \xmark & \checkmark & \xmark & \checkmark & \checkmark \\
				LRP ($\varepsilon$-rule) & \checkmark & \xmark & \checkmark & \xmark & \checkmark  & \checkmark \\
				LRP ($\alpha$-$\beta$-rule)& \xmark & \xmark & $\hphantom{^{\dagger}}$\checkmark$^{\dagger}$ & \xmark & \checkmark	 & \checkmark \\ \hline
				DeepLift (rescale) & \checkmark & \checkmark & \xmark & \xmark & \xmark & \checkmark \\
				DeepLift (reveal-cancel) & \xmark & \checkmark & \xmark & \xmark & \xmark & \checkmark  \\
				DeepSHAP & \checkmark & \xmark & \xmark & \hphantom{$^{\ddagger}$}\checkmark$^{\ddagger}$ & \xmark & \checkmark \\
				\bottomrule \bottomrule
                    \multicolumn{7}{l}{\footnotesize $^*$\pkg{deeplift} applies the middle Riemann sum, whereas \pkg{captum}, \pkg{zennit} and \pkg{innsight} use right Riemann sum.}\\
                    \multicolumn{7}{l}{\footnotesize $^{\dagger}$\pkg{innvestigate} uses another definition of the positive and negative part of the bias vector (see. Appendix~\ref{app:lrp_bias}).}\\
                    \multicolumn{7}{l}{\footnotesize $^{\ddagger}$ For max pooling layers, \pkg{shap} uses the cross maximal value of the $\bm{x}$ and reference value $\bm{\tilde{x}}$ instead their individual maximal value.}\\
			\end{tabular}
		\end{adjustbox}
	\end{table}

\subsection{Validity comparison} \label{sec:val_validity}

For the validation, shallow untrained dense and convolutional models with the most commonly used layer types -- such as 2D convolution, 2D maximum/average pooling, and dense layers -- and normally distributed input data are generated. More specifically, $32$ different architectures are considered, using ReLU and hyperbolic tangent to include both constrained and unconstrained activation functions, with and without bias vectors, with varying pooling layers, and a different number of output nodes. From each of these architectures, $50$ randomly initialized models are created, resulting in $1\,600$ distinct models, and evaluated on normally distributed datasets with $16$ input instances each. The experimental details can be found in the Appendix \ref{app:validity}. Moreover, all figures and results are reproducible using the code in the reproduction material or on GitHub (\url{https://github.com/bips-hb/JSS_innsight}).

\begin{figure}[!t]
    \centering
    \begin{subfigure}[b]{0.33\textwidth}
        \resizebox{\linewidth}{!}{
            \includegraphics{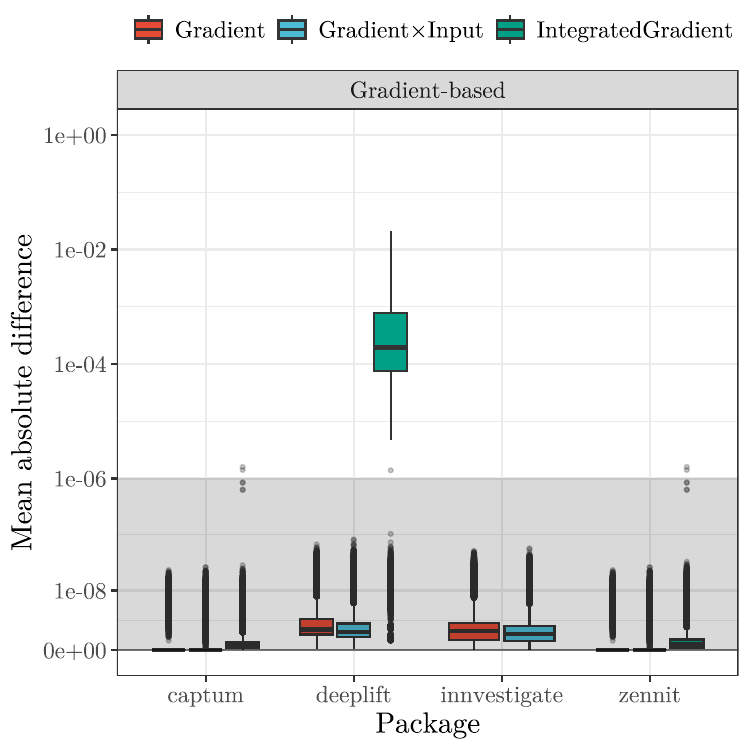}
        }
        \caption{Gradient-based (Sec.~\ref{sec:gradient_based})}
        \label{fig:result_mae_a}
    \end{subfigure}%
    \begin{subfigure}[b]{0.33\textwidth}
        \resizebox{\linewidth}{!}{
            \includegraphics{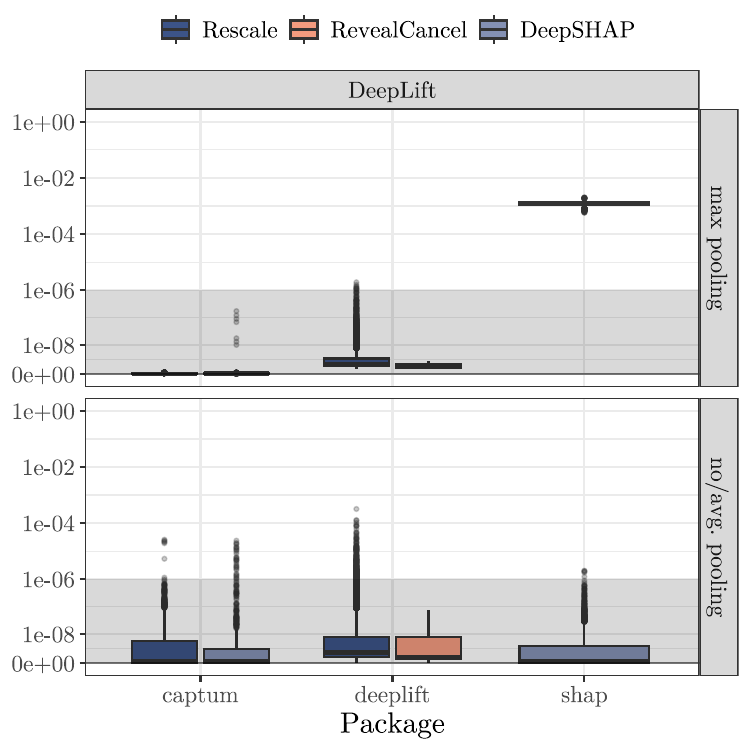}
        }
        \caption{DeepLift (Sec.~\ref{sec:Deeplift})}
        \label{fig:result_mae_b}
    \end{subfigure}%
    \begin{subfigure}[b]{0.33\textwidth}
        \resizebox{\linewidth}{!}{
            \includegraphics{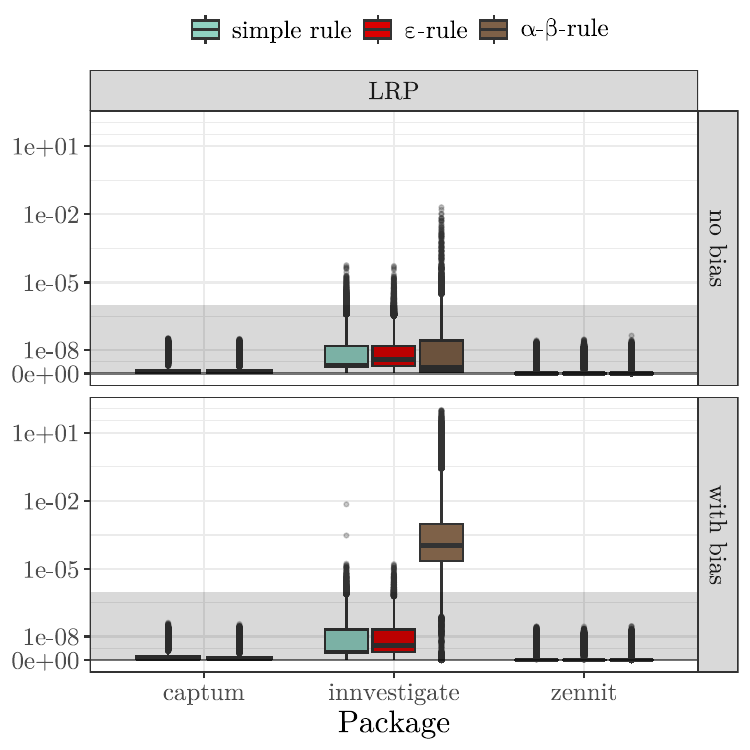}
        }
        \caption{LRP (Sec.~\ref{sec:LRP})}
        \label{fig:result_mae_c}
    \end{subfigure}
    \caption{Comparison of feature attribution methods' results of \pkg{innsight} and the reference implementations \pkg{captum}, \pkg{zennit}, \pkg{innvestigate}, \pkg{deeplift} and \pkg{shap} regarding the mean absolute difference as box plots over different model architectures and repetitions. It shows the results separated into (a) gradient-based methods, (b) DeepLift, and (c) LRP. The shaded gray area indicates the error tolerance of $10^{-6}$.}
    \label{fig:result_mae}
\end{figure}

As a measure of quality, the mean absolute difference between the results of \pkg{innsight} and the corresponding reference implementation over all input variables and output nodes is considered. Consequently, for each combination of method, model, input instance, and output node, a value for this quality measure is derived, leading to box plots for visualizing the differences. In addition to the box plots, an acceptable error range of up to $10^{-6}$ is highlighted in light gray to distinguish numerical tolerated differences caused by calculations of single-precision floating point numbers according to the IEEE 754 standard \citep{ieee_754} from abnormal discrepancies. The results are summarized in Figure~\ref{fig:result_mae}. 

For the gradient-based methods Gradient and Gradient$\times$Input, the method's results mostly coincide for all packages (see Fig.~\ref{fig:result_mae_a}). The only discrepancy is with the IntegratedGradient method for the package \pkg{deeplift}, but this is due to the fact that \pkg{innsight} approximates the integral with the right and \pkg{deeplift} with the middle Riemann sum\footnote{see \url{https://github.com/kundajelab/deeplift/blob/master/deeplift/util.py\#L261}}. The right Riemann sum is also used in \pkg{zennit}\footnote{see \url{https://github.com/chr5tphr/zennit/blob/0.5.1/src/zennit/attribution.py\#L453}}, whereas various approximation methods can be selected in \pkg{captum}. 

A similar picture results for the DeepLift method with the Rescale and RevealCancel rules, but with a few outliers with a maximum error of up to $10^{-3}$ for the Rescale rule (see Fig.~\ref{fig:result_mae_b}). However, all outliers with an error exceeding $10^{-6}$ originate from models with the hyperbolic tangent as activation and can thus be explained by numerical inaccuracies due to the saturated activation. In addition, minor discrepancies are probably caused by different treatments of vanishing denominators in the multipliers or numerical uncertainties between the backends \pkg{PyTorch}/\pkg{LibTorch} and \pkg{TensorFlow} in general. With the DeepSHAP method -- similar to IntegratedGradient -- the results of \pkg{innsight} largely match those of \pkg{captum}, but they differ from the results of \pkg{shap}. However, this is mainly due to the max pooling layer (see Fig.~\ref{fig:result_mae_b}), which is handled differently in \pkg{shap} than in \pkg{innsight} and \pkg{captum}. Since DeepSHAP is a repeated application of DeepLift with different reference values, the numerical inaccuracies from the DeepLift method accumulate and, thus, cause the visible outliers. 

For the LRP methods, a few adjustments are needed for the \pkg{innvestigate} and \pkg{captum} packages since they use only the simple rule for average pooling layers, which is modified in \pkg{innsight} using composite rules. Apart from that, the results from \pkg{innsight} compared to \pkg{captum} or \pkg{zennit} for the simple, $\varepsilon$-rule and $\alpha$-$\beta$-rule differ negligibly and are far below the maximally tolerated error of $10^{-6}$ (see Fig.~\ref{fig:result_mae_c}). For the simple and $\varepsilon$-rule, \pkg{innvestigate} is consistent with \pkg{innsight} except for a few deviations. Again, almost all of the cases with errors exceeding $10^{-6}$ are caused by a saturated hyperbolic tangent activation, lower errors on different stabilizers for the denominators in the relevance messages, and general numerical inaccuracies between their backends. However, significant discrepancies can be observed using the $\alpha$-$\beta$-rule, which only occur in models with a bias vector (see Fig.~\ref{fig:result_mae_c} bottom). The reason for this is a different interpretation of the positive or negative part of the bias vector, which is discussed in more detail in the Appendix~\ref{app:lrp_bias}.

\subsection{Runtime comparison}

In addition to comparing whether \pkg{innsight}'s results are consistent with the reference implementations, a runtime comparison is also conducted concerning the number of output nodes, hidden units or filters, hidden layers, batch size, and, for images, the size of the input images. It must be noted again that the packages based on \pkg{Keras} and \pkg{innsight} first convert the passed model, and the \pkg{PyTorch}-based packages use hooks to overwrite the automated backward pass while executing, making them considerably faster. Therefore, in the results, only the execution time excluding the conversion step -- as far as possible -- is presented and not the total time. For comparisons of the total time needed to calculate an explanation, see Appendix~\ref{app:additional_figs}. However, the \pkg{innvestigate} package has a special characteristic in this regard since the entire conversion process and the construction of the underlying graph only happens during the analysis of the first batch of input data\footnote{See the GitHub issue \href{https://github.com/albermax/innvestigate/issues/50}{\#50} and \href{https://github.com/albermax/innvestigate/issues/129}{\#129} for the \pkg{innvestigate} package.}. For this reason, conversion times are almost hardly present in the results. Since this simulation assumes that an interpretation method is being applied for the first time to a model and only to a single batch of input instances, the results of \pkg{innvestigate} are slightly biased and would be notably quicker if the same model is employed with more input batches. 

Analogously to the comparison from Section~\ref{sec:val_validity}, untrained dense and convolutional neural networks, and normally distributed input data are used for the time comparisons. Depending on the type of time comparison, the hyperparameters for the number of output nodes, number of hidden units or filter size, number of hidden layers, batch size, and the size of the input images are varied. The hyperparameters not considered in the respective comparison remain unchanged and take default values, i.e., one output node, $128$ hidden units for the tabular model and $5$ filters for the image model, two layers in total, a batch size of $16$ and an input image size of $64$$\times64$. In addition, $20$ replicates of each architecture are created to compensate for potential numerical fluctuations. For a more detailed simulation description or analysis of the results, including the total time, please refer to Appendix~\ref{app:time} and \ref{app:additional_figs}, and for a reproduction of the results, see the reproduction material or the code in the GitHub repository at \url{https://github.com/bips-hb/JSS_innsight/}.

\begin{figure}[!t]
    \centering
    \begin{subfigure}[t]{\textwidth}
        \resizebox{\textwidth}{!}{
            \includegraphics{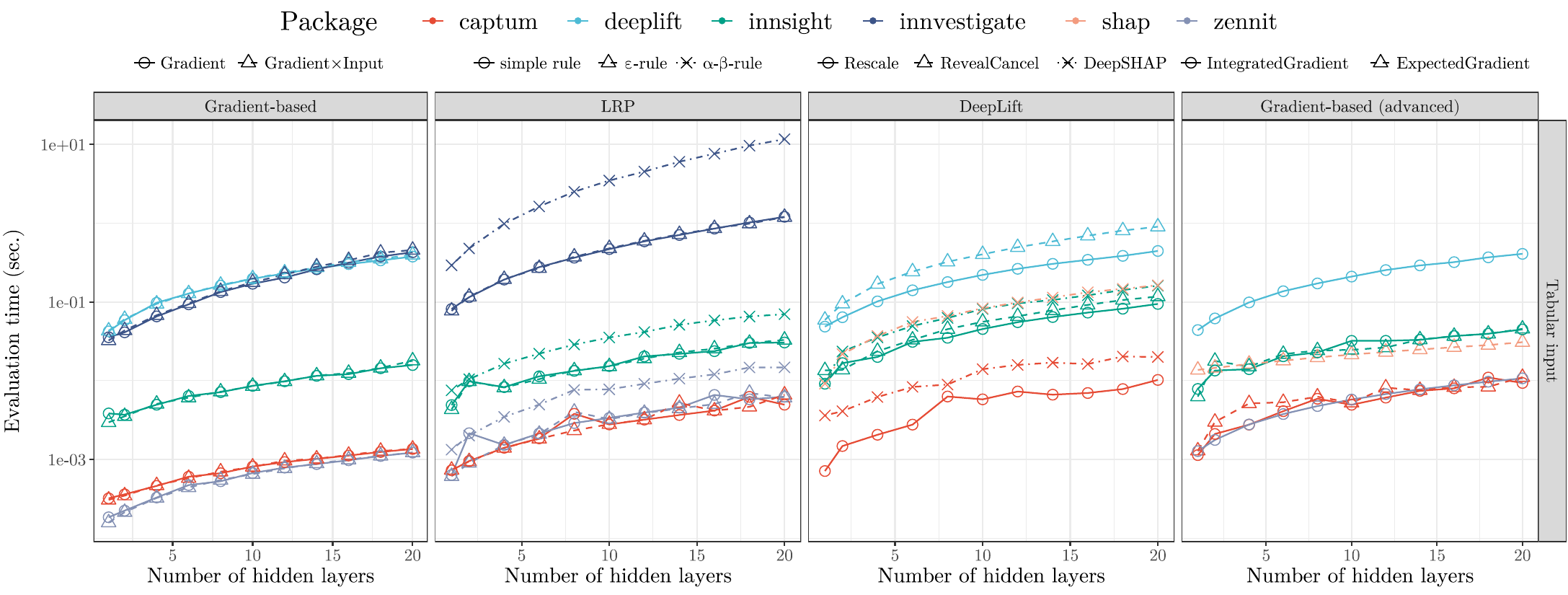}}
        \caption{Time comparison for a varying number of hidden layers (L)}
        \label{fig:val_time_L_eval}
    \end{subfigure}
    \hskip5pt
    \begin{subfigure}[b]{\textwidth}
        \resizebox{\textwidth}{!}{
            \includegraphics{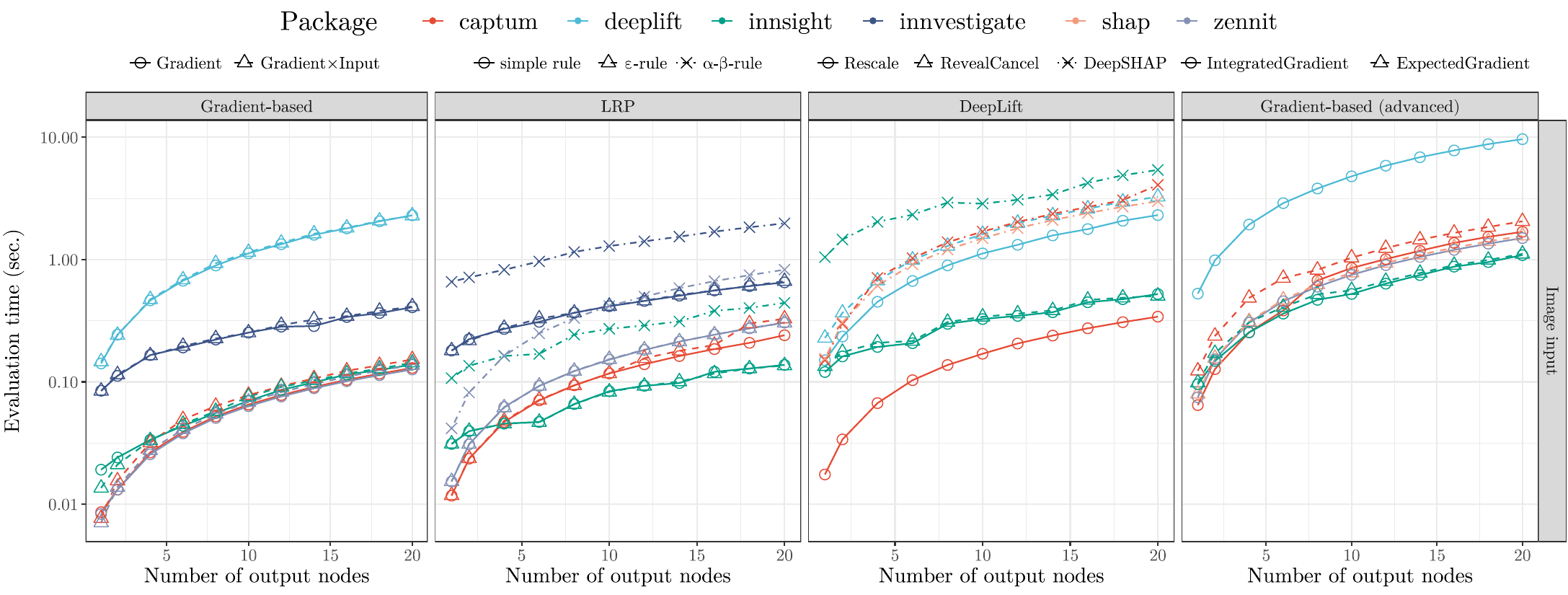}}
        \caption{Time comparison for a varying number of output nodes (C)}
        \label{fig:val_time_C_eval}
    \end{subfigure}
    \caption{Package's average evaluation time in seconds over $20$ repetitions for applying different feature attribution methods on models with (a) a varying number of hidden layers and (b) a varying number of output nodes (only image data).}
    \label{fig:val_time_eval}
\end{figure}

In general, comparing the runtimes of the different packages reveals that \pkg{innsight} is faster than \pkg{innvestigate} and \pkg{deeplift} (which are based on \pkg{Keras}), but slower than \pkg{captum}, \pkg{zennit} and \pkg{shap} (which are based on \pkg{PyTorch}). This overall trend is particularly evident when adding more layers to dense models: \pkg{innsight} is $10$-$15$ times slower than \pkg{captum} and \pkg{zennit}, but still for the same order of magnitude faster or even faster than \pkg{innvestigate} and \pkg{deeplift} (see Fig.~\ref{fig:val_time_L_eval}). In addition, \pkg{innsight} is similarly fast as \pkg{shap} for the methods DeepSHAP and ExpectedGradient. This trend can largely be extended to image inputs as well, with the exception that the evaluation times in \pkg{innsight} approach those of \pkg{PyTorch}-based implementations very closely. However, the DeepLift methods, particularly with the Rescale rule and thus even more with DeepSHAP, are considerably slower when applied to images compared to \pkg{deeplift} (see Fig.\ref{fig:val_time_L}).

The same trend can also be observed for a varying number of output nodes. In the case of image data, \pkg{innsight} is similarly fast as the \pkg{PyTorch}-based packages and mostly considerably faster than \pkg{innvestigate} and \pkg{deeplift}. However, the exception concerning the DeepLift methods applies here as well: DeepLift using the Rescale rule runs almost $10$ times slower than in \pkg{captum}. Consequently, DeepSHAP is also considerably slower than the runtime in \pkg{shap} and \pkg{captum} (see Fig.\ref{fig:val_time_C_eval}). On the other hand, \pkg{innsight} stands out for all rule-based methods applied to dense models and performs similarly or even faster than the \pkg{PyTorch}-based packages (see Fig.\ref{fig:val_time_C}). The primary reason for this is that the results for several output nodes can be calculated at once in \pkg{innsight}. In contrast, all other implementations only allow the calculation for single nodes, and thus, the method's results are computed by iterative execution. 

These tendencies can also be observed in the other simulations' results when changing the number of hidden units/filters, batch size, or image size (see Fig.~\ref{fig:val_time_U}, \ref{fig:val_time_B} and \ref{fig:val_time_W} in the Appendix). In the gradient-based methods and LRP, the execution times of \pkg{innsight} are increasingly approaching the times of the \pkg{PyTorch}-based implementations. With a larger number of filters in the convolutional models, the runtimes are almost identical. At the same time, it can be seen that the \pkg{Keras}-based packages run considerably slower than all other packages. An exception to this is with a high number of filters in convolutional models: in these cases, \pkg{innvestigate} and \pkg{innsight} applied to standard gradient-based methods and LRP are faster than all other implementations. In addition, the previously observed pattern with the DeepLift methods repeats in the other simulations as well: the execution time of DeepLift with the Rescale rule for image data is noticeably slower than in the reference implementations, which also explains the high execution time of DeepSHAP. Nevertheless, \pkg{innsight} is considerably faster than \pkg{deeplift} in dense models running DeepLift methods, and, in the case of batch size, also faster than \pkg{shap}.

In summary, \pkg{innsight} is mostly faster than the \pkg{Keras}-based packages and becomes more comparable to the \pkg{PyTorch} packages with increasing number of input instances, hidden units/filters, image size, and number of output nodes. However, the evaluation time of DeepLift with the Rescale rule and DeepSHAP applied to images is a weakness of \pkg{innsight} and shows the largest discrepancy in the runtime comparison. Even though the packages in this section are compared regarding runtime and some packages' weaknesses are revealed, all considered implementations provide an explanation within a reasonable time of a few seconds, even for deep neural networks with several large images as inputs.


\section{Summary and discussion} \label{sec:summary}

In summary, we have presented \pkg{innsight}, an \proglang{R} package that provides the most well-known feature attribution methods for interpreting neural network predictions. After a detailed introduction of the implemented feature attribution methods, the internal structure utilizing \pkg{torch}'s fast array calculations and conversion function \code{convert()} for initializing an \code{Converter} object demonstrated how the deep-learning-model-agnostic approach was implemented to enable the analysis of models from any \proglang{R} package in an efficient way. This flexibility is complemented by a unified 3-step approach from model to plotted results, including multiple visualization tools based on \pkg{ggplot2} or \pkg{plotly} for interactive plots. The step-wise procedure was illustrated using a model on the tabular penguin dataset and a deep neural network on the melanoma dataset consisting of structured patient-level information and images. Furthermore, the results of the simulation study show that \pkg{innsight} returns nearly identical feature-wise explanations to the reference implementations \pkg{captum}, \pkg{zennit}, \pkg{innvestigate}, \pkg{deeplift}, and \pkg{shap} in \proglang{Python}. Any observed differences are either below the accepted error tolerance of $10^{-6}$, reflecting negligible numerical inaccuracies, or caused by variations in layer treatments, e.g., max pooling layer in \pkg{shap} for DeepSHAP or another integral estimation for IntegratedGradient in \pkg{deeplift}. In terms of runtime, the package shows that it is generally faster than the \pkg{Keras}-based packages and slower or comparable fast than \pkg{captum}, \pkg{zennit}, and  \pkg{shap}. It only suffers with the DeepLift method using the Rescale rule for image data, which is also reflected in the DeepSHAP method that builds on it. Apart from that, the package also has some limitations that could be improved in the future. For example, only converting sequential models (i.e., \code{nn_sequential}) from the \pkg{torch} package is possible because no structured network graph can be extracted from an arbitrary \code{nn_module}. For all the gradient-based methods, however, this is not strictly necessary, which is why these could be extended for arbitrary \code{nn_module}s. Nevertheless, passing a model as a list allows the user to do the conversion step on their own in such cases. The package can also be extended methodically and offer permutation-based methods, e.g., Occlusion or RISE \citep{zeiler2014visualizing,rise}. Furthermore, an activation function is assigned to a linear or convolutional layer only if it is defined in the layer itself or immediately after the layer. This behavior is especially relevant for the RevealCancel rule in the DeepLift method, because \pkg{innsight} handles separated activations with the Rescale rule, which is the case, for example, with the layer sequence of convolution, batch normalization, and activation. Moreover, even if it is possible in the \pkg{torch} package, the \pkg{innsight} package currently only supports computations on CPUs and not on GPUs.


\section*{Computational details}

A 64-bit Linux platform running Ubuntu 20.04 with an AMD Ryzen Threadrippper 3960X (24 cores, 48 threads) CPU including 256 gigabyte RAM and two NVIDIA Titan RTX GPUs was used for all computations. All comparisons and calculations with the reference implementations were performed in a separate session -- created by \pkg{callr} \citep{callr_R} -- using only a single CPU thread per job. An exception was the neural network training on the melanoma dataset using a single GPU, which was also the only code executed by \proglang{Python} and not from \proglang{R} 4.3.2 \citep{R}. Due to the package requirement mismatch, separate environments were created for each of the \pkg{Keras}-based and the \pkg{PyTorch}-based packages, i.e.,
\begin{itemize}
    \item \pkg{innvestigate} 2.0.2: Using \proglang{Python} 3.8.15 with \pkg{Keras} 2.10.0 and \pkg{TensorFlow} 2.10
    \item \pkg{deeplift} 0.6.13: Using \proglang{Python} 3.6.15 with \pkg{Keras} 2.2.4 and \pkg{TensorFlow} 1.15
    \item \pkg{captum} 0.6.0, \pkg{zennit} 0.5.0 and \pkg{shap} 0.44.0: Using \proglang{Python} 3.8.12 with \pkg{PyTorch} 1.13.1 (cpu)
\end{itemize}
The corresponding environments were loaded in \proglang{R}, and then the code was executed in \proglang{Python} using \pkg{reticulate} 1.34 \citep{reticulate_R}. In addition, the computer was used exclusively for the runtime measurements for the corresponding job and was not distorted by other simultaneous processes.

\newpage

\section*{Acknowledgments}

This project was funded by the German Research Foundation (DFG), Emmy Noether Grant 437611051.

\bibliography{refs}

\newpage

\begin{appendix}

\section{Details on LRP and DeepLift}

\subsection{Details on LRP}\label{sec:app_LRP}

The \textit{layer-wise relevance propagation (LRP)} method was introduced by \cite{bach_2015} and has a similar goal as the Gradient$\times$Input approach: decomposing the output into variable-wise relevances conforming to Equation~\ref{Eq_decompose}. The distinguishing aspect is that the prediction $\hat{y}_c$ is redistributed layer by layer from the output node back to the inputs according to the layer's weights and intermediate values. The entire procedure is accomplished by rule-based relevance messages defining how to redistribute the upper-layer relevance to the lower layer. More precisely, the relevance message $r_{i \leftarrow j}^{(l, l+1)}$ describes the amount of the relevance $R_j^{l + 1}$ of node $j$ in layer $l+1$ sent to the lower-layer node $i$. The relevance for the lower-layer node $i$ is calculated as the sum of all incoming relevance messages, i.e.,
\begin{align}\label{app:lrp_messages}
    R_i^l = \sum_j r_{i \leftarrow j}^{(l, l+1)}.
\end{align}

In the following, we briefly overview the most popular variations of relevance messages flowing from a node of index $j$ in layer $l+1$ to node $i$ in the preceding layer:
\begin{itemize}
    \item \textbf{The simple rule:} The fundamental rule on which all other variations of relevance messages are more or less based is the \textit{simple rule} (also known as \textit{LRP-0}). The relevances are redistributed to the lower layers according to the ratio between local and global pre-activation. Let $\bm{x}$ be the inputs of the preceding layer, $\bm{w}$ the weight matrix and $\bm{b}$ the bias vector of layer $l$, and $R_j^{l+1}$ the upper-layer relevance; then $x_i\, w_{ij}$ is the local and $z_j = b_j + \sum_{k} x_k\, w_{kj}$ the global pre-activation defining the simple rule as
    \begin{align*}
        r_{i \leftarrow j}^{(l,\, l+1)} = \frac{x_i\, w_{ij}}{z_j} \, R_j^{l +1}.
    \end{align*}
    \item \textbf{The $\varepsilon$-rule:} One issue with the simple rule is that it is numerically unstable when the global pre-activation $z_j$ vanishes and causes a division by zero. The \textit{$\varepsilon$-rule} (also known as \textit{LRP-$\varepsilon$}) tackles those situations by adding a stabilizer $\varepsilon > 0$ that moves the denominator away from zero, i.e.,
    \begin{align*}
            r_{i \leftarrow j}^{(l,\, l+1)} = \frac{x_i\, w_{ij}}{z_j + \text{sign}(z_j)\, \varepsilon}\, R_j^{l+1}.
    \end{align*}
    This inserted value $\varepsilon$ absorbs some of the relevance and can, therefore, be utilized to achieve sparser and less noisy results for the explanation. As $\varepsilon$ increases, a greater portion of the relevance is intercepted, sustaining only the most salient relevances for this relevance message.
\end{itemize}
Both variants have in common that they distribute the upper-layer relevance proportionally downward regarding the local and global pre-activations, i.e.,  $x_i w_{ij}$ and $z_j$. Even though the $\varepsilon$-rule avoids division by zero, numerical inconsistencies can occur in both variants for very deep models. Since the pre-activations are not necessarily guaranteed to be positive, the local pre-activations may take on substantial positive or negative values that cancel out in the global pre-activation, leading to magnified values in the preceding layer. As a result, larger relevances in the lower layers potentially accumulate in deep models and increasingly reach the limits of computational representation of floating point numbers. To prevent this blow-up of relevances, the authors introduced the \textit{$\alpha$-$\beta$-rule}, which treats positive and negative pre-activations separately:
\begin{itemize}
    \item \textbf{The $\alpha$-$\beta$-rule:} The \textit{$\alpha$-$\beta$-rule} was introduced to avoid numerical instabilities and enable a weighting between positive and negative relevances depending on the user's focus. This relevance message applies the simple rule to the positive and negative parts of the pre-activations, respectively, and takes the weighted sum of both. The weighting can be regulated by the hyperparameters $\alpha, \beta \in \mathbb{R}$ satisfying $\alpha + \beta = 1$. Mathematically formulated, the rule is defined as follows:
    \begin{align*}
        r_{i \leftarrow j}^{(l,\, l+1)} = \left(\alpha \frac{(x_i\, w_{ij})^+}{z_j^+} + \beta \frac{(x_i\, w_{ij})^-}{z_j^-}\right)\, R_j^{l +1}
    \end{align*}
    \begin{align*}
        \text{with}\quad z_j^\pm = (b_j)^\pm + \sum_k (x_k\, w_{kj})^\pm,\quad (\cdot)^+ = \max(\cdot, 0),\quad
        (\cdot)^- = \min(\cdot, 0).
    \end{align*}
\end{itemize}
Since the bias vector $b_j$ is included in the computation of the global pre-activations in all presented variants, this term absorbs a certain amount of the upper-layer relevance. Consequently, the LRP methods approximate the output prediction rather than providing an accurate representation of the targeted decomposition in Equation~\ref{Eq_decompose}.

There are even more variants of relevance messages discussed in the literature suitable for various situations or layer types: For example, the \textit{deep Taylor decomposition} (also called \textit{$z^+$-rule}) in ReLU models -- also achieved with the $\alpha$-$\beta$-rule with $\alpha = 1$ -- allows filtering out only positive relevances \citep{montavon_2017}, or the \textit{$\gamma$-rule} favoring positive over negative relevances \citep{montavon_2019}. Moreover, some rules are specifically designed for the input layer \citep{montavon_2017}. Due to the rule independence of how the lower-layer relevances are computed from the relevance messages in Equation~\ref{app:lrp_messages}, the rules can also be set individually for each layer, called \textit{composite-rule} \citep{montavon_2019, kohlbrenner_2020}.

\subsection{Details on DeepLift}\label{sec:app_Deeplift}

The \textit{deep learning important features (DeepLift)} method introduced by \cite{shrikumar_2017} behaves similarly to LRP in a layer-by-layer backpropagation fashion from a selected output node back to the input variables considering the simple rule. However, it incorporates a reference value $\bm{\tilde{x}}$ to compare the relevances with each other. Hence, the relevances of DeepLift represent the relative effect of the outputs of the instance to be explained $f(\bm{x})_c$ and the output of the reference value $f(\bm{\tilde{x}})_c$. By taking the difference, the bias term is eliminated in the relevance messages, preventing the relevance absorption and leading to an exact variable-wise decomposition of the difference-from-reference output $\Delta \hat{y}_c = f(\bm{x})_c - f(\bm{\tilde{x}})_c$, i.e.,
\begin{align*}
    \Delta \hat{y}_c = f(\bm{x})_c - f(\bm{\tilde{x}})_c = \sum_{i=1}^d R_i^c.
\end{align*}
In contrast to the LRP method, DeepLift defines a multiplier layer by layer, starting from the output layer and propagating to the input layer instead of directly determining the relevances in each intermediate stage. Based on these multipliers, the contribution of an arbitrary variable to the difference-from-reference output can be obtained by multiplying it by the corresponding difference-from-reference input. For an arbitrary layer with the layer's input $\bm{x}$, reference input $\bm{\tilde{x}}$ and multiplier $m_{\Delta \bm{x} \Delta \hat{y}_c}$, this means:
\begin{align}\label{eq:App_DeepLift_multiplier}
    \sum_{i}  m_{\Delta x_i \Delta \hat{y}_c} \left( x_i - \tilde{x}_i \right) = m_{\Delta \bm{x} \Delta \hat{y}_c} \cdot \left(\Delta \bm{x}\right)^\top = \Delta \hat{y}_c.
\end{align}
The multipliers fulfill a chain rule allowing the computation of the multiplier for the preceding layer given the already calculated one $m_{\Delta \bm{t}, \Delta \hat{y}_c}$, i.e.,
\begin{align}\label{eq:App_DeepLift_chain_rule}
    m_{\Delta x_i \Delta \hat{y}_c} = \sum_{j} m_{\Delta x_i \Delta t_j}\, m_{\Delta t_j \Delta \hat{y}_c}.
\end{align}
In other words, the chain rule justifies defining the multipliers for each layer or part of a layer separately before combining them with the upper-layer multipliers. The authors distinguish between the linear and nonlinear components of a layer and provide definitions of the multipliers for each of them:
\begin{itemize}
    \item \textbf{Linear rule:} For the linear components of a layer, such as matrix multiplication in dense or convolution layers, the weights of the corresponding layer are used as the multipliers, i.e., $m_{\Delta x_i \Delta z_j} = w_{ij}$.
    \item \textbf{Rescale rule:} This rule can be used for all nonlinear parts of a layer that can be reduced to a one-dimensional function $\sigma$, e.g., all activations such as ReLU, tanh, or sigmoid. In this case, the ratio between the difference-from-reference activation $\Delta \sigma(z)_j = \sigma(z_j) - \sigma(\tilde{z}_j)$ and the pre-activation $\Delta z_j = z_j - \tilde{z}_j$ gives the multiplier, i.e., $m_{\Delta z_j \Delta \sigma(z)_j} = \frac{\Delta \sigma(z)_j}{\Delta z_j}$. To avoid numerical instability caused by a vanishing denominator, the gradient of $\sigma$ at $z_j$ is used instead of the multiplier when $z_j$ is close to its reference value $\tilde{z}_j$.
    \item \textbf{RevealCancel rule:} This rule is designed for non-linearities $\sigma$ to propagate meaningful relevances for saturated activations and discontinuous gradients through the layers' activation part, even when activations like ReLU eliminate the values. Similar to the normal pre-activations in the $\alpha$-$\beta$-rule for LRP, the positive $\Delta z_j^+$ and negative $\Delta z_j^-$ difference-from-reference pre-activations are considered separately, ensuring the propagation of expressive contribution scores. Descriptively, the \textit{RevealCancel} rule can be explained in a way that the multiplier for the positive part $m_{\Delta z_j^+ \Delta y_j^+}$ is the ratio between the average effect of $\Delta z_j^+$ after activating, before and after the negative part $\Delta z_j^-$ has been added, and the positive difference-from-reference pre-activation $\Delta z_j^+$. In the same way, the negative multiplier $m_{\Delta z_j^- \Delta y_j^-}$ is given by the ratio of the average impact of $\Delta z_j^-$ after activating, before and after the positive part $\Delta z_j^+$ has been added, to $\Delta z_j^-$. Mathematically, the rule is defined as
    \begin{align*}
        m_{\Delta z_j^\pm \Delta y_j^\pm} = \frac{\frac{1}{2}\left( \sigma(\tilde{z}_j + \Delta z_j^\pm) - \sigma(\tilde{z}_j) + \sigma(\tilde{z}_j + \Delta z_j^\pm + \Delta z_j^\mp) - \sigma(\tilde{z}_j + \Delta z_j^\pm)\right)}{\Delta z_j^\pm}.
    \end{align*}
\end{itemize}
These rules, along with the chain rule (Equations~\ref{eq:App_DeepLift_multiplier}-\ref{eq:App_DeepLift_chain_rule}), enable the successive computation of the input variables' contributions $R_i$ to the difference-from-reference output $\Delta \hat{y}_c$ in a single backward pass. 

\section{Remarks on accepted models and layers}\label{app:model}

As described in Section~\ref{sec:step_1}, conversion functions are only provided by default for the packages \pkg{torch}, \pkg{keras}, and \pkg{neuralnet} to transfer a neural network into a list structure, which is explained in more detail in the next paragraph. However, any model in this list format can be directly passed to the \code{Converter}. When initializing the \code{Converter} object, a \pkg{torch}-based copy of the model is then generated from this list. This is one of the most crucial differences compared to packages like \pkg{captum}, \pkg{zennit}, and \pkg{shap}, as these packages override the automatic differentiation graph in the backward pass of the method through so-called \textit{backward hooks}\footnote{See, e.g., \code{register\_full\_backward\_hook()} in \pkg{PyTorch} (\url{https://pytorch.org/docs/stable/generated/torch.nn.Module.html\#torch.nn.Module.register\_full\_backward\_hook}) or \code{tf.RegisterGradient} in \pkg{TensorFlow}/\pkg{Keras} (\url{https://www.tensorflow.org/api_docs/python/tf/RegisterGradient}).}. Consequently, there is no need to analyze and copy the computational graph of the neural network; instead, adjusting the gradient computation of the relevant layers in the backward pass is sufficient. This feature is not available in version 0.12.0 of \pkg{torch} in \proglang{R}, which is why we opted for copying the layers as in the packages \pkg{innvestigate} and \pkg{deeplift} and extended it to a deep-learning-model-agnostic approach. However, this advantage comes with the requirement that all calculations in the network need to be registered and transferred to \pkg{torch}. For example, the gradient function does not need to be overridden and registered for a flatten layer in \pkg{captum}, \pkg{zennit} and \pkg{shap} -- as it only rearranges the values -- aligning with the existing automatic differentiation function. In \pkg{innsight}, the entire model is copied, so an equivalent in \pkg{torch} must be created for this layer as well. In general, every model from the \pkg{neuralnet} and \pkg{keras} packages can be converted. In the case of \pkg{keras}, this includes both sequential models created with \code{keras_model_sequential()} and non-sequential models created with \code{keras_model()}, as long as they only include the accepted layers listed in Table~\ref{tab_app:layers}. Since it is not possible to reconstruct a computational graph for \pkg{torch} models, only \code{nn_sequential()} models are accepted, and only the layers listed in Table~\ref{tab_app:layers} are recognized.

\begin{table}[h]
    \caption{Summary of the accepted layer types for the packages \pkg{keras} and \pkg{torch}, as well as the types for the layers provided as a list object.}\label{tab_app:layers}
    \centering
    \begin{adjustbox}{width=0.9\textwidth}
	\begin{tabular}{llll}
		\toprule\toprule
		 & \multicolumn{3}{c}{\textbf{Package}}\\
		\cmidrule[0.8pt]{2-4}
		   & \pkg{keras} & \pkg{torch}  & as \code{list} (\code{type =}) \\
		\midrule[0.6pt]
		Dense								& \code{layer_dense()}		& \code{nn_linear()} 	& \code{"Dense"}	\\ \hline
		\multirow{2}{*}{Convolution}		& \code{layer_conv_1d()}	& \code{nn_conv1d()}	& \code{"Conv1D"}	\\
											& \code{layer_conv_2d()}	& \code{nn_conv2d()} 	& \code{"Conv2D"} 	\\ \hline
		\multirow{8}{*}{Pooling}		& 	\code{layer_max_pooling_1d()}	& \code{nn_max_pool1d()}	& \code{"MaxPooling1D"}	\\
										& 	\code{layer_max_pooling_2d()}	& \code{nn_max_pool2d()}	& \code{"MaxPooling2D"}	\\
										& 	\code{layer_average_pooling_1d()}	& \code{nn_avg_pool1d()}	& \code{"AveragePooling1D"}	\\
										& 	\code{layer_average_pooling_2d()}	& \code{nn_avg_pool2d()}	& \code{"AveragePooling2D"}	\\
										& 	\code{layer_max_pooling_1d()}	& 	& \code{"GlobalPooling"}	\\
										& 	\code{layer_max_pooling_2d()}	& 	&	\\
										& 	\code{layer_average_pooling_1d()}	& 	&	\\
										& 	\code{layer_average_pooling_2d()}	&	&	\\ \hline
		Batch		& 	\code{layer_batch_normalization()}	& \code{nn_batch_norm1d()}	& \code{"BatchNorm"}	\\
		Normalization								& 		& \code{nn_batch_norm2d()}	&	\\ \hline
		\multirow{6}{*}{Activation}		& 	\code{layer_activation_relu()}	& \code{nn_relu()}	& \code{"Activation"}	\\
										& 	\code{layer_activation_leaky_relu()}	&  \code{nn_leaky_relu()}	& with \code{"relu"}, \code{"softplus"},	\\ 
										& 	\code{layer_activation_softmax()}	&  \code{nn_softplus()}	& \code{"sigmoid"}, \code{"softmax"}, \\
										& 	\code{layer_activation()} with	&  \code{nn_sigmoid()}	& \code{"tanh"}, \code{"linear"}	\\
										& 	\code{"relu"}, \code{"softplus"},	&  \code{nn_softmax()}	&	\\
										& 	\code{"sigmoid"}, \code{"softmax"}, \code{"tanh"}	&  \code{nn_tanh()}	&	\\ \hline
		\multirow{7}{*}{Other}		& 	\code{layer_input()}	& \code{nn_flatten()}	& \code{"Flatten"}	\\
									& 	\code{layer_flatten()}	&  \code{nn_dropout()}	& \code{"Add"}	\\ 
									& 	\code{layer_add()}	&  	& \code{"Concatenate"}	\\
									& 	\code{layer_concatenate()}	&  	& \code{"Padding"}	\\
									& 	\code{layer_zero_padding_1d()}	&  	&	\\
									& 	\code{layer_zero_padding_2d()}	&  	&	\\
									& 	\code{layer_dropout()}	&  	&	\\ \hline
		\bottomrule
	\end{tabular}
    \end{adjustbox}
\end{table}

Internally, a model is being transferred from the packages \pkg{keras}, \pkg{torch}, and \pkg{neuralnet} into a list, from which a \pkg{torch}-based model is subsequently created. This list requires the entries \code{"input_dim"} representing the input dimension excluding the batch dimension, \code{"layers"} for the model's layers (again a list), and \code{"input_nodes"}/\code{"output_nodes"} for the indices of the model's input and output layers from \code{"layers"}. Additionally, input and output labels can be specified using the entries \code{"input_names"} and \code{"output_names"}. The input and output names are identical to the fields in the converter object, but when passing a model, they are optional and will be automatically filled otherwise. They always represent a list for each input or output layer, and then contain a list of names for each dimension. For example, \code{list(list("a", "b", "c"), list("1", "2", "3"))} is valid as input names for a model with one input layer that expects a two-dimensional input of shape $3$$\times$$3$. The entry \code{"layers"} again forms a list of individual layers of the neural network. For each layer, the \code{"type"} entry specifies the layer type, and, depending on the type, other relevant components of the layer. For example, a dense layer (\code{type = "Dense"}) has the entries \code{"weight"} for the weight matrix and \code{"bias"} for the bias vector. In addition to these layer-specific entries, each layer has entries \code{"input_layers"} and \code{"output_layers"}, indicating the indices in \code{"layers"} of the incoming and outgoing layers. All available layer types are listed in Table~\ref{tab_app:layers}. For a more detailed description of the entries and the requirements for the other layer types, please refer to the vignette "In-depth explanation" (see \code{vignette("detailed\_overview", package = "innsight")} or the online documentation at \url{https://bips-hb.github.io/innsight/articles/detailed_overview.html}). For the model trained on the bike sharing dataset in Section~\ref{sec:usage}, the argument \code{save_model_as_list} in the \code{convert()} function can be used to exemplify the list structure for a dense model:

\begin{CodeInput}
R> conv <- convert(model,
+    output_names = c("Number of rented bikes/10,000"),
+    save_model_as_list = TRUE)
R> str(conv$model_as_list, max.level = 3)
\end{CodeInput}
\begin{CodeOutput}
List of 7
 $ layers      :List of 2
  ..$ Dense_1:List of 8
  .. ..$ type           : chr "Dense"
  .. ..$ weight         :Float [1:64, 1:5]
  .. ..$ bias           :Float [1:64]
  .. ..$ activation_name: chr "logistic"
  .. ..$ dim_in         : int 5
  .. ..$ dim_out        : int 64
  .. ..$ input_layers   : num 0
  .. ..$ output_layers  : num 2
  ..$ Dense_2:List of 8
  .. ..$ type           : chr "Dense"
  .. ..$ weight         :Float [1:1, 1:64]
  .. ..$ bias           :Float [1:1]
  .. ..$ activation_name: chr "linear"
  .. ..$ dim_in         : int 64
  .. ..$ dim_out        : int 1
  .. ..$ input_layers   : num 1
  .. ..$ output_layers  : num -1
 $ input_dim   :List of 1
  ..$ : int 5
 $ output_dim  :List of 1
  ..$ : int 1
 $ input_names :List of 1
  ..$ :List of 1
  .. ..$ : Factor w/ 5 levels "holiday","workingday",..: 1 2 3 4 5
 $ output_names:List of 1
  ..$ :List of 1
  .. ..$ : Factor w/ 1 level "Number of rented bikes/10,000": 1
 $ input_nodes : num 1
 $ output_nodes: int 2
\end{CodeOutput}

\section{Advanced visualization}\label{sec:app_usage_advanced}

In Sections~\ref{sec:step_3} the basic \code{plot()} and \code{plot_global()} functions have already been explained. As mentioned there, these functions create either an object of the \code{S4} class \code{innsight_ggplot2} (if \code{as_plotly = FALSE}) or one of the \code{S4} class \code{innsight_plotly} (if \code{as_plotly = TRUE}). These functions are intended to generalize the usual \pkg{ggplot2}, or \pkg{plotly} objects since, with these packages, the limits of clear visualization possibilities for models with multiple input layers are quickly reached. For example, two charts with different scales for each output node or class need to be generated in a model with images and tabular data as inputs. In this case, a \pkg{ggplot2}-based or \pkg{plotly}-based plot is generated for each single input instance and output node and then combined into one large visualization using \code{arrangeGrob()} from \pkg{gridExtra} \citep{gridextra_R} or \code{subplot()} from \pkg{plotly}, respectively. In contrast, the \code{S4} class \code{innsight_ggplot2} behaves as a wrapper for the \pkg{ggplot2} object for ordinary models with only one input or output layer. Nevertheless, instances of the \code{innsight_ggplot2} class can be treated and modified as regular \pkg{ggplot2} objects providing a \pkg{ggplot2}-typical usage by adding, for example, themes, scales, or geometric objects; hence the intermediate step via this class is generally not noticeable to the user. For example, the following code is valid:
\begin{Code}
plot(method) +
    ggplot2::theme_bw() +
    ggplot2::xlab("My new x label") +
    ggplot2::scale_y_continuous(trans = "pseudo_log") +
    ggplot2::geom_text(ggplot2::aes(label = signif(value))
\end{Code}
Conveniently, all \pkg{ggplot2} objects are based on the same \code{data.frame}, which is also obtained via the \code{get_result()} method (see Sec.~\ref{sec:get_result}), i.e., the corresponding column names can be used as variables in the \pkg{ggplot2} objects, as can be seen in the last line of the code chunk above. For objects of the \code{innsight_plotly} class, the entire plot is always created using the \code{plotly::subplot()} function. However, this has the consequence that individually assigned modifications are partially overwritten by the grouping, which is why the usual \pkg{plotly}-typical adaptations can only be performed after the \code{innsight_plotly} object has been printed and returned by the generic \code{print()} function for this class, i.e.,
\begin{Code}
print(plot(method, as_plotly = TRUE)) 
    plotly::hide_colorbar() 
    plotly::layout(xaxis = list(title = "My new x label"))
\end{Code}
In addition, generic functions for both \code{S4} classes are implemented, which provide a deeper and more detailed examination of an already created plot through indexing or indexed modification. Section~\ref{sec:illu_melanom} demonstrates the application and illustration of some of these generic methods using visualized explanations of a model that takes tabular and image data as inputs. However, for a more detailed description and usage of these classes, please refer to the vignette "In-depth explanation" (see \code{vignette("detailed\_overview", package = "innsight")} or the online documentation at \url{https://bips-hb.github.io/innsight/articles/detailed_overview.html}).

\section{Simulation details} \label{app:simulation}

\subsection{Validity comparison} \label{app:validity}

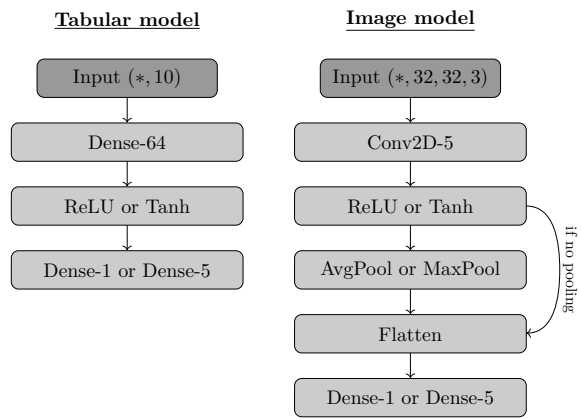
\begin{wrapfigure}[17]{r}{0.5\textwidth}
    \centering
    \resizebox{0.5\textwidth}{!}{
        \input{tikz/appendix_models_1.tikz}
    }
    \caption{The basic setup of the architectures for the comparison simulation. The expression \newline "-n" indicates the number $n$ of units for dense and the number of filters for convolutional layers.}
    \label{fig:app_sim_models}
\end{wrapfigure}
In order to verify the correctness of the \pkg{innsight} package, a comparative simulation is performed with the reference implementations \pkg{zennit}, \pkg{captum}, \pkg{innvestigate}, \pkg{deeplift} and \pkg{shap} using convolutional and dense neural networks, for which the basic structure is shown in Figure~\ref{fig:app_sim_models}. The dense architecture starts with an input of $10$ input variables, then has $64$ units in the middle hidden layer and either one or five output units. In addition, either ReLU or hyperbolic tangent is used to consider both unbounded and bounded activation functions. The convolutional architecture starts with inputs of shape $32$$\times32$$\times3$ and a convolutional layer with five filters and a kernel size of $5$$\times$$5$, followed by activation with ReLU or hyperbolic tangent. Models are created with and without a pooling layer, which comes at this point. Average or maximum pooling layers with a kernel size of $3$$\times$$3$ are used. Nevertheless, if no pooling layer is considered, strides of $2$$\times$$2$ are used in the preceding convolutional layer to get a similar number of units after flattening. A dense layer with one or five output nodes follows the flattening. Based on these models and normally distributed dataset with $16$ input instances, the following methods are compared:
\begin{itemize}
    \item Gradient-based: Gradient, Gradient$\times$Input, IntegatedGradient with \code{n = 20} and with zeros and normally distributed reference values.
    \item LRP: simple rule, $\varepsilon$-rule with $\varepsilon = 0.01$, $\alpha$-$\beta$-rule with $\alpha = 1$ and $\alpha = 2$.
    \item DeepLift: Rescale and RevealCancel rule with zeros and normally distributed reference values each.
    \item DeepSHAP with $32$ baseline values.
\end{itemize}
In this comparison, the methods SmoothGrad and ExpectedGradient are excluded because they are based on sampling, and consequently, they would only yield the same results with the exact same seed and calculation order. Basically, both methods, however, rely on the gradient calculation of Gradient or IntegratedGradient. Therefore, an agreement with these methods can also imply the validity of SmoothGrad and ExpectedGradient.

\subsection{Runtime comparison} \label{app:time}

\begin{wrapfigure}{r}{0.5\textwidth}
    \resizebox{0.49\textwidth}{!}{
        \input{tikz/time_models.tikz}
    }
    \captionof{figure}{Model architectures for the runtime comparison. The hyperparameters for the number of outputs (C), number of hidden units or filter size (U), number of hidden layers (L), batch size (B), and the height/width of the input images (W) were varied in each case. }
    \label{fig:val_time_models}
\end{wrapfigure}
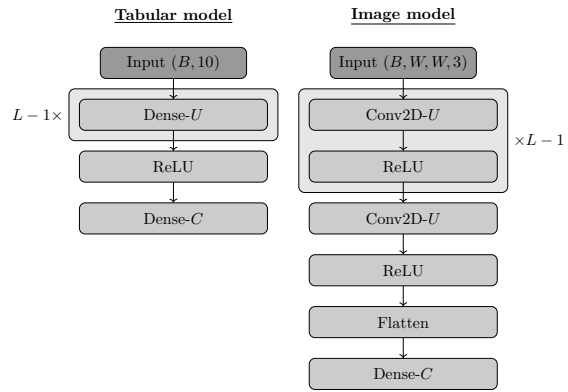
Besides the consistency of the methods' results, a runtime comparison of basic dense and convolutional models between \pkg{innsight} and the reference implementations \pkg{zennit}, \pkg{captum}, \pkg{innvestigate} and \pkg{deeplift} is also performed regarding several aspects: The number of output nodes (C), hidden units or filters (U), depth of the model (L), batch size (B), and for images the height/width (W) are varied resulting in the different architectures described in more detail in Figure~\ref{fig:val_time_models}. In the image model, for the 2D convolutional layer, which is repeated $L-1$ times, a kernel size of $5$$\times$$5$ with default strides of $1$$\times$$1$ and padding is applied so that the output shape corresponds to the input shape. But for the subsequent convolutional layer, a valid padding with strides of $\frac{W - 4}{6}$$\times$$\frac{W - 4}{6}$ is used, producing an equal number of flattened values regardless of the height and width $W$ of the input image. Overall, the time comparison is performed for the following methods, and the average time of $20$ replications is calculated:

\begin{itemize}
    \item Gradient-based: Gradient, Gradient$\times$Input, IntegratedGradient with \code{n = 10} and a normally distributed reference value and ExpectedGradient with $20$ normally distributed reference values and $10$ samples.
    \item LRP: simple rule, $\varepsilon$-rule with $\varepsilon = 0.01$, $\alpha$-$\beta$-rule with $\alpha = 2$.
    \item DeepLift: Rescale and RevealCancel rule with a normally distributed reference value.
    \item DeepSHAP with $20$ normally distributed reference values.
\end{itemize}

In addition to the total time required by a method of one of the considered packages for generating an explanation, the pure execution time is also measured separately. How exactly the time measurement of each method of the packages is accomplished can be found in the reproduction material or in the GitHub repository at \url{https://github.com/bips-hb/JSS_innsight}.

\subsubsection{Number of output nodes (C)}

In the time analysis regarding the number of output nodes, $20$ dense and image models of each of the architecture shown in Figure~\ref{fig:val_time_models} are created for $C = 1, 2, 4,6, \ldots, 20$. The default values are set for the other hyperparameters, i.e., $W = 64$, $U = 128$ for the tabular models, $U = 5$ for image models, $L = 2$, and $B = 16$. Since \pkg{innsight} is explicitly designed to analyze multiple output nodes or output classes simultaneously, it is only possible to generate the results in one run with \pkg{innsight}. All other implementations are forced to perform a for-loop over the output nodes.

\subsubsection{Number of layers (L)}

In the time analysis regarding the number of hidden layers, $20$ dense and image models of each of the architecture shown in Figure~\ref{fig:val_time_models} are created for $L = 1, 2, 4,6, \ldots, 20$. The default values are set for the other hyperparameters, i.e., $C = 1$, $W = 64$, $U = 128$ for the tabular models and $U = 5$ for image models, and $B = 16$.

\subsubsection{Number of hidden units/filters (U)}

In the time analysis regarding the number of hidden units for dense models and number of filters for image models, $20$ models each of the architecture shown in Figure~\ref{fig:val_time_models} are created for $U = 128, 256, 512,768, \ldots, 2560$ for the tabular and $U=10, 50, 100,150, \ldots, 500$ for the image model. The default values are set for the other hyperparameters, i.e., $C = 1$, $W = 64$, $L = 2$, and $B = 16$.

\subsubsection{Batch size (B)}

In the time analysis regarding the number of input instances, $20$ dense and image models of each of the architecture shown in Figure~\ref{fig:val_time_models} are created for $B = 32, 64, 128, 192, \ldots, 640$ for tabular and $B = 16, 32, 64, 96, \ldots, 320$ for image models. The default values are set for the other hyperparameters, i.e., $C = 1$, $W = 64$, $U = 128$ for the tabular models and $U = 5$ for image models, and $L = 2$.

\subsubsection{Height and width of the image inputs (W)}

In the time analysis regarding the height/width of the image inputs, $20$ image models for each of the architecture shown in Figure~\ref{fig:val_time_models} are created for $W = 16, 32,64, 96, \ldots, 320$. The default values are set for the other hyperparameters, i.e., $C = 1$, $U = 128$ for the tabular models and $U = 5$ for image models, and $L = 2$.

\newpage

\subsection{Additional figures} \label{app:additional_figs}

\subsubsection{Number of output nodes (C)}

\begin{figure}[!h]
    \centering
    \resizebox{\textwidth}{!}{
    \includegraphics{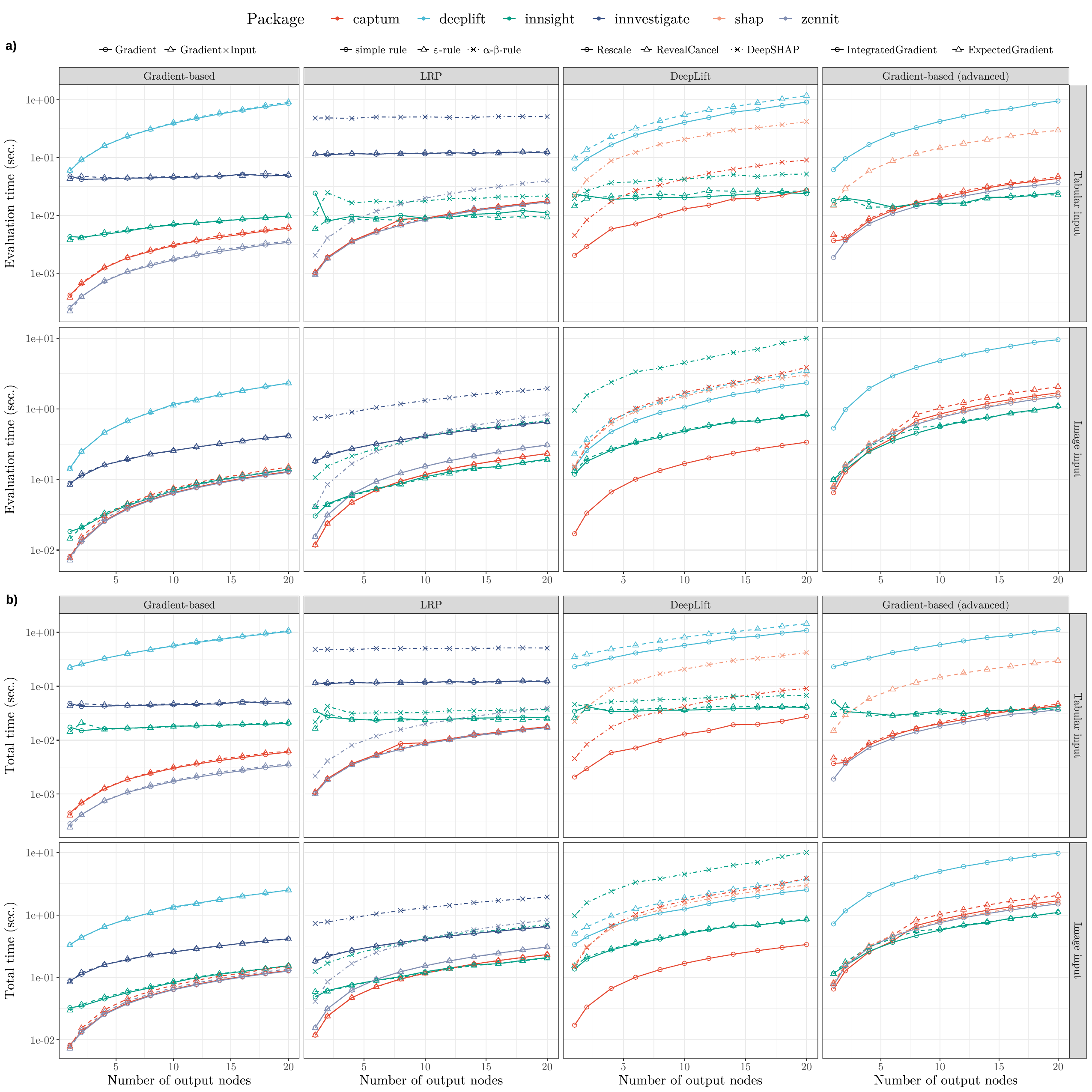}
    }
    \caption{Package’s average a) evaluation and b) total runtime in seconds over $20$ repetitions for applying different feature attribution methods on models with a varying number of output nodes.}
    \label{fig:val_time_C}
\end{figure}

\newpage

\subsubsection{Number of layers (L)}

\begin{figure}[!h]
    \centering
    \resizebox{\textwidth}{!}{
    \includegraphics{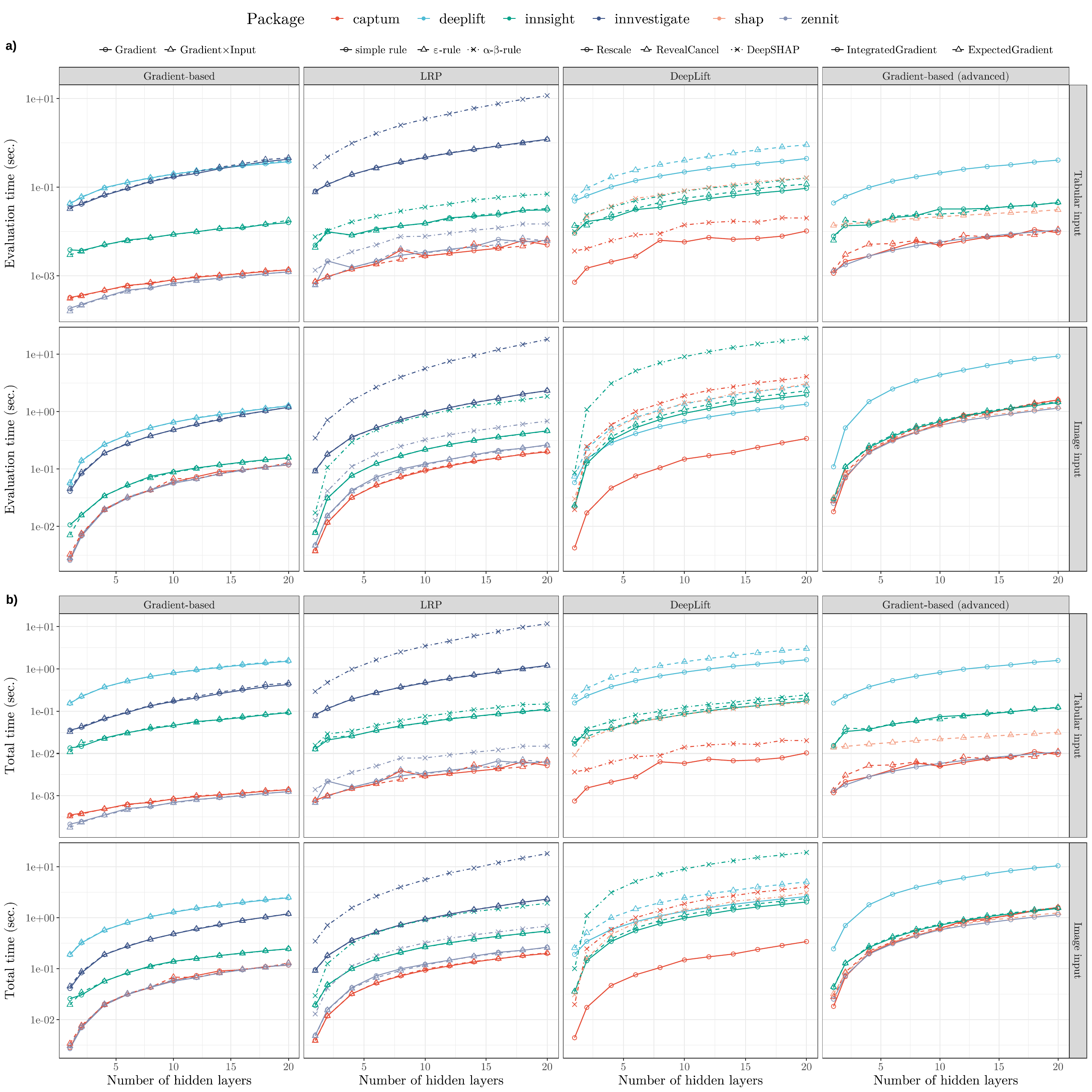}
    }
    \caption{Package’s average a) evaluation and b) total runtime in seconds over $20$ repetitions for applying different feature attribution methods on models with a varying number of hidden layers.}
    \label{fig:val_time_L}
\end{figure}

\newpage

\subsubsection{Number of hidden units/filters (U)}

\begin{figure}[!h]
    \centering
    \resizebox{\textwidth}{!}{
    \includegraphics{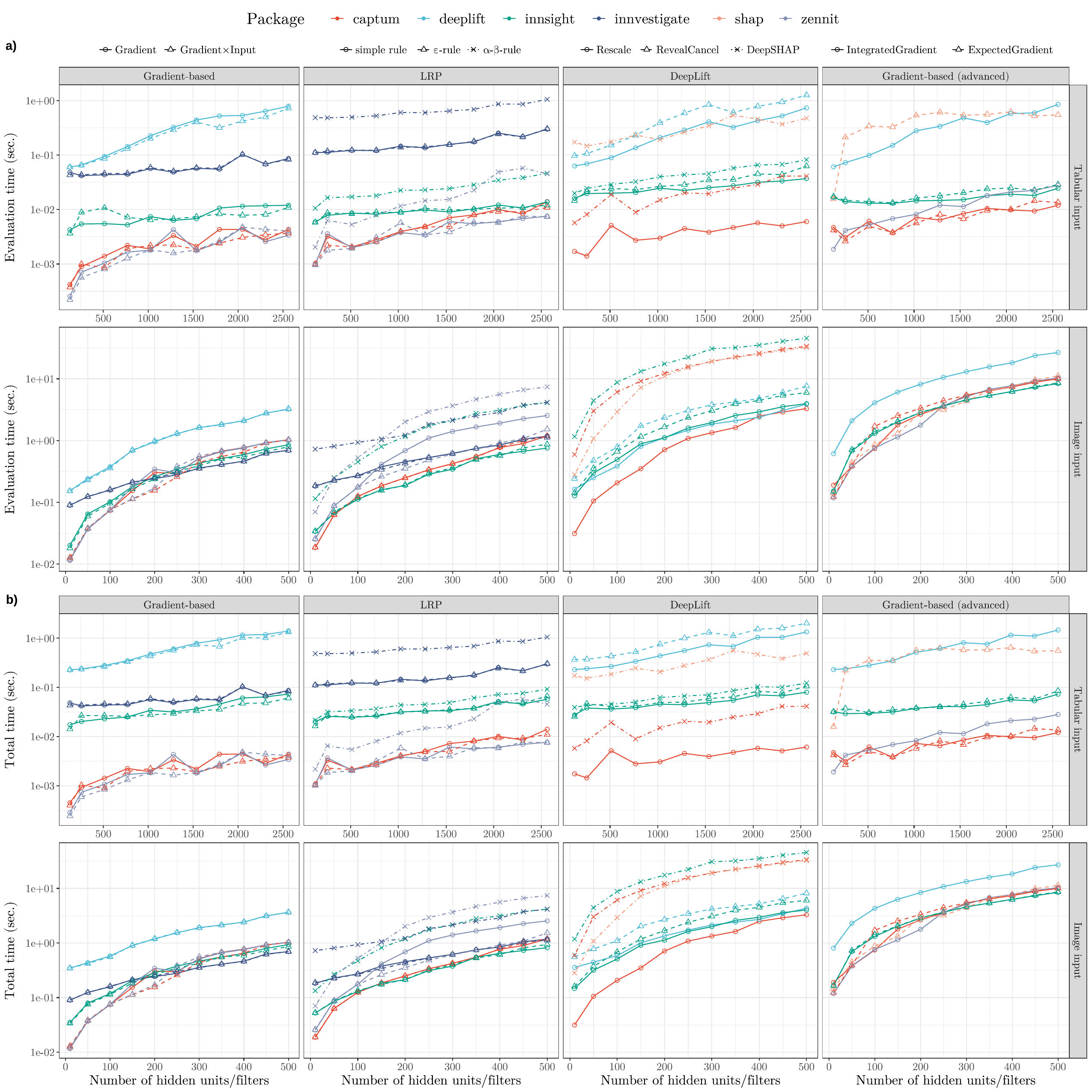}
    }
    \caption{Package’s average a) evaluation and b) total runtime in seconds over $20$ repetitions for applying different feature attribution methods on models with a varying number of hidden units/filters.}
    \label{fig:val_time_U}
\end{figure}

\newpage

\subsubsection{Batch size (B)}

\begin{figure}[!h]
    \centering
    \resizebox{\textwidth}{!}{
    \includegraphics{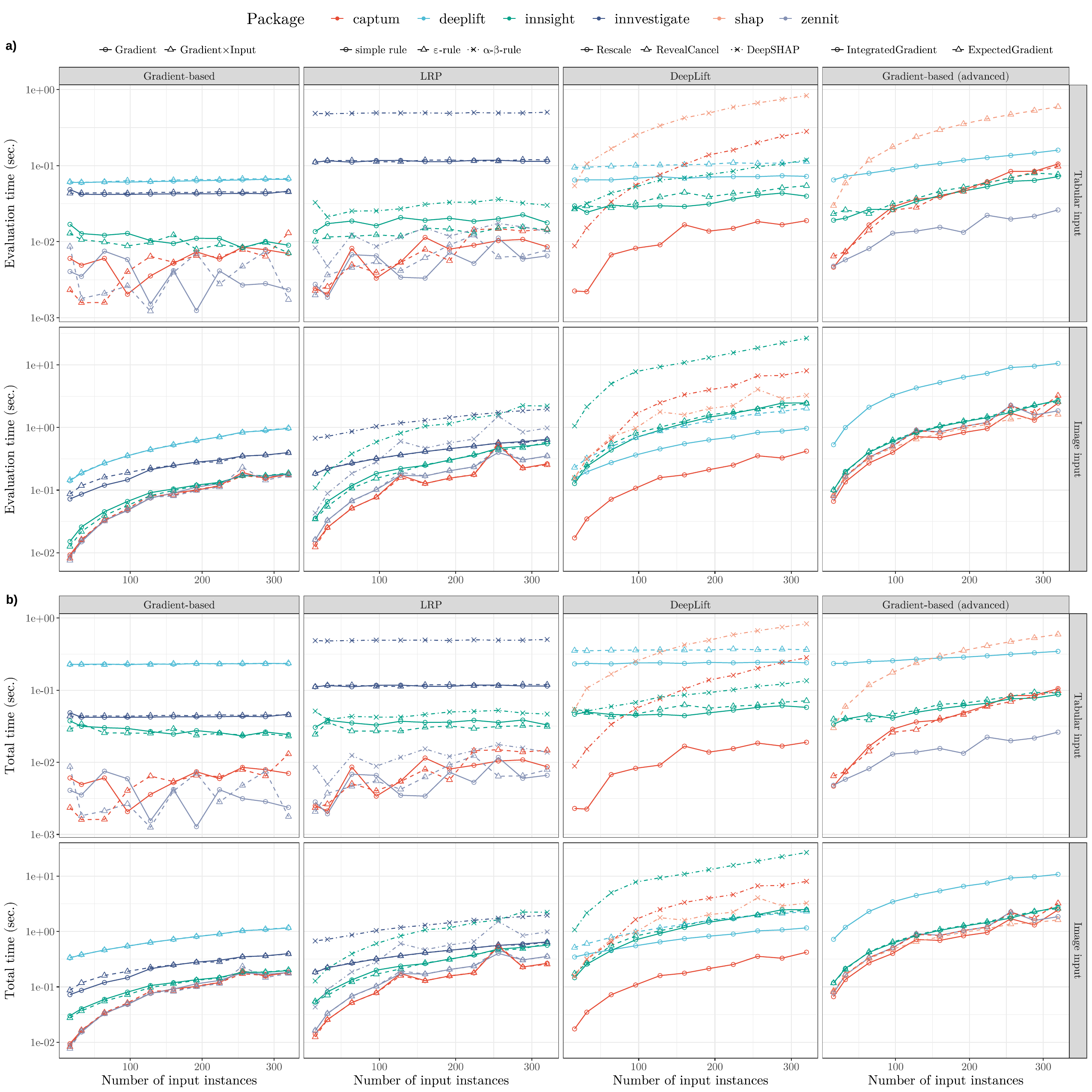}
    }
    \caption{Package’s average a) evaluation and b) total runtime in seconds over $20$ repetitions for applying different feature attribution methods on models with a varying batch size.}
    \label{fig:val_time_B}
\end{figure}

\newpage

\subsubsection{Height and width of the image inputs (W)}

\begin{figure}[!h]
    \centering
    \resizebox{\textwidth}{!}{
    \includegraphics{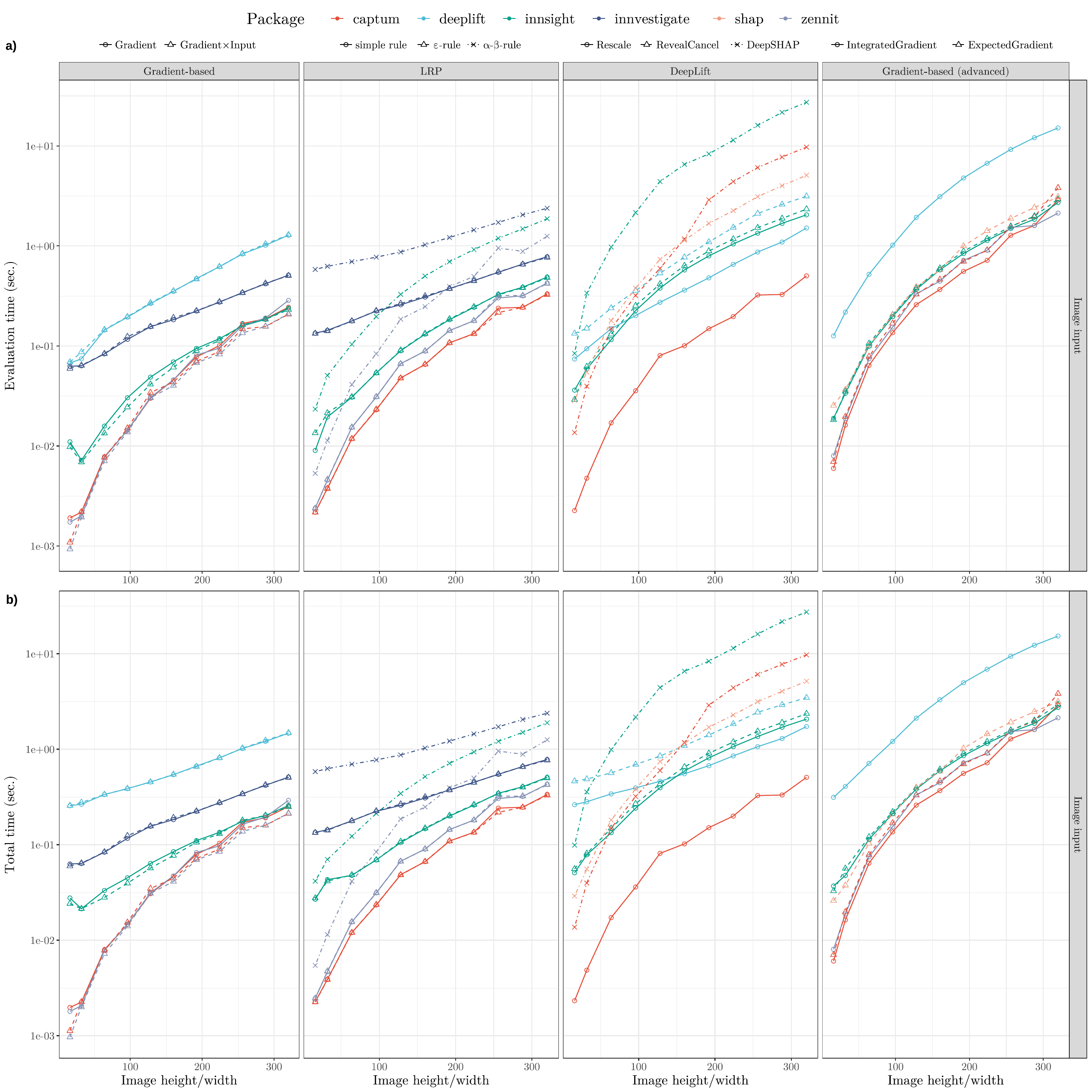}
    }
    \caption{Package’s average a) evaluation and b) total runtime in seconds over $20$ repetitions for applying different feature attribution methods on models with a varying input image size.}
    \label{fig:val_time_W}
\end{figure}

\newpage

\section[LRP with bias for innvestigate]{LRP with bias for \pkg{innvestigate}}\label{app:lrp_bias}

As already observed in Section~\ref{sec:validation}, the results of \pkg{innsight} and \pkg{innvestigate} differ significantly in the LRP method with the $\alpha$-$\beta$-rule for models with a bias vector. In the following, this problem is analyzed using a very simple neural network with only one dense layer consisting of one input variable and one output node. This layer has a weight of $w = 1$, a bias vector of $b = -0.25$, and is applied to the input $x = 1$. Mathematically, this results in the following input relevance for $x$:
\begin{align*}
    R_x &= \left(\alpha \frac{(x w)^+}{(x w)^+ + (b)^+} + \beta \frac{(x w)^-}{(x w)^- + (b)^-} \right) \hat{y}\\
    &= \left(\alpha \frac{(1)^+}{(1)^+ + (-0.25)^+} + \beta \frac{(1)^-}{(1)^- + (-0.25)^-} \right) 0.75\\
    &= \left(\alpha \cdot 1 + \beta \cdot 0 \right) 0.75 = 0.75\alpha
\end{align*}
This yields, for example, in a relevance of $R_x=0.75$ for the $\alpha$-$\beta$-rule with $\alpha = 1$ and $R_x = 1.5$ for $\alpha = 2$, which are exactly the values that \pkg{innsight} outputs. The package \pkg{innvestigate}, on the other hand, outputs relevance $1$ and $2$, which is probably because the bias vector $b$ is included in the calculation of the positive part despite the negative sign. This short comparison can be reproduced with the reproduction material or in our GitHub repository (\url{https://github.com/bips-hb/JSS_innsight}) and is based on version 2.0.2 of \pkg{innvestigate}.

\end{appendix}


\end{document}

%% file: tikz/feature_attribution.tikz
\begin{tikzpicture}
		%
		%

		\draw[draw = BIPSBlue, fill = BIPSBlue!20, rounded corners] (-1.6, -1.2) rectangle (7, 3.5);
		\begin{scope}
		    \clip (-1.65, 2.9) rectangle (7.05, 3.55);
		    \draw[draw = BIPSBlue, fill = BIPSBlue!50, rounded corners] (-1.6, -1.2) rectangle (7, 3.5);
		\end{scope}
		\node[scale = 1.25] at (2.7, 3.2) {\textbf{Forward pass}};


		\draw[draw = BIPSBlue, fill = BIPSBlue!20, rounded corners] (11.2, -1.2) rectangle (19.9, 3.5);
		\begin{scope}
		    \clip (11.15, 2.9) rectangle (19.95, 3.55);
		    \draw[draw = BIPSBlue, fill = BIPSBlue!50, rounded corners] (11.2, -1.2) rectangle (19.9, 3.5);
		\end{scope}
		\node[scale = 1.25] at (15.5, 3.2) {\textbf{Backward pass}};

		%
		%

		\draw[draw=gray!10, fill = gray!10, rounded corners] (-1.4,-1) rectangle (2.5,2.7);
		\draw[draw=gray!20, fill = gray!20, rounded corners] (-1.4,-1) rectangle (1,2.7); 
		\draw[draw=gray!40, fill = gray!40, rounded corners] (-1.4,-1) rectangle (-0.9,2.7);

		\node at (0,1.675) (image) {\includegraphics[width=1.5cm]{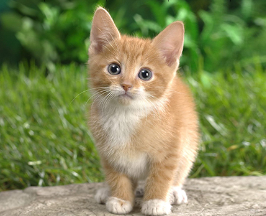}};
		\node[inner sep=0pt] (table) at (0,0) {\includegraphics[width=1.65cm]{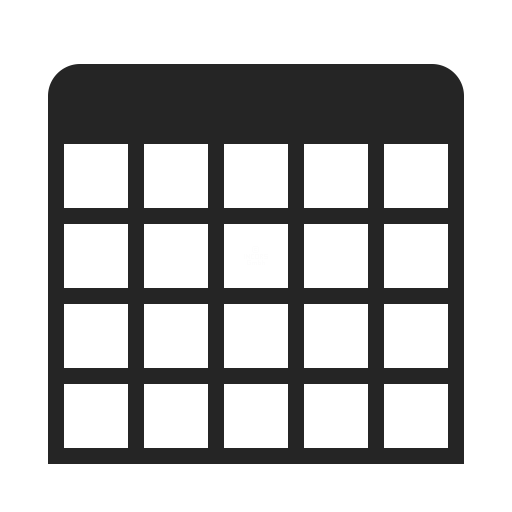}};

		\node[rotate = 90] at (-1.15, 1.75) {Image};
		\node[rotate = 90] at (-1.15,0) {Tabular};
		\node at (1.85, 2.2) {$\mathbf{x} \in \mathbb{R}^p$};
		\node at (1.75, 0.875) {$\begin{pmatrix} x_1\\ \vdots \\ x_p \end{pmatrix}$};

		%
		%
		\draw[draw=black, fill = black, rounded corners] (4, 0.25) rectangle (6.5,1.5) node[pos=.5, color = white, align=center, scale = 0.8] (NN) {{\large $f(\mathbf{x})$} \\ Neural network};
		\draw[->, thick] (2.55, 0.875) -- (3.95, 0.875);
		\draw[->, thick] (6.55, 0.875) -- (8.45, 0.875); 

		%
		%

		\draw[draw=gray!10, fill = gray!10, rounded corners] (8.5,-1) rectangle (9.8,2.7);
  		\draw[darkgray, fill = gray!40, rounded corners] (8.5,0.1) rectangle (9.8, 0.6);

		\node[scale = 1] at (9.15, 3) {\textbf{Prediction}};
		\node at (9.2, 2.2) {$\bm{\hat{y}} \in \mathbb{R}^C$};
		\node at (9.15, 0.35) {$\begin{pmatrix} \hat{y}_1 \\ \vdots \\ \hat{y}_c \\ \vdots \\ \hat{y}_C\end{pmatrix}$};

		%
		%
		\draw[draw=black, fill = white, rounded corners] (11.8, 0.25) rectangle (14.7,1.5) node[pos=.5, color = black, align=center, scale = 0.8] (FA) {Feature attribution \\ method};
		\draw[->, thick] (9.85, 0.35) -- (11.75, 0.875);
		\draw[->, thick] (14.75, 0.875) -- (15.75, 0.875);

		%
		%

		\draw[draw=gray!10, fill = gray!10, rounded corners] (15.8,-1) rectangle (19.7,2.7);
		\draw[gray!20, fill=gray!20, rounded corners] (17.1, -1) rectangle (19.7, 2.7);
		\draw[draw=gray!40, fill = gray!40, rounded corners] (19.2,-1) rectangle (19.7,2.7);

		\def\x{0.1875}
		\node at (18.15,1.675) {\includegraphics[width = 1.5cm]{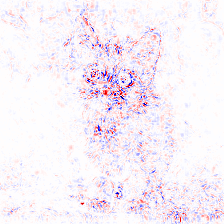}};

		\draw[->] (17.4, 0) -- (19, 0);
		\draw[->] (17.4, -0.75) -- (17.4, 0.75);
		\draw[fill = red!28.5] (17.4 + \x / 2, 0) rectangle (17.4 + \x * 1.5, 0.2);
		\draw[fill = blue!14] (17.4 + \x * 2, 0) rectangle (17.4 + \x * 3, -0.1);
		\draw[fill = blue!71] (17.4 + \x * 3.5, 0) rectangle (17.4 + \x * 4.5, -0.5);
		\draw[fill = red!85] (17.4 + \x * 5, 0) rectangle (17.4 + \x * 6, 0.6);
		\draw[fill = red!62] (17.4 + \x * 6.5, 0) rectangle (17.4 + \x * 7.5, 0.44);

		\node[rotate = 90] at (19.45, 1.75) {Image};
		\node[rotate = 90] at (19.45, 0) {Tabular};
		\node[scale = 0.9] at (16.45, 2.2) {$\bm{R^c} \in \mathbb{R}^p$};
		\node at (16.45, 0.875) {$\begin{pmatrix} R^c_1\\ \vdots \\ R^c_p \end{pmatrix}$};
	\end{tikzpicture}

%% file: tikz/gradientbased.tikz
\begin{tikzpicture}[x = 1cm, y = 1.05cm]
	\draw[BIPSBlue!15, fill = BIPSBlue!15, rounded corners] (-3,1) rectangle (3.9,-1.3);
	
	\draw[BIPSBlue!30, fill = BIPSBlue!30, rounded corners] (-2.9,0.9) rectangle (-1.2,-1.2);
	\node[fill = BIPSBlue!50, rounded corners, text width = 0.25\textwidth, text height = 0.4cm, align = center, scale = 0.35] at (-2.05, 0.15) {
		\textbf{Gradient}\vspace{-0.1cm}
		\begin{equation*}
				\frac{\partial f(\bm{x})_c}{x_i}
		\end{equation*}};
	\node[fill = BIPSBlue!50, rounded corners, text width = 0.25\textwidth, text height = 0.4cm, align = center, scale = 0.35] at (-2.05, -0.7) {
		\textbf{SmoothGrad}\vspace{-0.1cm}
		\begin{equation*} \mathbb{E}_\varepsilon \left[\frac{\partial f(\bm{x} + \bm{\varepsilon})_c}{\partial x_i + \varepsilon_i} \right]
	\end{equation*}};

	\draw[BIPSBlue!30, fill = BIPSBlue!30, rounded corners] (-1,0.9) rectangle (1 ,-1.2);
	\node[fill = BIPSBlue!50, rounded corners, text height = 0.4cm, text width = 0.3\textwidth, align = center, scale = 0.35] (A) at (0, 0.15) {
		\textbf{Grad$\times$Input}\vspace{-0.1cm}
		\begin{equation*}
			\frac{\partial f(\bm{x})_c}{x_i} \cdot x_i
	\end{equation*}};
	\node[fill = BIPSBlue!50, rounded corners, text height = 0.4cm, text width = 0.3\textwidth, align = center, scale = 0.35] (B) at (0, -0.7) {
		\textbf{SmoothGrad$\times$Input}\vspace{-0.1cm}
		\begin{equation*}
			 \mathbb{E}_\varepsilon \left[\frac{\partial f(\bm{x} + \bm{\varepsilon})_c}{\partial x_i + \varepsilon_i}\ (x_i + \varepsilon_i) \right]
	\end{equation*}};

	\draw[BIPSBlue!30, fill = BIPSBlue!30, rounded corners] (1.2,0.9) rectangle (3.8 ,-0.1);
	\node[fill = BIPSBlue!50, rounded corners, text width = 0.43\textwidth, text height = 0.4cm, align = center, scale = 0.35] (A) at (2.5, 0.28) {
		\textbf{IntegratedGradient}\vspace{-0.1cm}
		\begin{equation*}
			(x_i - x_i') \int_{\alpha = 0}^1 \frac{\partial f(\bm{\tilde{x}} + \alpha ( \bm{x} - \bm{\tilde{x}}))_c}{\partial x_i} d \alpha
	\end{equation*}};

	\draw[BIPSBlue!30, fill = BIPSBlue!30, rounded corners] (1.2,-0.2) rectangle (3.8 ,-1.2);
	\node[fill = BIPSBlue!50, rounded corners, text width = 0.43\textwidth, text height = 0.4cm, align = center, scale = 0.35] (A) at (2.5, -0.8) {
		\textbf{ExpectedGradient}\vspace{-0.1cm}
		\begin{equation*}
			\underset{\substack{\bm{\tilde{x}}\sim D\\ \alpha \sim \mathcal{U}(0,1)}}{\scalebox{1.6}{$\mathbb{E}$}} \left[(x_i - \tilde{x}_i) \frac{\partial f(\bm{\tilde{x}} + \alpha ( \bm{x} - \bm{\tilde{x}}))_c}{\partial x_i} \right]
	\end{equation*}};

	\node[scale = 0.4, align= center] at (-2.05, 0.7) {\textbf{Output sensitivity}};
	\node[scale = 0.4] at (0, 0.7) {\textbf{Decomposition of} $f(\bm{x})_c$};
	\node[scale = 0.4] at (2.5, 0.7) {\textbf{Decomposition of} $f(\bm{x})_c - f(\bm{\tilde{x}})_c$};
	\node[scale = 0.38] at (2.5, -0.325) {\textbf{Decomposition of} $f(\bm{x})_c - \mathbb{E}_{\bm{\tilde{x}}} \left[ f(\bm{\tilde{x}})_c \right]$};
	
\end{tikzpicture}

%% file: tikz/lrp.tikz
\begin{tikzpicture}
    	\def\yshift{0.45}
    	\def\lwidth{1.75}
    	\tikzset{>={Latex[width=5,length=5]}}
    	\tikzstyle{node}=[very thick,circle,minimum size=22,inner sep=0.5,outer sep=0.6]
        \tikzstyle{connect}=[->,thick,black!60,shorten >=1]
        \tikzset{
        	input/.style={node, draw=BIPSBlue, fill=BIPSBlue!45},
        	input_b/.style={node, minimum size =18.5, draw=BIPSBlue, pattern=mydots, pattern color=BIPSBlue!25},
        	hidden/.style={node, minimum size=15, draw=gray, fill=gray!25},
        	output/.style={node, draw=red!80!black, fill=red!80!black!25},
        }

        \pgfdeclarepatternformonly{mydots}{\pgfqpoint{-1pt}{-1pt}}{\pgfqpoint{1pt}{1pt}}{\pgfqpoint{1.5pt}{1.5pt}}
        {%
        	\pgfpathcircle{\pgfqpoint{0pt}{0pt}}{.5pt}%
        	\pgfusepath{fill}%
        }%
    	
    	
    	\begin{scope}[scale = 1.5, transform shape, x = 2.5cm, y = 1.5cm]
    	\begin{scope}
    	\clip (0.79,-2.01) rectangle (3.01, 1.2);
    	\draw[gray, rounded corners, fill = gray!20] (0.8,-2.0) rectangle (3, 1.55); 
    	\end{scope}
    	\begin{scope}
    	\clip (0.79,0.9) rectangle (3.01, 1.31);
    	\draw[draw = gray, fill = gray!40, rounded corners] (0.8,-3.25) rectangle (3, 1.3); 
    	\end{scope}
    	\node[input_b, pattern=mydots, line width = \lwidth] (N1-1) at (1.5,4/2 - 1.4) {};
    	\node[input, line width = \lwidth] (N1-2) at (1.1,4/2 - 2) {$R_1^{l}$};
    	\node[input, line width = \lwidth] (N1-4) at (1.1,4/2 - 3 - \yshift) {$R_i^{l}$};
    	\node[hidden, line width = \lwidth] (N2-1) at (2,4/2 - 2) {};
    	\node[hidden, line width = \lwidth] (N2-3) at (2,4/2 - 3 - \yshift) {};
    	\node[output, line width = \lwidth] (N3-1) at (2.65,4/2 - 2) {$R_1^{l+1}$};
    	\node[output, line width = \lwidth] (N3-3) at (2.65,4/2 - 3 - \yshift) {$R_j^{l+1}$};
    	\path (N1-4) --++ (0,0.9+\yshift) node[midway,scale=1.5] {$\vdots$};
    	\path (N2-3) --++ (0,0.9+\yshift) node[midway,scale=1.5] {$\vdots$};
    	\path (N3-3) --++ (0,0.9+\yshift) node[midway,scale=1.5] {$\vdots$};
    	\draw[white,line width=1.2, shorten <=1] (N1-1) -- (N2-1);
    	\draw[<-, red!70, line width = 1, shorten <= 1, >={Latex[width=4,length=4]}] (N1-1) -- (N2-1);
    	\draw[white,line width=0.6, shorten >=1] (N1-1) -- (N2-3);
    	\draw[<-, red!70, line width = 0.5, shorten >= 1, >={Latex[width=2,length=2]}] (N1-1) -- (N2-3);
    	\draw[white,line width=6.2, shorten <=5] (N1-2) -- (N2-1);
    	\draw[<-, red!70, line width = 6, shorten <= 1, >={Latex[width=9,length=6]}] (N1-2) -- (N2-1) node[black, pos = 0.5, sloped, yshift = 0.75em] {$r_{1 \leftarrow 1}$};
    	\draw[white,line width=1.2, shorten >=1] (N1-2) -- (N2-3);
    	\draw[<-, red!70, line width = 1, shorten >= 1, >={Latex[width=4,length=4]}] (N1-2) -- (N2-3) node[black, pos = 0.65, sloped, yshift = 0.7em] {$r_{1 \leftarrow j}$};
    	\draw[white,line width=3.2, shorten <=3] (N1-4) -- (N2-1);
    	\draw[<-, red!70, line width = 3, shorten <= 1, >={Latex[width=6,length=5]}] (N1-4) -- (N2-1) node[black, pos = 0.35, sloped, yshift = 0.65em] {$r_{i \leftarrow 1}$};
    	\draw[white,line width=0.6, shorten >=1] (N1-4) -- (N2-3);
    	\draw[<-, red!70, line width = 0.5, shorten >= 1, >={Latex[width=2,length=2]}] (N1-4) -- (N2-3) node[black, pos = 0.5, sloped, yshift = 0.4em] {$r_{i \leftarrow j}$};
    	\draw[white,line width= 5.2, shorten <=5] (N2-1) -- (N3-1);
    	\draw[red!70, <-, line width = 10, shorten >= 1, >={Latex[width=18,length=12]}] (N2-1) -- (N3-1);
    	\draw[white,line width=2.2,shorten >=1] (N2-3) -- (N3-3);
    	\draw[<-, red!70, line width = 2, shorten >= 1, >={Latex[width=8,length=7]}] (N2-3) -- (N3-3);
    	\node[scale = 1.3] at (1.875,1.075) {Hidden layer};
    	\draw[BIPSBlue!75, line width = 8, <-, >={Latex[width=20,length=10]}] (0.1,1.8) -- (3.6, 1.8) node[pos = 0.5, fill = white, scale = 1.5] {Backward};
    	\end{scope}

    	\begin{scope}[scale = 1.2, transform shape, x = 2.5cm, y = 1.5cm]
    	\draw[BIPSBlue, fill = BIPSBlue!20, rounded corners] (0, -2.5 - \yshift) rectangle (0.5, 1.5);
    	\draw[red, fill = red!20, rounded corners] (4.25, -1 - 0.5 * \yshift) rectangle (4.75, -0.5 * \yshift);
    	
    	\node[BIPSBlue] at (0.25, 2 - 0.25) {\Large \textbf{Input}};
    	\node[input, line width = \lwidth] (N0-1) at (0.25, 4/2 - 1) {$R_1^1$};
    	\node[input, line width = \lwidth] (N0-2) at (0.25, 4/2 - 2) {$R_2^1$};
    	\node[input, line width = \lwidth] (N0-3) at (0.25, 4/2 - 3) {$R_3^1$};
    	\node[input, line width = \lwidth] (N0-4) at (0.25, 4/2 - 4 - \yshift) {$R_d^1$};
    	\path (N0-4) --++ (0,0.9+\yshift) node[midway,scale=1.9] {$\vdots$};
    	
    	\node[red] at (4.5, 0.0) {\Large \textbf{Output}};
    	\node[output, line width = \lwidth] (out) at (4.5, -0.5 - 0.5 * \yshift) {\large $\hat{y}_c$};
    	
    	\draw[white,line width=0.4 * \lwidth,shorten >=3] (N0-1) -- (N1-2);
    	\draw[red!70, <-, line width = \lwidth] (N0-1) -- (N1-2) node[pos = 0.5, sloped, fill = white] {$\cdots$};
    	\draw[white,line width=0.4 * \lwidth,shorten >=4] (N0-1) -- (N1-4);
    	\draw[red!70, <-, line width = 1.5 * \lwidth] (N0-1) -- (N1-4) node[pos = 0.5, sloped, fill = white] {$\cdots$};
    	
    	\draw[white,line width=0.6 * \lwidth,shorten >=3] (N0-2) -- (N1-2);
    	\draw[red!70, <-, line width = 3* \lwidth, >={Latex[width=8,length=8]}] (N0-2) -- (N1-2) node[pos = 0.5, sloped, fill = white] {$\cdots$};
    	\draw[white,line width=0.2 * \lwidth,shorten >=3] (N0-2) -- (N1-4);
    	\draw[red!70, <-, line width = 0.5 * \lwidth] (N0-2) -- (N1-4) node[pos = 0.5, sloped, fill = white] {$\cdots$};
    	
    	\draw[white,line width=0.4 * \lwidth,shorten >=3] (N0-3) -- (N1-2);
    	\draw[red!70, <-, line width = \lwidth] (N0-3) -- (N1-2) node[pos = 0.5, sloped, fill = white] {$\cdots$};
    	\draw[white,line width=0.4 * \lwidth,shorten >=3] (N0-3) -- (N1-4);
    	\draw[red!70, <-, line width = \lwidth] (N0-3) -- (N1-4) node[pos = 0.5, sloped, fill = white] {$\cdots$};
    	
    	\draw[white,line width=0.7 * \lwidth,shorten >=3] (N0-4) -- (N1-2);
    	\draw[red!70, <-, line width = 2 *\lwidth, >={Latex[width=6,length=6]}] (N0-4) -- (N1-2) node[pos = 0.5, sloped, fill = white] {$\cdots$};
    	\draw[white,line width=0.2 * \lwidth,shorten >=3] (N0-4) -- (N1-4);
    	\draw[red!70, <-, line width = 0.5 *\lwidth] (N0-4) -- (N1-4) node[pos = 0.5, sloped, fill = white] {$\cdots$};
    	
    	\draw[white,line width=2.2,shorten >=1] (N3-1) -- (out);
    	\draw[<-, red, line width = 10, >={Latex[width=18,length=12]}] (N3-1) -- (out) node[pos = 0.5, sloped, fill = white] {$\cdots$};
    	\draw[white,line width=2.2, shorten >=1] (N3-3) -- (out);
    	\draw[<-, red, line width = 2] (N3-3) -- (out) node[pos = 0.5, sloped, fill = white] {$\cdots$};
    	
    	\end{scope}
    	\end{tikzpicture}

%% file: tikz/lrp_detail.tikz
\def\layersep{2cm}
    \begin{tikzpicture}[shorten >=1pt,draw=black!50, node distance=\layersep]
		\tikzstyle{node}=[very thick,circle,minimum size=22,inner sep=0.5,outer sep=0.6]
		\tikzset{
			input/.style={node, draw=BIPSBlue, fill=BIPSBlue!45},
			input_b/.style={node, minimum size =18.5, draw=BIPSBlue, pattern=mydots, pattern color=BIPSBlue!25},
			hidden/.style={node, minimum size=15, draw=gray, fill=gray!25},
			output/.style={node, draw=red!80!black, fill=red!80!black!25},
		}
		
		\draw [gray, fill = gray!20, rounded corners] (-0.55, -2.5) rectangle (\layersep * 1.75, -8.25); 
	
		\node[input] (I-1) at (0,-1*2 -1) {$x_1$};
		\node[xshift = -2.8cm, yshift = -0.15cm, scale = 0.8] at (I-1) {$R_1^{l} =  \underbrace{\frac{x_1 w_{11}}{z_1} R_1^{l+1}}_{r_{1 \leftarrow 1}^{(l, l + 1)}} + \underbrace{\frac{x_1 w_{12}}{z_2} R_2^{l+1}}_{r_{1 \leftarrow 2}^{(l, l + 1)}}$};
		\node[input] (I-2) at (0,-2*2 -1) {$x_2$};
		\node[xshift = -2.8cm, yshift = -0.15cm, scale = 0.8] at (I-2) {$R_2^{l} =  \underbrace{\frac{x_2 w_{21}}{z_1} R_1^{l+1}}_{r_{2 \leftarrow 1}^{(l, l + 1)}} + \underbrace{\frac{x_2 w_{22}}{z_2} R_2^{l+1}}_{r_{2 \leftarrow 2}^{(l, l + 1)}}$};
		\node[input] (I-3) at (0,-3*2 -1) {$x_3$};
		\node[xshift = -2.8cm, yshift = -0.15cm, scale = 0.8] at (I-3) {$R_3^{l} =  \underbrace{\frac{x_3 w_{31}}{z_1} R_1^{l+1}}_{r_{3 \leftarrow 1}^{(l, l + 1)}} + \underbrace{\frac{x_3 w_{32}}{z_2} R_2^{l+1}}_{r_{3 \leftarrow 2}^{(l, l + 1)}}$};

		\foreach \name / \y in {1,...,2}
		\node[hidden] (H-\name) at (\layersep * 0.8,-\y*2 -1*2) {$z_\y$};
		
		\foreach \name / \y in {1,...,2}
		\node[output] (O-\name) at (\layersep * 1.5,-\y*2 -1*2) {$y_\y$};
		
		\node[xshift = 1.2cm, scale = 1.3] at (O-1) {$R_1^{l+1}$};
		\node[xshift = 1.2cm, scale = 1.3] at (O-2) {$R_2^{l+1}$};
		
		\draw (H-1) -- (O-1) node[midway, above] {\resizebox{0.5cm}{!}{\begin{tikzpicture}
				\draw[-, line width = 2pt] (-1,0) -- (0,0) -- (0.70710678118,0.70710678118);
				\end{tikzpicture}}};
		\draw (H-2) -- (O-2) node[midway, above] {\resizebox{0.5cm}{!}{\begin{tikzpicture}
				\draw[-, line width = 2pt] (-1,0) -- (0,0) -- (0.70710678118,0.70710678118);
				\end{tikzpicture}}};
		
		\node[scale = 0.9] at (\layersep * 0.8, -7.85) {$z_j = \sum_{i=1}^4 w_{ij} x_i + b_j$};
		
		\foreach \source in {1,...,3}
		\draw (I-\source) -- (H-2) node[above, pos = \source / 4, sloped] {$w_{\source 2}$};
		
		\foreach \source in {1,...,3}
		\draw (I-\source) -- (H-1) node[above, pos = 1 / 4, sloped] {$w_{\source 1}$};	
\end{tikzpicture}

%% file: tikz/innsight.tikz
\begin{tikzpicture}
		\draw[BIPSBlue!20, very thick, fill = BIPSBlue!20, rounded corners] (0,0) rectangle (4,-3);
		\node[BIPSBlue, scale = 0.7] at (2, -0.4) {\Large \texttt{innsight}};
		\draw[torchorange, very thick, fill = torchorange!20, rounded corners] (0.2,-0.8) rectangle (3.8,-2.8);
		\node[torchorange, scale = 0.7] at (1, -1.8) {\Large \texttt{torch}};
		\draw[gray, very thick, fill = gray!20, rounded corners] (2,-1) rectangle (3.6,-2.6);
		\node[gray, scale = 0.4] at (2.8, -1.4) {\Huge \texttt{C++}};
		\node[gray, scale = 0.7] at (2.8, -1.9) {\large \texttt{LibTorch}};
\end{tikzpicture}

%% file: tikz/innsight_intern.tikz
\begin{tikzpicture}
    \tikzset{>=latex}
	\draw[gray!80, fill = gray!10, rounded corners] (-2.5, -0.5) rectangle (2.5, -7);
	\node[black!60] at (-1.5, -0.9) {\code{Converter}};
	
	\draw[->] (0, 0) -- (0, -1.5) node[pos=0.6, scale = 0.6, right, align = center] {conversion\\ method};
	\draw[->] (0, -2.5) -- (0, -3);
	
	\draw[gray!30, rounded corners, fill = gray!30] (0.75, 1.48) rectangle (2.5, 1.02) node[black, pos = 0.6, scale = 0.5] {\code{nn\_sequential}};
	\draw[gray!30, rounded corners, fill = gray!30] (0.75, 0.98) rectangle (2.5, 0.52) node[black, pos = 0.54, scale = 0.5] {\code{keras\_model}};
	\draw[gray!30, rounded corners, fill = gray!30] (0.75, 0.48) rectangle (2.5, 0.02) node[black, pos = 0.49, scale = 0.5] {\code{neuralnet}};
	\draw[BIPSBlue!20, rounded corners, fill = BIPSBlue!20] (-1, 1.5) rectangle (1, 0) node[black, pos = 0.5] {\code{model}};
	
	\draw[gray!30, rounded corners, fill = gray!30] (-1, -1.5) rectangle (1, -2.5) node[black, midway, scale = 0.9] {named list};
	\draw[gray!30, rounded corners, fill = gray!30] (-1.5, -3) rectangle (1.5, -4) node[black, midway, scale = 0.8] {\code{ConvertedModel}};
	
	\draw[BIPSBlue!20, rounded corners, fill = BIPSBlue!20] (-5, -3) rectangle (-3, -4) node[black, pos = 0.5, align = center, scale = 0.9] {\code{model} as \\ named list};
	\draw[->] (-3, -3.5) -- (-1.5, -3.5) node[midway, above, align = center, scale = 0.475] {including checks};
	
	\draw[gray!30, fill = gray!30, rounded corners] (-2.25, -4.5) rectangle (-0.5, -6.7);
	\node[black!60, scale = 0.8] at (-1.375, -4.8) {Fields};
	\draw[gray!75, fill = gray!75, rounded corners] (-2.1, -5.2) rectangle (-0.65, -5.7) node[black, midway, scale = 0.7] {\code{\$model}};
	\draw[gray!75, fill = gray!75, rounded corners] (-2.1, -5.9) rectangle (-0.65, -6.4) node[black, midway, scale = 0.48] {\code{\$model\_as\_list}};
	\draw[->] (0, -4) -- (0, -5.45) -- (-0.65, -5.45);
	\draw[->] (1, -2) to[out = 0, in = 90] (2, -4.1) to[out = -90, in = 0] (0, -6.15) to (-0.65,-6.15);
	\node[rotate = -90, scale = 0.6] at (2.15, -4.1) {if \code{save\_model\_as\_list = TRUE}};
	\draw[->] (-2.9, -3.5) -- (-2.9, -6.15) node[pos = 0.6, below, align = center, scale = 0.4, rotate = -90] {if \code{save\_model\_as\_list = TRUE}} -- (-2.1, -6.15);
\end{tikzpicture}

%% file: tikz/visualization_tools.tikz
\begin{tikzpicture}[x = 1cm, y = 0.8cm]
	
	\draw[BIPSBlue!20, fill = BIPSBlue!20, rounded corners] (-5,0) rectangle (5.75, -12.25);
	\draw[BIPSBlue!20, fill = BIPSBlue!20, rounded corners] (6.25,0) rectangle (17, -12.25);
	
	\node[scale = 1.5] at (0.375, -0.75) {\code{plot()}};
	\node[scale = 1.5] at (11.625, -0.75) {\code{plot_global()} (\code{boxplot()})};
	
	\draw[BIPSBlue!20, fill = white, rounded corners] (-4.6, -2.6) rectangle (5.6, -12.25); 
	\begin{scope}
		\clip (-4.6, -1.5) rectangle (-3.3, -12.25);
		\draw[BIPSBlue!20, fill = BIPSBlue!40, rounded corners] (-4.6, -2.6) rectangle (5.6, -12.25);
	\end{scope}
	\begin{scope}
		\clip (-3.3, -1.5) rectangle (5.6, -2.6); 
		\draw[BIPSBlue!40, fill = BIPSBlue!40, rounded corners] (-3.3, -1.5) rectangle (5.6, -4);
	\end{scope}
	\node[scale = 1] at (-1.1, -2.05) {Tabular/signal};
	\node[scale = 1] at (3.3125, -2.05) {Image};
	\node[rotate = 90, scale = 0.9] at (-3.95, -5) {\code{as_plotly = FALSE}};
	\node[rotate = 90, scale = 0.9] at (-3.95, -9.75) {\code{as\_plotly = TRUE}};
	
	\node at (3.34, -5.025) {\resizebox{0.29\textwidth}{!}{\includegraphics{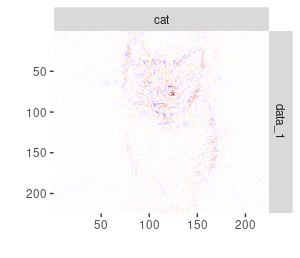}}};
	\node at (-1.05, -5.025) {\resizebox{0.29\textwidth}{!}{\includegraphics{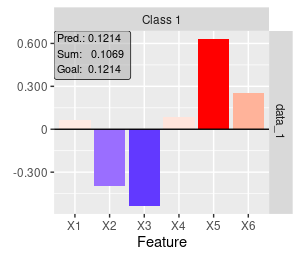}}};
	\node at (3.34, -9.82) {\resizebox{0.29\textwidth}{!}{\includegraphics{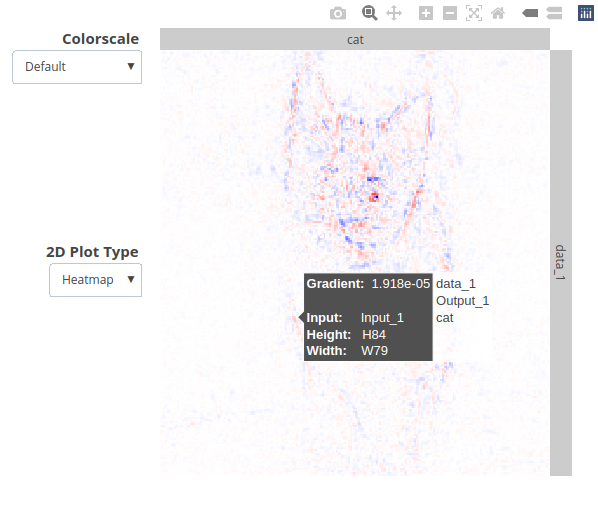}}};
	\node at (-1.05, -9.82) {\resizebox{0.29\textwidth}{!}{\includegraphics{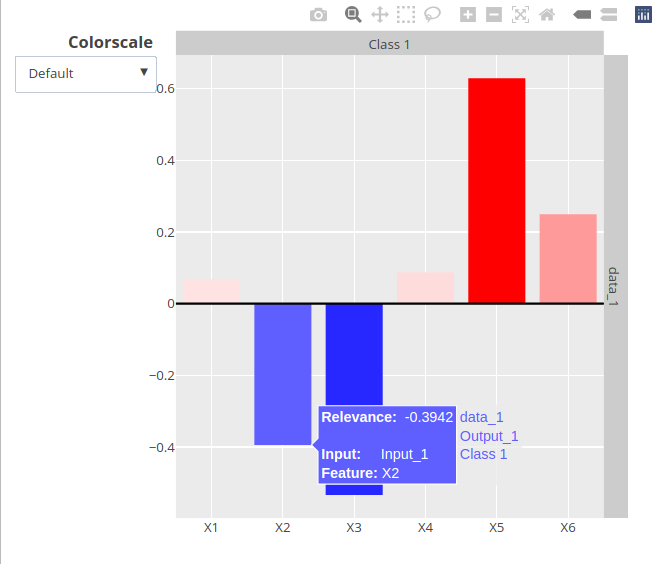}}};

	\begin{scope}[shift={(11.25,0)}]
		\draw[BIPSBlue!40, fill = white, rounded corners] (-4.6, -2.6) rectangle (5.6, -12.24); 
		\begin{scope}
		\clip (-4.6, -1.5) rectangle (-3.3, -12.25);
		\draw[BIPSBlue!40, fill = BIPSBlue!40, rounded corners] (-4.6, -2.6) rectangle (5.6, -12.25);
		\end{scope}
		\begin{scope}
		\clip (-3.3, -1.5) rectangle (5.6, -2.6); 
		\draw[BIPSBlue!40, fill = BIPSBlue!40, rounded corners] (-3.3, -1.5) rectangle (5.6, -4);
		\end{scope}
		\node[scale = 1] at (-1.1, -2.05) {Tabular/signal};
		\node[scale = 1] at (3.3125, -2.05) {Image};
		\node[rotate = 90, scale = 0.9] at (-3.95, -5) {\code{as_plotly = FALSE}};
		\node[rotate = 90, scale = 0.9] at (-3.95, -9.75) {\code{as\_plotly = TRUE}};
		
		\node at (3.33, -5.055) {\resizebox{0.293\textwidth}{!}{\includegraphics{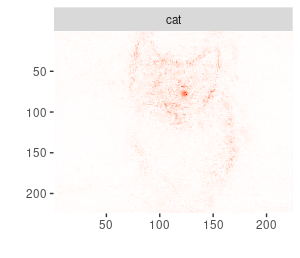}}};
		\node at (-1.025, -5.055) {\resizebox{0.293\textwidth}{!}{\includegraphics{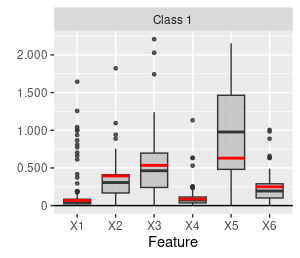}}};
		\node at (3.33, -9.795) {\resizebox{0.293\textwidth}{!}{\includegraphics{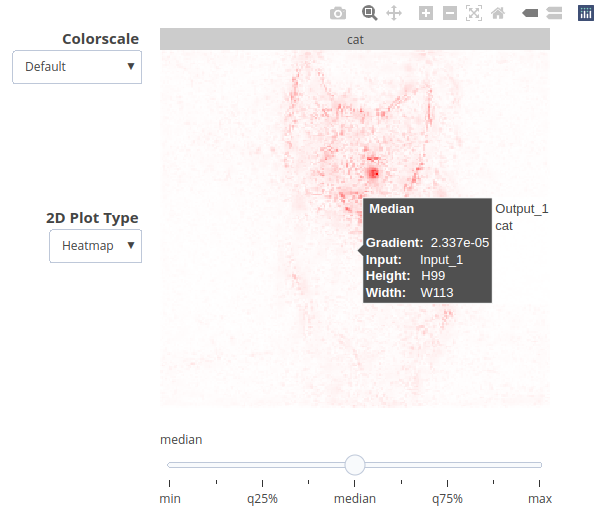}}};
		\node at (-1.025, -9.795) {\resizebox{0.293\textwidth}{!}{\includegraphics{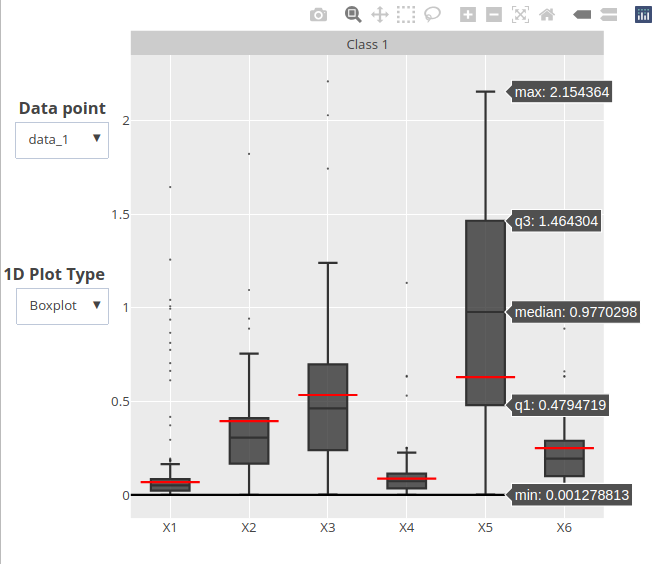}}};
	\end{scope}	
\end{tikzpicture}

%% file: tikz/melanoma_model.tikz
\begin{tikzpicture}
	\def\width{5}
	\def\height{0.75}
	\def\dist{0.5}
	
	\begin{scope}[shift = {(-8,0)}]
		\draw[rounded corners, fill = black!40] (0.5, 0) rectangle (\width - 0.5, \height) node[pos = 0.5] {Input $(*,10)$};
	
		\def\dist{1}
		\def\height{1.25}
		\draw[rounded corners, fill = black!10] (-1, -0.5 * \dist) rectangle (\width + 0.5, -3 * \height - 3.5 * \dist);
		\node[rotate = 90, scale = 1.2] at (-0.5, -1.5 * \height - 1.75 * \dist) {Tabular model};

		\draw[rounded corners, fill = black!20] (0, -1 * \dist) rectangle (\width, -1 * \dist - 1 * \height) node[pos = 0.5] {Dense-$32$ (ReLU)};
		\draw[rounded corners, fill = black!20] (0, -2 * \dist - 1 * \height) rectangle (\width, -2 * \dist - 2 * \height) node[pos = 0.5] {Dense-$16$ (ReLU)};
		\draw[rounded corners, fill = black!20] (0, -3 * \dist - 2 * \height) rectangle (\width, -3 * \dist - 3 * \height) node[pos = 0.5] {Dense-$8$ (linear)};
		
		\draw[->] (0.5*\width, -0*\dist - 0*\height) -- (0.5*\width,-1*\dist - 0*\height);
		\draw[->] (0.5*\width, -1*\dist - 1*\height) -- (0.5*\width,-2*\dist - 1*\height);
		\draw[->] (0.5*\width, -2*\dist - 2*\height) -- (0.5*\width,-3*\dist - 2*\height);
	\end{scope}
	
	\draw[rounded corners, fill = black!10] (-1.25, -0.5 * \dist) rectangle (\width + 2.5, -9 * \height - 6 * \dist);
	\draw[rounded corners, fill = black!15] (- 0.5*\dist, -0.875 * \dist ) rectangle (\width + 2.5 - 0.5*\dist, -4.125 * \dist - 6 * \height);
	\node[rotate = 90, scale = 1.2] at (-0.75, -4.5 * \height - 3 * \dist) {Image model}; 
	
	\draw[rounded corners, fill = black!40] (0.5, 0) rectangle (\width - 0.5, \height) node[pos = 0.5] {Input $(*,224,224,3)$};
	
	\draw[rounded corners, fill = black!20] (0, -1.25 * \dist) rectangle (\width, -1.25 * \dist - 1 * \height) node[pos = 0.5] {Conv2D-$n$ (ReLU)};
	\draw[rounded corners, fill = black!20] (0, -1.75 * \dist  - 1 * \height) rectangle (\width, -1.75 * \dist - 2 * \height) node[pos = 0.5] {2x Conv2D-$n$ (ReLU};
	\draw[rounded corners, fill = black!20] (0.25 * \width, -2.25 * \dist  - 2 * \height) rectangle (0.75*\width, -2.25 * \dist - 3 * \height) node[pos = 0.5] {Add};
	\draw[rounded corners, fill = black!20] (0, -2.75 * \dist  - 3 * \height) rectangle (\width, -2.75 * \dist - 4 * \height) node[pos = 0.5] {2x Conv2D-$n$ (ReLU)};
	\draw[rounded corners, fill = black!20] (0.25 * \width, -3.25 * \dist  - 4 * \height) rectangle (0.75 * \width, -3.25 * \dist - 5 * \height) node[pos = 0.5] {Add};
	\draw[rounded corners, fill = black!20] (0, -3.75 * \dist  - 5 * \height) rectangle (\width, -3.75 * \dist - 6 * \height) node[pos = 0.5] {AvgPool2D};
	
	\draw[->] (\width, -1.25*\dist - 0.5 * \height) -- (\width + 0.5, -1.25*\dist - 0.5 * \height) -- (\width + 0.5, -2.25*\dist - 2.25 * \height) -- (0.75*\width, -2.25*\dist - 2.25 * \height);
	\draw[->] (0.75 * \width, -2.25*\dist - 2.75 * \height) -- (\width + 0.5, -2.25*\dist - 2.75 * \height) -- (\width + 0.5, -3.25*\dist - 4.5 * \height) -- (0.75 * \width, -3.25*\dist - 4.5 * \height);
	
	\draw[rounded corners, fill = black!20] (0, -4.5 * \dist  - 6 * \height) rectangle (\width, -4.5 * \dist - 7 * \height) node[pos = 0.5] {Flatten};
	\draw[rounded corners, fill = black!20] (0, -5 * \dist  - 7 * \height) rectangle (\width, -5 * \dist - 8 * \height) node[pos = 0.5] {Dense-$512$ (ReLU)};
	\draw[rounded corners, fill = black!20] (0, -5.5 * \dist  - 8 * \height) rectangle (\width, -5.5 * \dist - 9 * \height) node[pos = 0.5] {Dense-$256$};
	\node[align = center, rotate = 90] at (\width + 1.37, -2.5 * \dist - 3 * \height) {for $n = 32, 64, 128$ and $256$};
	
	\draw[->] (0.5*\width, -0*\dist - 0*\height) -- (0.5*\width,-1.25*\dist - 0*\height);
	\draw[->] (0.5*\width, -1.25*\dist - 1*\height) -- (0.5*\width,-1.75*\dist - 1*\height);
	\draw[->] (0.5*\width, -1.75*\dist - 2*\height) -- (0.5*\width,-2.25*\dist - 2*\height);
	\draw[->] (0.5*\width, -2.25*\dist - 3*\height) -- (0.5*\width,-2.75*\dist - 3*\height);
	\draw[->] (0.5*\width, -2.75*\dist - 4*\height) -- (0.5*\width,-3.25*\dist - 4*\height);
	\draw[->] (0.5*\width, -3.25*\dist - 5*\height) -- (0.5*\width,-3.75*\dist - 5*\height);
	\draw[->] (0.5*\width, -3.75*\dist - 6*\height) -- (0.5*\width,-4.5*\dist - 6*\height);
	\draw[->] (0.5*\width, -4.5*\dist - 7*\height) -- (0.5*\width,-5*\dist - 7*\height);
	\draw[->] (0.5*\width, -5*\dist - 8*\height) -- (0.5*\width,-5.5*\dist - 8*\height);
	
	\begin{scope}[shift = {(-4, -11.5)}]
		\draw[rounded corners, fill = black!20] (-0.5*\width, 0) rectangle (1.5*\width, \height) node[pos = 0.5] {Concatenation};
		\draw[rounded corners, fill = black!20] (0, -1 * \dist) rectangle (\width, -1 * \dist - 1 * \height) node[pos = 0.5] {Dense-$256$ (ReLU)};
		\draw[rounded corners, fill = black!20] (0, -2 * \dist - 1 * \height) rectangle (\width, -2 * \dist - 2 * \height) node[pos = 0.5] {Dense-$1$ (Sigmoid)};
		
		\draw[->] (0.5*\width, -0*\dist - 0*\height) -- (0.5*\width,-1*\dist - 0*\height);
		\draw[->] (0.5*\width, -1*\dist - 1*\height) -- (0.5*\width,-2*\dist - 1*\height);
	\end{scope}
	
	\draw[->] (-8 + 0.5*\width, -3 * 1.25 - 3 * 1) -- (-8 + 0.5*\width, -10) -- (-4 + 0.5*\width, -10) -- (-4 + 0.5 * \width, -11.5 + \height);
	\draw[->] (0.5*\width, -5.5*\dist - 9*\height) -- (0.5*\width, -10) -- (-4 + 0.5*\width, -10) -- (-4 + 0.5 * \width, -11.5 + \height);
\end{tikzpicture}

%% file: tikz/appendix_models_1.tikz
\begin{tikzpicture}

	\begin{scope}[shift = {(-5.5,0)}]
		\draw[rounded corners, fill = black!40] (0.25,0) rectangle (3.75,0.75) node[pos = 0.5] {Input $(*,10)$};
		\draw[rounded corners, fill = black!20] (-0.25,-0.5) rectangle (4.25,-1.25) node[pos = 0.5] {Dense-$64$};
		\draw[rounded corners, fill = black!20] (-0.25,-1.75) rectangle (4.25,-2.5) node[pos = 0.5] {ReLU or Tanh};
		\draw[rounded corners, fill = black!20] (-0.25,-3) rectangle (4.25,-3.75) node[pos = 0.5] {Dense-$1$ or Dense-$5$};
		
		\draw[->] (2,0) -- (2,-0.49);
		\draw[->] (2,-1.25) -- (2,-1.74);
		\draw[->] (2,-2.5) -- (2,-2.99);
		
		\node at (2, 1.5) {\underline{\textbf{Tabular model}}};
	\end{scope}
	
	\draw[rounded corners, fill = black!40] (0.25,0) rectangle (3.75,0.75) node[pos = 0.5] {Input $(*, 32,32,3)$};
	\draw[rounded corners, fill = black!20] (-0.25,-0.5) rectangle (4.25,-1.25) node[pos = 0.5] {Conv2D-$5$};
	\draw[rounded corners, fill = black!20] (-0.25,-1.75) rectangle (4.25,-2.5) node[pos = 0.5] {ReLU or Tanh};
	\draw[rounded corners, fill = black!20] (-0.25,-3) rectangle (4.25,-3.75) node[pos = 0.5] {AvgPool or MaxPool};
	\draw[rounded corners, fill = black!20] (-0.25,-4.25) rectangle (4.25,-5) node[pos = 0.5] {Flatten};
	\draw[rounded corners, fill = black!20] (-0.25,-5.5) rectangle (4.25,-6.25) node[pos = 0.5] {Dense-$1$ or Dense-$5$};
	
	\draw[->] (2,0) -- (2,-0.49);
	\draw[->] (2,-1.25) -- (2,-1.74);
	\draw[->] (2,-2.5) -- (2,-2.99);
	\draw[->] (2,-3.75) -- (2,-4.24);
	\draw[->] (2,-5) -- (2,-5.49);
	
	\draw[->] (4.25, -2.125) to[out = 0, in = 90] (4.9, -3.375) to[out = -90, in = 0] (4.25, -4.6125);
	\node[rotate = -90, scale = 0.8] at (5.1, -3.375) {if no pooling};
	\node at (2, 1.5) {\underline{\textbf{Image model}}};
\end{tikzpicture}

%% file: tikz/time_models.tikz
\begin{tikzpicture}

    \begin{scope}[shift = {(-5.5,0)}]
    \draw[rounded corners, fill = black!10] (-0.5,-0.25) rectangle (4.5,-1.5);
    \node at (-1.25, -0.875) {$L - 1 \times$};
    
    \draw[rounded corners, fill = black!40] (0.25,0) rectangle (3.75,0.75) node[pos = 0.5] {Input $(B,10)$};
    \draw[rounded corners, fill = black!20] (-0.25,-0.5) rectangle (4.25,-1.25) node[pos = 0.5] {Dense-$U$};
    \draw[rounded corners, fill = black!20] (-0.25,-1.75) rectangle (4.25,-2.5) node[pos = 0.5] {ReLU};
    \draw[rounded corners, fill = black!20] (-0.25,-3) rectangle (4.25,-3.75) node[pos = 0.5] {Dense-$C$};
    
    \draw[->] (2,0) -- (2,-0.49);
    \draw[->] (2,-1.25) -- (2,-1.74);
    \draw[->] (2,-2.5) -- (2,-2.99);
    
    \node at (2, 1.5) {\underline{\textbf{Tabular model}}};
    \end{scope}
    
    
    \draw[rounded corners, fill = black!10] (-0.5,-0.25) rectangle (4.5,-2.75);
    \node at (5.25, -1.5) {$\times L - 1$};
    
    \draw[rounded corners, fill = black!40] (0.25,0) rectangle (3.75,0.75) node[pos = 0.5] {Input $(B,W,W,3)$};
    \draw[rounded corners, fill = black!20] (-0.25,-0.5) rectangle (4.25,-1.25) node[pos = 0.5] {Conv2D-$U$};
    \draw[rounded corners, fill = black!20] (-0.25,-1.75) rectangle (4.25,-2.5) node[pos = 0.5] {ReLU};
    \draw[rounded corners, fill = black!20] (-0.25,-3) rectangle (4.25,-3.75) node[pos = 0.5] {Conv2D-$U$};
    \draw[rounded corners, fill = black!20] (-0.25,-4.25) rectangle (4.25,-5) node[pos = 0.5] {ReLU};
    \draw[rounded corners, fill = black!20] (-0.25,-5.5) rectangle (4.25,-6.25) node[pos = 0.5] {Flatten};
    \draw[rounded corners, fill = black!20] (-0.25,-6.75) rectangle (4.25,-7.5) node[pos = 0.5] {Dense-$C$};
    
    \draw[->] (2,0) -- (2,-0.49);
    \draw[->] (2,-1.25) -- (2,-1.74);
    \draw[->] (2,-2.5) -- (2,-2.99);
    \draw[->] (2,-3.75) -- (2,-4.24);
    \draw[->] (2,-5) -- (2,-5.49);
    \draw[->] (2,-6.25) -- (2,-6.74);
    
    \node at (2, 1.5) {\underline{\textbf{Image model}}};
\end{tikzpicture}